\documentclass{article}
\usepackage{arxiv}

\usepackage[utf8]{inputenc}
\usepackage[T1]{fontenc}
\usepackage{newunicodechar}
\newunicodechar{’}{'}
\usepackage{amsmath,amsfonts,bm}

\def\eqref#1{equation~\ref{#1}}
\def\1{\bm{1}}

\DeclareMathAlphabet{\mathsfit}{\encodingdefault}{\sfdefault}{m}{sl}
\SetMathAlphabet{\mathsfit}{bold}{\encodingdefault}{\sfdefault}{bx}{n}

\usepackage{amssymb}
\usepackage{graphicx}
\usepackage{fontawesome5}
\usepackage{booktabs}
\usepackage{xcolor}
\usepackage{microtype}
\usepackage{natbib}

\usepackage{url}
\usepackage{hyperref}
\newcommand{\YiFeiORCID}{0000-0002-7427-1212}
\newcommand{\TianxinORCID}{0009-0001-0160-2228}
\newcommand{\HanboORCID}{0000-0003-1133-8544}
\newcommand{\JianQiaoORCID}{0000-0003-0133-9551}

\newcommand{\GitHubURL}{https://github.com/YifeiCAO/MindGym}

\newcommand{\authororcid}[1]{%
    \begingroup
    \edef\orcidvalue{#1}%
    \ifx\orcidvalue\empty\else
        \hspace{0.25em}\href{https://orcid.org/\orcidvalue}{%
            \raisebox{-0.12em}{\includegraphics[height=0.95em]{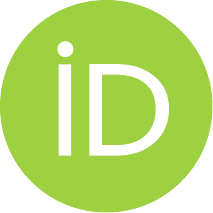}}}%
    \fi
    \endgroup
}
\newcommand{\projectlinks}{%
    \href{\GitHubURL}{\faGithub}\\[2pt]
    {\small\color{blue}\expandafter\url\expandafter{\GitHubURL}}%
}
\newcommand{\equalcontrib}{\footnotemark[1]\hspace{0.45em}}

\definecolor{GPTBlack}{HTML}{000000}
\definecolor{DeepSeekBlue}{HTML}{5786FE}
\definecolor{QwenPurple}{HTML}{6950EF}

\newcommand{\mglogo}[1]{%
    \raisebox{-0.15em}{%
        \includegraphics[height=1.15em]{assets/#1.png}%
    }%
}
\newcommand{\mgcell}[4]{%
    \shortstack{#2\\[-1pt]{\scriptsize\textcolor{gray}{[#3, #4]}}}%
}

\title{\texttt{mini-psych-agent}: Evaluating Human-Like Behavior in LLM Agents through Interactive Psychological Environments}

\author{
    Yi-Fei Cao\equalcontrib\authororcid{\YiFeiORCID} \\
    Department of Psychology\\
    University of California, Los Angeles\\
    \texttt{caoyifei@g.ucla.edu}
    \And
    Tianxin Huang\equalcontrib\authororcid{\TianxinORCID} \\
    School of Computer Science\\
    McGill University\\
    \texttt{tianxin.huang@mail.mcgill.ca}
    \AND
    Hanbo Xie\equalcontrib\authororcid{\HanboORCID} \\
    School of Psychological and Brain Sciences\\
    Georgia Institute of Technology\\
    \texttt{hxie88@gatech.edu}
    \And
    Jian-Qiao Zhu\authororcid{\JianQiaoORCID} \\
    Department of Psychology\\
    The University of Hong Kong\\
    \texttt{zhujq@hku.hk}
}

\date{\projectlinks}
\renewcommand{\shorttitle}{\texttt{mini-psych-agent}}
\hypersetup{
    pdftitle={mini-psych-agent: Evaluating Human-Like Behavior in LLM Agents through Interactive Psychological Environments},
    pdfauthor={Yi-Fei Cao, Tianxin Huang, Hanbo Xie, Jian-Qiao Zhu},
    pdfsubject={Interactive psychological environments and human-like behavior in LLM agents},
    pdfkeywords={LLM agents, psychological tasks, human behavior, agent memory},
    hidelinks
}

\begin{document}

\maketitle
\begingroup
\renewcommand{\thefootnote}{\fnsymbol{footnote}}
\footnotetext[1]{Equal contribution.}
\endgroup

\begin{abstract}

A Large Language Model (LLM), its harness, and the test-time compute budget jointly determine its behavior. Consequently, the same underlying model, embedded in different agentic workflows or allocated different amounts of test-time compute, can exhibit markedly different performance. Psychologists face an analogous challenge when evaluating human performance: the same person may behave differently under different incentives and levels of deliberation. Psychological tasks therefore provide an ideal testbed for studying such behavioral variation. However, when adapted to LLMs, these tasks are typically text-transcribed, replaying fixed, language-rendered experiments as static prompts. This format is ill-suited to agent evaluation, where actions can alter future observations and task states. Conversely, although dynamic environments are prevalent in the evaluation of LLM agents, they rarely ask whether the resulting behavior is human-like. Here, we bridge this gap by transforming psychological tasks into dynamic environments in which agents can interact and by using a learned human-policy model to evaluate how compatible the resulting behavioral trajectories are with human behavior. Across these environments, we find that this compatibility varies with an LLM's agentic configuration and the psychological task, and that memory raises it mainly by making choices more predictable rather than more human-like. These findings motivate a more holistic evaluation of LLM agents in psychology-inspired environments, accounting for both the backbone model and its agentic scaffold.

\end{abstract}

\section{Introduction}

\begin{figure}[t]
    \centering
    \includegraphics[width=\linewidth]{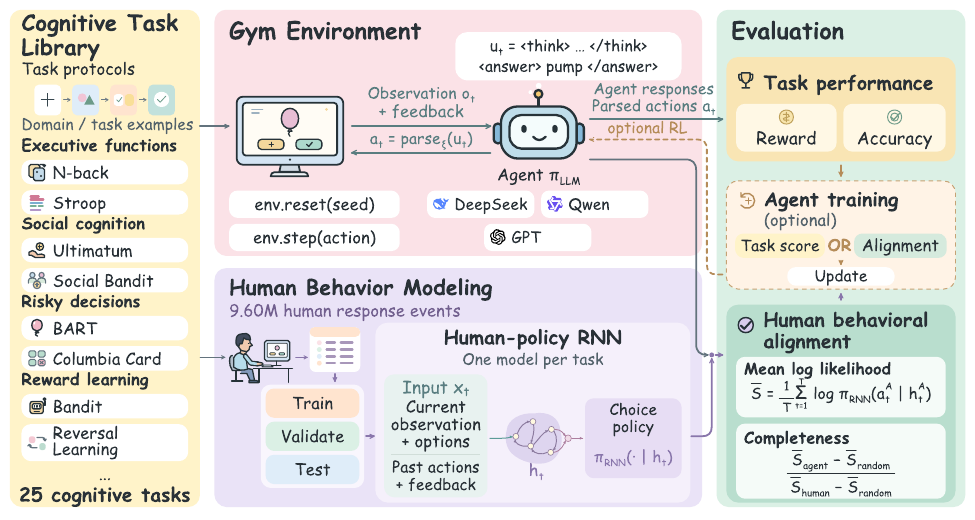}
    \caption{Overview of \texttt{MindGym}.
LLM agents interact with 25 psychological tasks through
a Gym-style interface. Their trajectories are evaluated
for task performance and alignment with task-specific
RNN models trained on human behavior.
Displayed tasks and backbones are illustrative.
No training or fine-tuning of LLMs were performed
in this study; the dashed training loop illustrates
a possible extension only.
}
    \label{fig:mindgym}

\end{figure}

Psychological tasks are originally designed to isolate and test specific aspects of human cognition. To understand aspects of machine cognition, it is therefore natural for psychologists to adapt their carefully designed tasks to examine the behavior of LLMs. However, a common approach to evaluating LLMs on psychological tasks is to replay, in natural language, the experimental experiences originally encountered by human participants \citep{binz2023using, binz2025foundation, binz2026post, bai2025explicitly, frank2026cognitive, Ryan2025, zhu2024incoherent}. For example, the Centaur model, which is a finetuned Llama-3.1-70B model using psychological data, transcribes each experiment into a text sequence containing the task instructions, stimuli, and trial-by-trial interaction history \citep{binz2025foundation}. At each response point, the LLM is conditioned on the participant’s experience up to, but not including, the response at that trial and evaluated according to the likelihood it assigns to the participant’s observed response.

Verbatim replay provides a controlled and statistically principled measure of human-likeness through next-response prediction, following a long tradition of cognitive modeling \citep{daw2011trial, wilson2019ten}. But it does not directly assess closed-loop behavior, in which an agent’s actions determine its subsequent observations and task states. As a result, replaying a human participant’s experience evaluates model behavior only under the distribution of task states and histories generated by the participant \citep{binz2025foundation, xie2025centaur}. This can conceal compounding errors: at every trial, the model is always returned to the participant’s recorded history and therefore never has to recover from the consequences of its own earlier non-human-like decisions.

But an autonomous agent may take an unusual action and enter a state that few or no human participants encountered. Replay provides no evidence about its behavior in such states. The supplied human history may also be unlikely under the model’s own policy. For example, the replay transcript may state that ``you previously chose option $A$,'' even though the model would almost certainly have chosen option $B$. The model is thus asked to predict a response conditional on a history that it would rarely generate through its own decisions.

Evaluating autonomous agents therefore also requires assessing behavior under the histories generated by their own actions. To support this evaluation, we developed \texttt{MindGym}, a suite of interactive environments comprising 25 distinct psychological tasks that probe executive functions (e.g., the N-back task), social cognition (e.g., the ultimatum game), risky choice (e.g., the Balloon Analogue Risk Task), and reward learning (e.g., bandit tasks). Unlike approaches that only transcribe participants’ experiences into text, we transform these tasks into interactive environments in which both humans and LLM agents can act through a defined set of permissible actions. As a result, these environments allow entirely new action sequences to be generated and evaluated under instructions similar to those given to the original human participants.

\texttt{MindGym} thus offers a unique opportunity to examine the behavior of LLM agents and evaluate their workflow designs through psychology-inspired environments. We created five \texttt{mini-psych-agent} variants by varying how agents maintain and use an external memory file (i.e., \texttt{MEMORY.md}). These variants range from a stateless baseline without persistent memory to more sophisticated approaches to context management, including retrieval-augmented generation (RAG) and surprisal-based episodic memory. External memory provides a practical mechanism for maintaining persistent agent state across model calls, allowing information from earlier interactions to be stored, updated, and retrieved to guide subsequent behavior.

Evaluating the human-likeness of \texttt{mini-psych-agent} variants in \texttt{MindGym} introduces an additional challenge: how to assess human-likeness when few or no matching human behavioral sequences are available. Our agents can generate novel trajectories that no human participant in the dataset has experienced. We address this challenge by training recurrent neural networks (RNNs) on human behavioral sequences, with one RNN for each of the 25 psychological tasks. Given sufficient human data, these RNNs can generalize to unseen histories within each task and serve as learned evaluators of agent behavior. We then freeze the RNNs and assess how compatible agent behavior is with the human policy using the likelihood these learned human-policy models assign to agent-generated actions, conditioned on the corresponding observations and interaction histories.

Our contributions are threefold. (i) We standardized and abstracted a range of psychological experiments into interactive environments in which agents can act and receive feedback. (ii) We developed novel likelihood metrics for evaluating how compatible behavioral sequences are with human policies learned by RNNs, reported alongside KL divergence and trajectory-level features. (iii) We systematically evaluated different memory-management strategies for LLM agents across interactive psychological environments.

\section{Related Work}

\textbf{Evaluating human alignment of LLMs.}
Psychological experiments are now widely adapted to characterize LLM behavior, often through static prompts such as vignettes, questionnaires, and choice problems \citep[e.g.,][]{binz2023using, zhu2024incoherent, bai2025explicitly, Ryan2025}. For example, CogGym standardizes 258 commonsense reasoning experiments and compares 50 LLMs with humans on matched trials \citep{ying2026coggym}. However, fewer studies evaluate LLMs in closed-loop interactions, where the model's own choices determine its subsequent task states \citep[e.g.,][]{binz2023using, codaforno2024cogbench, schubert2024incontext, akata2025playing}. These studies often summarize behavior using a few task-specific signatures, such as regret, stay probabilities, cooperation rates, or fitted model parameters (e.g., CogBench~\citep{codaforno2024cogbench} and
CogArena~\citep{fernandez2025cogarena}). Although informative, these summaries require bespoke analyses for each task and retain only selected aspects of trial-level behavior. In parallel, interactive environments are well established for training and evaluating non-LLM agents. Psychlab implements psychological experiments for both humans and deep RL agents \citep{leibo2018psychlab}, while NeuroGym provides neuroscience tasks for training neural networks \citep{molano2022neurogym}. For LLM agents, game-based benchmarks such as BALROG and TextArena provide interactive environments for evaluating task performance \citep{paglieri2025balrog, guertler2025textarena}. However, these benchmarks primarily assess game progress or skill ratings rather than the human-likeness of the LLM behavior. A remaining challenge is therefore to combine interactive environments with a common measure of human-likeness for trajectories generated by LLM agents.

\textbf{Learned models of human behavior.}
Psychologists commonly develop computational models to explain cognitive mechanisms or predict human behavior. Whether interpretable cognitive models or black-box neural networks, these models are typically themselves the object of evaluation. For sequential tasks, a common criterion is how well they predict each response from a participant's preceding history \citep{daw2011trial, wilson2019ten}. Flexible learned models increasingly outperform hand-crafted theories in risky and strategic choice \citep{peterson2021using, zhu2025capturing} and sequential decision-making \citep{dezfouli2019models, jian2025discovering}, while hybrid neural-cognitive models retain interpretable structure \citep{eckstein2026hybrid, cao2025hybrid}. In sufficiently data-rich settings, neural networks can also approach empirically estimated predictive limits \citep{kuperwajs2023using, yan2026understanding}, capturing maximal explainable variances in human data. Learned neural network models have also guided cognitive theory refinement through scientific regret minimization \citep{agrawal2020scaling} and provided behavioral representations for evaluating LLM-generated behavior \citep{lei2026computational}. In this paper, we use RNNs trained on human trajectories in each psychological task to evaluate the human-likeness of trajectories generated by autonomous LLM agents in the corresponding interactive environment. This extends the role of learned models from objects of evaluation to evaluators of novel agent-generated behavior within the same task.

\textbf{Memory management in LLM agents.}
An LLM agent's behavior depends not only on its backbone but also on its scaffold \citep{yang2024sweagent, kapoor2024agents}, of which memory management is a central component. CoALA frames LLM agents as cognitive architectures with working and long-term memory \citep{sumers2024coala}, while concrete designs include RAG \citep{lewis2020retrieval}, memory streams with retrieval and reflection \citep{park2023generative}, OS-inspired context management \citep{packer2023memgpt}, and surprise-based event segmentation \citep{fountas2025humaninspired}. Memory benchmarks, however, commonly emphasize task performance, including question answering over long interaction histories and, more recently, on-policy conversational evaluation \citep{wu2025longmemeval, jiayang2026amemgym}. Memory management also shapes decisions: in a study of multi-armed bandits, satisfactory exploration emerged only when GPT-4 combined chain-of-thought reasoning with an externally summarized interaction history \citep{krishnamurthy2024can}. Prior work has also examined how memory contributes to behavioral believability \citep{park2023generative} and persona-consistent decisions \citep{peng2026heart}. To our knowledge, however, no prior work has systematically examined how different memory-management strategies affect the human-likeness of agent-generated trajectories across a suite of interactive psychological tasks. We address this gap by testing whether a memory design elicits more human-like behavior and how this relationship varies across tasks.

\section{Evaluation Framework}
\label{sec:framework}

\textbf{Interactive psychological environments.}
\label{sec:environments}
Following the agent-environment interface used by OpenAI Gym \citep{brockman2016gym}, we implement the  \texttt{MindGym} environments using Gymnasium \citep{towers2024gymnasium}. Each environment supports interaction through natural-language instructions and observations, together with structured, task-specific actions. Specifically, each environment maintains an internal state that evolves in response to the agent's actions. As illustrated in Figure~\ref{fig:mindgym}, interaction under task configuration $\xi$ proceeds as
\begin{align}
    u_t \sim \pi_\text{LLM}(\cdot \mid c_t), ~~
     a_t = \operatorname{parse}_\xi(u_t), ~~
    s_{t+1} \sim P_\xi(\cdot \mid s_t, a_t), ~~
     o_{t+1} = g_\xi(s_{t+1}).
\end{align}
where $t$ denotes the decision step and $c_t$ is the input context constructed by the agent runtime. This context combines the task instructions and current textual observation $o_t$ with any memory or interaction history supplied by the agent architecture (see Equation \ref{eq:agent_runtime_construction} for details). Conditioned on $c_t$, the LLM generates its response $u_t$ under policy $\pi$, which is further parsed into an executable action $a_t$, which updates the internal environment state $s_t$ through the transition kernel $P_\xi$. The rendering function $g_\xi$ exposes participant-visible information and feedback permitted by the task protocol, while internal scoring and diagnostic information are recorded separately. Repeated interaction produces a behavioral trajectory shaped by the agent's own choices and their consequences.

The present study evaluates 25 psychological tasks spanning executive control, feedback learning and adaptation, information acquisition, social learning and strategic interaction, risk and decision making, planning and transfer, memory retrieval and integration, and affective updating. These categories reflect the primary demands of the implemented tasks. We draw on human behavioral datasets from
Psych-201 \citep{binz2026post} and datasets
identified through OpenCogData \citep{opencogdata}.
We selected tasks with human behavioral data from more
than 500 participants to support reliable training and
held-out evaluation of task-specific RNN models of
human behavior. For each task, we draw on published paradigms and available experimental materials to specify the available actions, outcome dynamics, feedback schedule, and stopping conditions. The full task inventory and adaptations required for language-based interaction are documented in Appendix~\ref{app:environments}.

\textbf{LLM agents and memory configurations.}
\label{sec:agents}
We implement \texttt{mini-psych-agent} as a lightweight, model-agnostic harness that combines an LLM with explicit memory management and an execution loop. At decision step $t$ under task $\xi$, the runtime constructs the input context, obtains an LLM response, validates the proposed decision, and updates memory after execution:
\begin{align}
    c_t = C(I_\xi, o_t, m_t^\phi, \mathcal{A}_t),~~
     u_t \sim \pi_\text{LLM}(\cdot \mid c_t), ~~
    a_t = \operatorname{parse}_\xi(u_t),~~
     M_{t+1} = U_\phi(M_t, e_t). 
     \label{eq:agent_runtime_construction}
\end{align}
where $C(\cdot)$ constructs the context, and $U_\phi$ updates memory according to memory-management strategy $\phi$. Here, $I_\xi$ denotes the task instructions, $M_t$ the stored memory state, $m_t^\phi$ the memory content supplied to the LLM under strategy $\phi$, and $\mathcal{A}_t$ defines the set of valid decisions. The tuple $e_t=(o_t,a_t,o_{t+1})$ records the completed, participant-visible transition. Only validated actions are executed. Memory is reset between runs (i.e., $M_0=\varnothing$), and the LLM's parameters remain fixed.

Let $H_t=(e_0,\ldots,e_{t-1})$ denote the recorded public interaction history. We compare five configurations that differ in the memory content supplied to the LLM:
\begin{equation}
m_t^\phi =
\begin{cases}
\varnothing,
    & \text{Stateless}, \\[2pt]
H_t,
    & \text{Cumulative}, \\[2pt]
\operatorname{Last}_{6}(H_t),
    & \text{Window}, \\[2pt]
\operatorname{Top}_{6}(H_t;\rho_t),
    & \text{Retrieval}, \\[2pt]
(\sigma_1,\ldots,\sigma_{J_t}) \mathbin{\Vert} B_t,
    & \text{Surprise}.
\end{cases}
\label{eq:memory_configurations}
\end{equation}
where $\Vert$ denotes concatenation. All configurations also receive the current observation $o_t$ and task instructions $I_\xi$. The Stateless agent therefore receives no additional memory beyond these inputs. On the other extreme, the Cumulative agent always receives its full recorded interaction history at every decision step. 

Between the Stateless and Cumulative extremes, we consider more selective memory strategies inspired by cognitive psychology \citep{miller1956magical, tulving1973encoding}. The Window agent receives up to six most recent interaction records. The Retrieval agent instead receives up to six records with the highest relevance scores, $\rho_t(e_i)=\cos(\operatorname{tf}(o_t),\operatorname{tf}(o_i))$, where $\operatorname{tf}$ denotes word-count vectors and ties favor more recent records.

Finally, inspired by surprise-based event segmentation in EM-LLM \citep{fountas2025humaninspired}, our Surprise agent uses elicited forecasts of public outcomes or stimuli to partition experience into events. Each closed event $E_j$ is replaced by a textual summary $\sigma_j=\operatorname{Summarize}_\text{LLM}(E_j)$. At decision step $t$, the agent receives the $J_t$ completed summaries in chronological order, followed by the current event's uncompressed records $B_t$.

\textbf{Learned human-policy model.}
\label{sec:human_policy}
To assess human-like behavior in states visited by few or no human participants, we train a separate RNN on human sequences for each of the 25 psychological tasks. Each model is trained to approximate the human policy conditioned on the current task context and preceding interaction history, providing a task-specific behavioral reference:
\begin{align}
\pi_\text{RNN}(a_t \mid x_{1:t}) \approx \pi_\text{human}(a_t \mid x_{1:t}),
\label{eq:rnn_approximate_human_policy}
\end{align}
where $x_{1:t}$ denotes the sequence of structured inputs available before action $a_t$ is selected, including the current stimuli or options and preceding responses and feedback. That is, $x_{1:t}$ is an abstraction of $c_t$ in Equation \ref{eq:agent_runtime_construction}.

Crucially, our RNNs use symbolic encodings of task features and responses, following common practice in neural network models of human behavior \citep{jian2025discovering, eckstein2026hybrid, cao2025hybrid, kuperwajs2023using, yan2026understanding}. Unlike LLM-based approaches that represent psychological tasks in natural language \citep{binz2023using}, our RNN models learn directly from these structured representations. They therefore operate on task abstractions rather than directly processing the stimuli presented to participants, as ``stimulus-computable'' models do \citep{frank2026cognitive}. This abstraction focuses learning on task-relevant behavioral dependencies, with the aim of supporting generalization to novel interaction histories within the same task.

\textbf{Training task-specific RNNs.}
For each task, we use a single-layer gated recurrent unit (GRU) \citep{cho2014learning} to maintain a recurrent state and generate task-specific predictions:
\begin{equation}
    h_t = \operatorname{GRU}
    \bigl(h_{t-1}, x_t \bigr),
    ~~
    a_t = g(h_t),
\end{equation}
where $x_t$ encodes the task inputs available before response $a_t$, $h_t$ is the hidden state of GRU, and $g(\cdot)$ produces predictions $a_t$, comprising categorical action probabilities, continuous predictions, or both, depending on the psychological task. Recurrent states are reset at task-defined sequence boundaries.

We trained these RNN models through next-action prediction with teacher forcing: each human response is predicted from the current task inputs and preceding observed human responses and feedback. Categorical decisions are trained with cross-entropy, and continuous decisions with task-specific regression losses; tasks with multiple response types combine the corresponding objectives. Human data are partitioned approximately 80/10/10 into training, validation, and test sets by source-defined identity groups, with each group assigned to a single partition. Checkpoints are selected based on validation performance and frozen for subsequent evaluation. Task-specific encodings, losses, and training configurations are detailed in Appendix~\ref{app:human_policy}.

\begin{figure}[t]
    \centering
    \includegraphics[width=\linewidth]{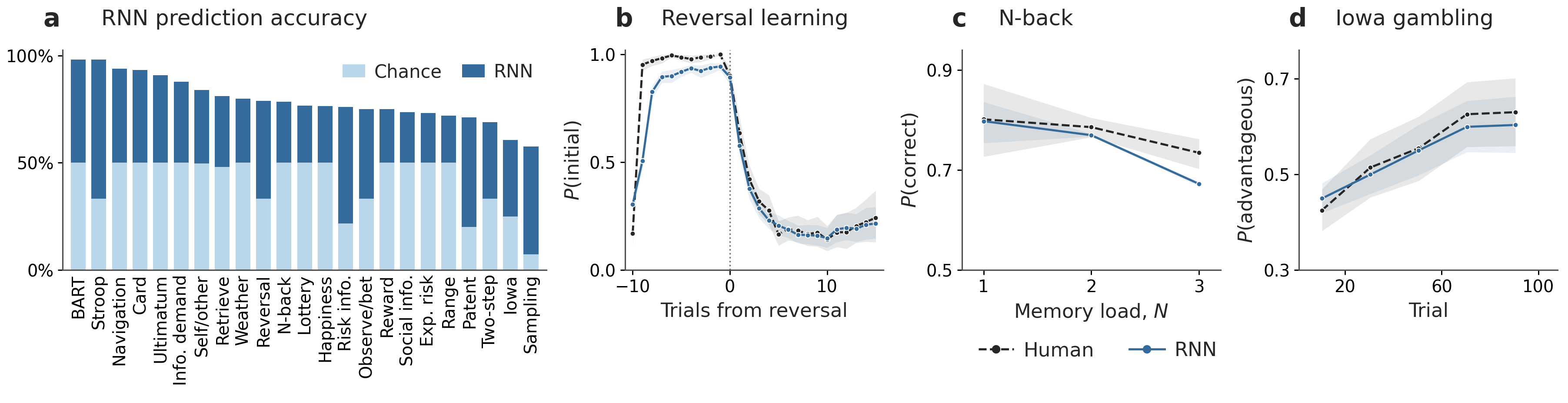}
    \caption{
    \textbf{Predictive validation of the human-policy RNN models.}
    \textbf{(a)} Held-out response prediction accuracy across
    the 23 tasks with categorical decisions. Full bar heights show
    RNN accuracy; pale blue marks uniform-choice chance levels.
    \textbf{(b--d)} Human responses and RNN probabilities for
    choosing the pre-reversal best option, responding correctly
    in N-back, and choosing advantageous Iowa decks, respectively.
    Predictions condition on recorded human histories.
    Shading denotes pointwise 95\% participant-session cluster
    bootstrap intervals.
    }
    \label{fig:rnn_results}
\end{figure}

\textbf{Predictive validation.}
\label{sec:human_policy_validation}
On held-out human trajectories, the RNNs exceeded uniform-random prediction accuracy by $37.6 \pm 11.4$ percentage points (mean $\pm$ SD across tasks), with above-chance performance on all 23 tasks with categorical decisions (Figure~\ref{fig:rnn_results}a). The RNNs also capture qualitative patterns in human behavior, including behavioral adaptation following reversals in the reversal learning task (see Figure~\ref{fig:rnn_results}b), declining accuracy with increasing N-back load (see Figure~\ref{fig:rnn_results}c), and increasing preference for advantageous decks in the Iowa Gambling Task (see Figure~\ref{fig:rnn_results}d). These findings are consistent with previous RNN modeling results and support the use of learned RNNs as black-box behavioral references for human trajectories \citep{kuperwajs2023using, jian2025discovering, eckstein2026hybrid, yan2026understanding}. Full predictive metrics, including results for continuous decisions, are reported in Appendix~\ref{app:human_policy}. Appendix~\ref{app:reference_validation} extends these behavioral checks using qualitative features from 16 tasks: the RNN curves stay within 0.06 of the human curves in 14 tasks and within 0.09 in the other two.

\textbf{Behavioral alignment.}
We now turn to our primary evaluation objective: quantifying the human-likeness of behavioral trajectories generated by LLM agents using a learned human-policy RNN. For task $\xi$, we use $\pi_{\text{RNN}}(a_t \mid x_{1:t})$ to score the agent's action $a_t = \operatorname{parse}_\xi(u_t)$, where $u_t \sim \pi_{\text{LLM}}(\cdot \mid c_t)$ and $x_{1:t}$ contains the current task inputs and the agent's own preceding interaction history. The RNN thus serves as a learned proxy for how human participants might respond under agent-generated histories, including those rarely or never observed in the human data. This evaluation relies on the RNN's ability to generalize within the task beyond the human trajectories on which it was trained.

Specifically, let $\tau^A=(o_1^A, a_1^A, o_2^A, a_2^A,\ldots,o_T^A,a_T^A)$ denote an LLM agent's trajectory, consisting of $T$ observations and corresponding actions. At each decision step, we evaluate the agent's action using $\pi_\text{RNN}(a_t^A \mid x_{1:t}^A)=\pi_\text{RNN}(a_t^A \mid h_t^A)$, where $h_t^A$ is the RNN's hidden state after processing the structured interaction history $(o_1^A,a_1^A,\ldots,o_t^A)$, before observing action $a_t^A$. We calculate a human-likeness score based on mean log likelihoods, which measure the likelihood assigned to the agent's actions by the RNN. Formally, the score $\bar{S} = \frac{1}{T}\sum_{t=1}^T \log \pi_\text{RNN}(a_t^A \mid h_t^A)$, which is the mean of the per-decision log likelihoods assigned by the RNN along the agent's trajectory (see Appendix~\ref{app:justify_rnn_scores}).

\textbf{KL divergence from the human policy.}
Because $\bar S$ is computed from the agent's sampled actions alone, it increases both when the agent's action distribution moves toward the human policy and when the agent merely chooses more deterministically (Equation~\ref{eq: score_action entropy_KL}). To separate the two, we also estimate the agent's action entropy $\mathcal{H}_\text{Agent}$ and the per-decision Kullback--Leibler divergence $D_{\mathrm{KL}}$ of the agent's action distribution from $\pi_\text{RNN}$ along the agent's own trajectory. $D_{\mathrm{KL}}$ is zero when the agent chooses with the RNN's probabilities and equals $-\bar S$ for a deterministic policy, so it does not reward repetitive choice. Both quantities require the agent's action probabilities, which we approximate using decision tokens' log probabilities that a backbone returns them (i.e., the open-weight DeepSeek and Qwen models, see Appendix~\ref{app:kl} for details).

\textbf{Optimal behavior in psychological tasks.}
Of the 25 psychological tasks, 17 have a clear definition of optimal behavior, including the Stroop and N-back tasks. In these tasks, behavior can be evaluated against an optimal decision $a_t^*$ as a measure of task performance. The remaining tasks elicit preferences from human participants and agents, without a clearly defined optimality baseline. These 17 tasks therefore allow us to examine alignment with both human behavior and optimal behavior.

\section{Experiments}

We evaluated \texttt{mini-psych-agent} on 25 psychological tasks across GPT-5.6 Luna~\citep{openai2026gpt56luna}, DeepSeek V4.1 Flash~\citep{deepseekai2026deepseekv41flash}, and Qwen3.8 Flash~\citep{alibaba2026qwen38flash}. Each backbone LLM was evaluated under five memory configurations (Stateless, Cumulative, Window, Retrieval, and Surprise), with five environment seeds per task, matched across configurations. Our analyses included 1,690 episodes across the 23 tasks with categorical decisions. Results for the 2 tasks with continuous or mixed decisions are reported in Appendix~\ref{app:continuous_results}.

For the subset of 17 tasks with a defined optimality baseline, we measured task performance as the proportion of decisions that exactly matched the task-specific optimal action. We assessed human-likeness using $\bar S$ and its random-to-human completeness, defined as $\frac{\bar S_{\mathrm{agent}}-\bar S_{\mathrm{random}}}{\bar S_{\mathrm{human}}-\bar S_{\mathrm{random}}}$. Comparisons between memory configurations were paired by seed within each task and backbone, and cross-task summaries weighted tasks equally. Unless otherwise stated, intervals are 95\% confidence intervals obtained by bootstrapping tasks. Details of decoding and episode accounting are provided in Appendix~\ref{app:agents}.

\textbf{Memory effects on task performance.}
\label{sec:memory_performance}
We evaluated how memory influences LLM agents' task performance, averaging task-specific optimality rates for cross-task aggregation. Across the 17 tasks, averaging the four memory configurations yielded a performance gain of $0.075$ over Stateless (CI $[0.030,0.123]$), with positive gains in 14 tasks. At the task--backbone level, individual memory configurations improved performance relative to Stateless in 149 of 204 comparisons (73.0\%), reduced it in 45, and tied in 10. Qwen showed consistent gains, improving in 15 of 17 tasks, with a mean gain of $0.084$ (CI $[0.048,0.123]$).

\begin{figure*}[t]
    \centering
    \includegraphics[width=\textwidth]{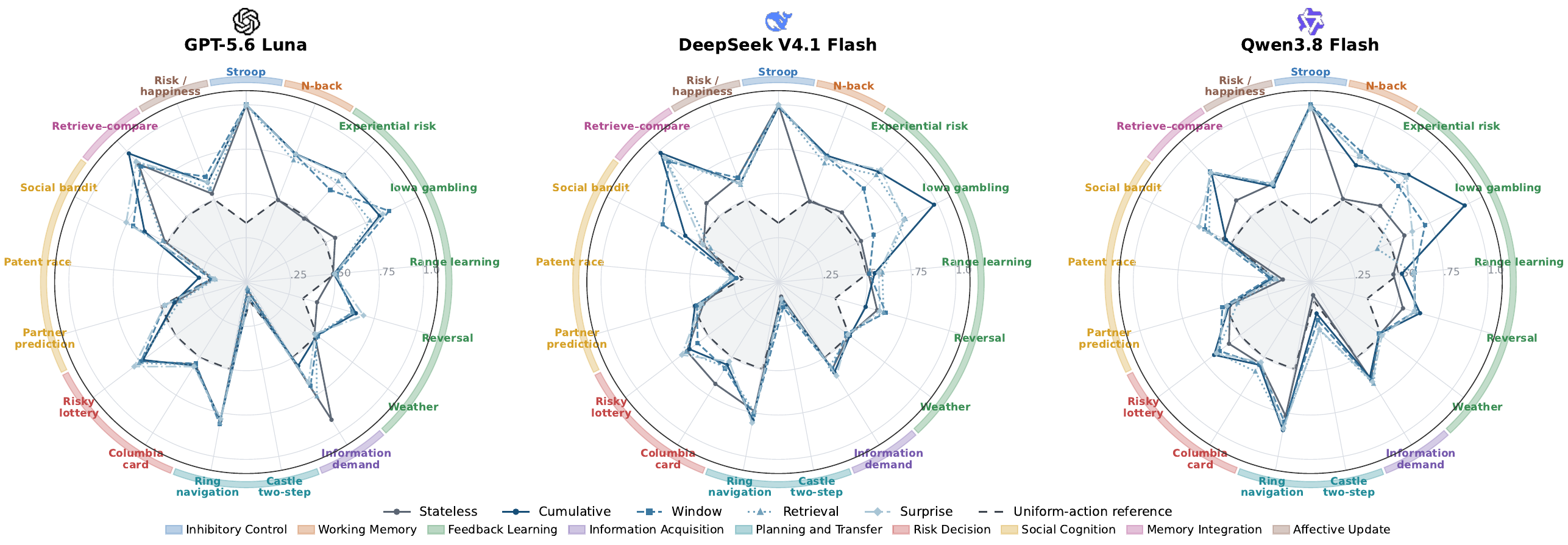}
    \caption{
    Jagged intelligence visualized using optimality rates across memory configurations and backbone LLMs on the 17 psychological tasks with clearly defined optimality baselines. Outer colors denote categories for psychological tasks, and lines represent memory configurations. Each axis shows a task-specific optimality rate, with 1 indicating that every decision matches an optimal decision. For the Risky lottery and Risk / happiness tasks, optimality is defined by the risk-neutral choice with the higher expected value; ties and decisions with ambiguous probabilities are excluded. The dashed boundary and gray region show the optimality baseline for an agent that selects uniformly among the available actions.
    }
    \label{fig:task_performance_radar}

\end{figure*}

As shown in Figure~\ref{fig:task_performance_radar}, more specific patterns emerged across tasks with different history requirements. In experiential risk learning, Cumulative outperformed Window in all
15 paired backbone--seed comparisons, with a mean gain of 10.9
percentage points in optimal-choice rate. In Castle two-step,
its optimal-terminal rate was 1.5 percentage points lower than
Window's on average, with positive differences in 4 of 15 comparisons. In N-back, Window outperformed Retrieval by 3.3 percentage points in balanced accuracy, with gains in all 15 comparisons. These contrasts suggest benefits from accumulated experience in experiential risk learning and from preserving recent observations in their original order in N-back. The best memory configuration nevertheless varied across tasks and backbone LLMs (Appendix~\ref{app:performance_statistics}).

\textbf{Memory effects on human-likeness.}
\label{sec:memory_alignment}
We find that including interaction history in the context supplied to the LLM improves average alignment with the learned human policy. Across the 23 tasks, mean completeness increased from $0.664$ under Stateless to $0.885$ when averaged over the four other memory configurations, which provide interaction history in different forms ($\Delta=0.222$, CI $[0.119,0.336]$). Mean log likelihood $\bar S$ also increased by $0.168$ nats per decision (CI $[0.096,0.246]$). All four memory configurations had higher task-averaged completeness than Stateless for every backbone, with Cumulative scoring highest (Table~\ref{tab:memory_overall}). Averaged across the four configurations, completeness gains were $0.125$ for GPT, $0.284$ for DeepSeek, and $0.256$ for Qwen. Holm-corrected Wilcoxon tests yielded $p=0.030$, $p=0.015$, and $p<0.001$, respectively (Appendix~\ref{app:inference}).

Task-specific memory effects were also evident as measured by $\bar S$. Cumulative exceeded Window by $0.234$ nats in experiential risk learning and $0.251$ nats in Castle two-step. In N-back, Window exceeded Retrieval by $0.127$ nats, with positive differences in all 15 paired backbone--seed comparisons. These patterns suggest that the benefits of memory depend on the history demands of particular tasks, although better human-policy fit need not coincide with better task performance, as illustrated by Castle two-step.

\begin{figure*}[t!]
    \centering
    \includegraphics[width=\textwidth]{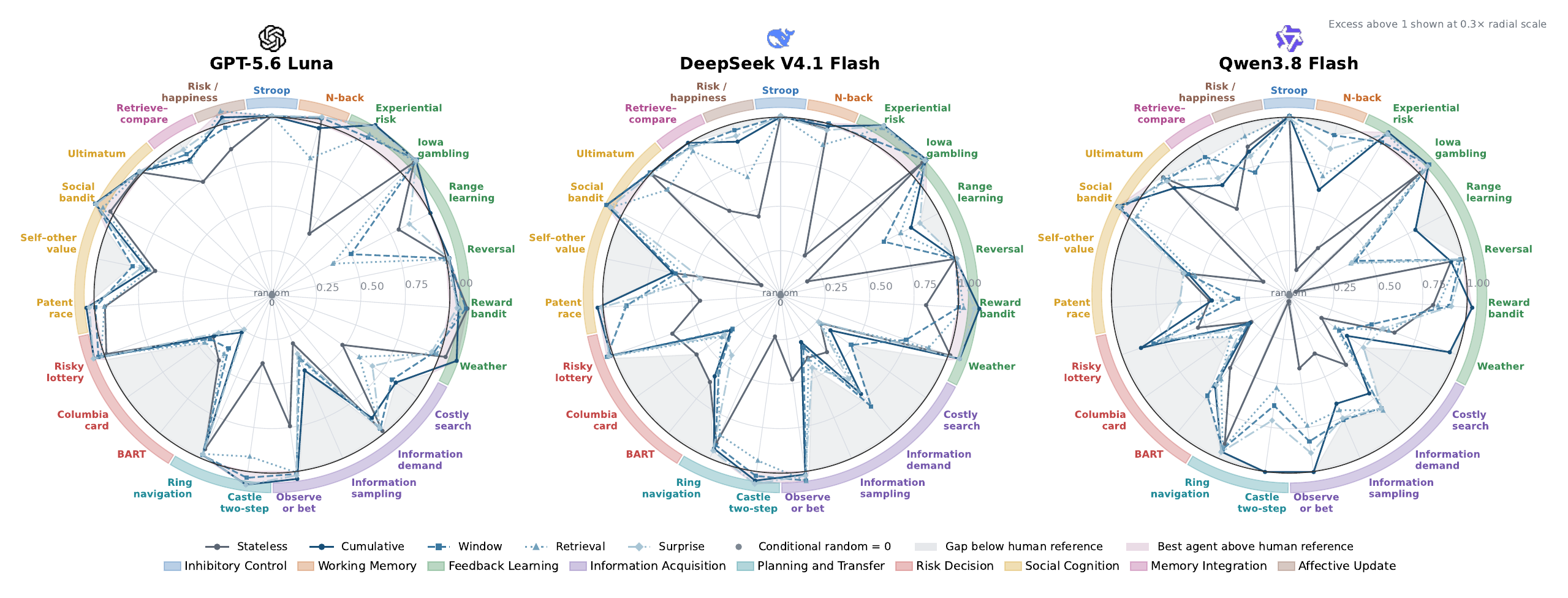}
    \caption{
    Jagged human alignment visualized using random-to-human completeness across the 23 psychological tasks with categorical decisions. The random reference averages RNN log likelihoods uniformly over valid actions along agent-visited histories; the human reference scores held-out participants on their own histories. Completeness of 0 therefore corresponds to the conditional random baseline, and 1 to the average held-out human log likelihood under the task-specific RNN. The human reference is a benchmark rather than an upper bound, and values are not clipped. Gray shading shows the shortfall of the highest agent completeness relative to 1; purple shading indicates completeness above 1. Excess above 1 is displayed with a radial scaling factor of 0.3. Outer colors denote psychological task categories, and lines represent memory configurations.
    }
    \label{fig:human_policy_completeness_radar}
\end{figure*}

Task-level completeness profiles, which express gains in human-policy fit over the random baseline relative to the held-out human reference, are shown in Figure~\ref{fig:human_policy_completeness_radar}. Compact memory configurations captured much of the aggregate benefit. Window and Surprise retained 79\% and 90\% of Cumulative's mean completeness gain over Stateless while using 9\% and 34\% of its prompt tokens, respectively, including memory-update calls. These ratios are computed from task-averaged gains and token use (see Appendix~\ref{app:agents} for details). Cumulative's mean completeness marginally exceeded Window's by $0.056$, although the paired task-level difference was not statistically significant ($p=0.273$; see Appendix~\ref{app:inference} for details).

\begin{table*}[t]
    \centering
    \small
    \setlength{\tabcolsep}{7pt}
    \renewcommand{\arraystretch}{1.15}
    \caption{
    Average random-to-human completeness across the 23 psychological tasks with categorical decisions. 
    }
    \begin{tabular}{@{}lccccc@{}}
        \toprule
        \textbf{Model}
        & Stateless & Cumulative & Window & Retrieval & Surprise \\
        \midrule

        \mglogo{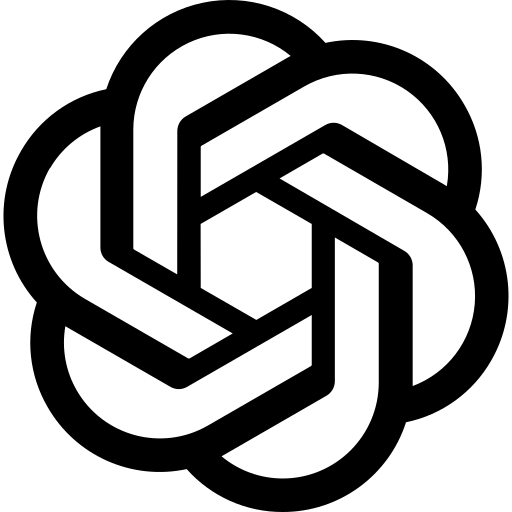}\;
        \textcolor{GPTBlack}{GPT-5.6 Luna}
        & \mgcell{GPTBlack}{0.833}{0.709}{0.954}
        & \mgcell{GPTBlack}{\textbf{1.005}}{0.880}{1.120}
        & \mgcell{GPTBlack}{0.951}{0.833}{1.061}
        & \mgcell{GPTBlack}{0.903}{0.789}{1.012}
        & \mgcell{GPTBlack}{0.971}{0.843}{1.086} \\

        \addlinespace[3pt]

        \mglogo{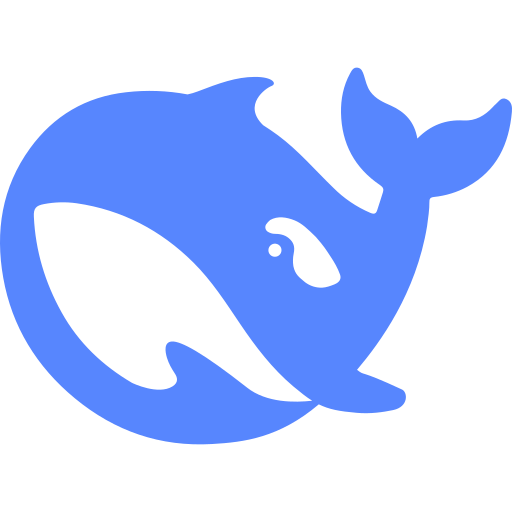}\;
        \textcolor{DeepSeekBlue}{DeepSeek V4.1 Flash}
        & \mgcell{DeepSeekBlue}{0.623}{0.494}{0.757}
        & \mgcell{DeepSeekBlue}{\textbf{0.944}}{0.808}{1.072}
        & \mgcell{DeepSeekBlue}{0.889}{0.775}{1.001}
        & \mgcell{DeepSeekBlue}{0.869}{0.749}{0.986}
        & \mgcell{DeepSeekBlue}{0.926}{0.791}{1.053} \\

        \addlinespace[3pt]

        \mglogo{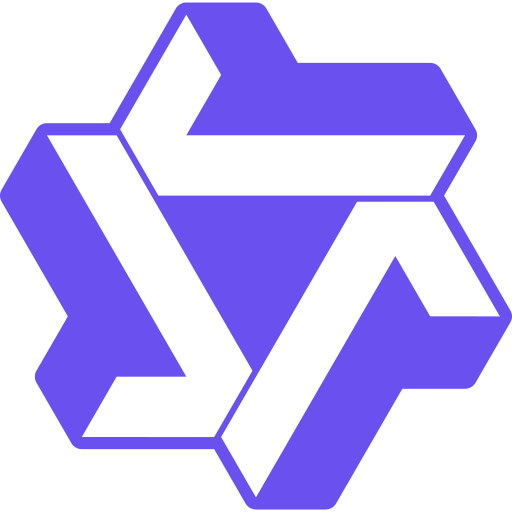}\;
        \textcolor{QwenPurple}{Qwen3.8 Flash}
        & \mgcell{QwenPurple}{0.536}{0.409}{0.664}
        & \mgcell{QwenPurple}{\textbf{0.835}}{0.729}{0.940}
        & \mgcell{QwenPurple}{0.775}{0.665}{0.881}
        & \mgcell{QwenPurple}{0.750}{0.651}{0.846}
        & \mgcell{QwenPurple}{0.806}{0.709}{0.904} \\

        \bottomrule
    \end{tabular} \\
    \textit{Note.} Brackets show 95\% confidence intervals obtained by bootstrapping tasks; boldface indicates the highest mean for each backbone LLM. Within each task, completeness is computed from episode-averaged log likelihoods using matched seeds, with 1 indicating that the agent's $\bar S$ matches the held-out human reference score.
    \label{tab:memory_overall}
\end{table*}

\textbf{Choice predictability and KL divergence from the human policy.}
We estimated action entropy and $D_{\mathrm{KL}}$ on the original trajectories of DeepSeek and Qwen for 20 tasks (Appendix~\ref{app:kl}); GPT-5.6 Luna is a closed-source model whose interface returned no log probabilities. Compared to Stateless, Memory lowered action entropy by 0.150--0.192 nats for DeepSeek and 0.099--0.127 nats for Qwen, which accounts for 76--95\% of the gain in $\bar S$ under Cumulative, Window, and Surprise; under Retrieval, the KL divergence increased. The KL divergence itself changed less (Table~\ref{tab:kl_overall}). It was lowest under Cumulative for both backbones, by $0.054$ relative to Stateless for DeepSeek (CI $[0.019,0.092]$) and $0.031$ for Qwen (CI $[-0.009,0.072]$), and highest under Retrieval. Cumulative therefore remains the configuration closest to the human policy, but most of the improvement in Table~\ref{tab:memory_overall} reflects more predictable choices. Because action probabilities are approximated from model logits and limited only to open-weight models, we treat $D_{\mathrm{KL}}$ as a complementary metric.

\begin{table*}[t]
    \centering
    \small
    \setlength{\tabcolsep}{7pt}
    \renewcommand{\arraystretch}{1.15}
    \caption{
    Divergence $D_{\mathrm{KL}}$ of the agent's action distribution from the human-policy RNN (nats per decision; lower is closer), averaged over the 20 tasks for which action probabilities could be recovered. 
    }
    \begin{tabular}{@{}lccccc@{}}
        \toprule
        \textbf{Model}
        & Stateless & Cumulative & Window & Retrieval & Surprise \\
        \midrule
        \mglogo{deepseek}\;
        \textcolor{DeepSeekBlue}{DeepSeek V4.1 Flash}
        & \mgcell{DeepSeekBlue}{0.200}{0.154}{0.249}
        & \mgcell{DeepSeekBlue}{\textbf{0.146}}{0.108}{0.187}
        & \mgcell{DeepSeekBlue}{0.192}{0.147}{0.242}
        & \mgcell{DeepSeekBlue}{0.241}{0.179}{0.307}
        & \mgcell{DeepSeekBlue}{0.186}{0.123}{0.269} \\
        \addlinespace[3pt]
        \mglogo{qwen}\;
        \textcolor{QwenPurple}{Qwen3.8 Flash}
        & \mgcell{QwenPurple}{0.166}{0.121}{0.215}
        & \mgcell{QwenPurple}{\textbf{0.136}}{0.096}{0.180}
        & \mgcell{QwenPurple}{0.148}{0.107}{0.193}
        & \mgcell{QwenPurple}{0.186}{0.134}{0.243}
        & \mgcell{QwenPurple}{0.139}{0.104}{0.177} \\
        \bottomrule
    \end{tabular} \\
    \textit{Note.} Brackets show 95\% confidence intervals obtained by bootstrapping tasks; boldface indicates the lowest mean for each backbone LLM. GPT-5.6 Luna returned no token log probabilities.
    \label{tab:kl_overall}
\end{table*}

\textbf{Relationship between task performance and human-likeness.}
\label{sec:performance_alignment}
Task performance and completeness were positively associated on average. Within each task and backbone, we computed Spearman's correlation between mean performance and mean completeness across all five memory configurations, then averaged these correlations across backbones within each task. Across the 17 tasks with both measures, Spearman's $\rho$ was $0.428 \pm 0.329$ (mean $\pm$ SD across tasks; CI $[0.278,0.580]$).

However, not all tasks showed positive correlation between task performance and human-likeness from agent's trajectories. For instance, N-back illustrates how the two measures can diverge within a sequential task. GPT's balanced accuracy increased from $(49.7 \pm 0.4)\%$ under Stateless to $(77.5 \pm 1.6)\%$ under Cumulative (mean $\pm$ SD across five seeds), while $\bar S$ decreased from $-0.365$ to $-0.472$ nats and completeness fell from $1.101$ to $0.975$. Specifically, Stateless responded with ``non-match'' on 96.3\% of trials. An always ``non-match'' policy likewise achieved chance-level balanced accuracy but a score of $\bar S=-0.249$, exceeding the held-out humans' mean score of $\bar S=-0.458$. These results illustrate how repetitive choices can receive high likelihood under a history-conditioned human-policy model. Completeness therefore expresses mean log likelihood relative to the random and human references; it complements task performance but does not measure a distance between behavioral distributions (see Appendix~\ref{app:human_policy}). The KL divergence in Table~\ref{tab:kl_overall} does not share this property: the always ``non-match'' policy in the N-back task has $D_{\mathrm{KL}}=0.249$, whereas the memory configurations of DeepSeek and Qwen reach $0.042$--$0.197$ (see Appendix~\ref{app:kl} for details).

\textbf{Interaction histories and autonomous behavior.}
\label{sec:interaction_history}
We compared replay and closed-loop interaction across four tasks, two backbones (DeepSeek and Qwen), four memory configurations, and five seeds, yielding 160 matched episode pairs. In the replay condition, the environment followed a fixed trajectory sampled from the frozen human-policy RNN, and the forced actions were recorded in memory. Agent proposals were scored but did not alter the trajectory. Mean within-episode disagreement between proposed and forced actions ranged from 43.1\% to 53.0\% across configurations. In Castle two-step, the available second-stage options differed between matched closed-loop and replay trajectories in 66.6\% of rounds. Mean memory gains in $\bar S$ were $0.389$ and $0.265$ nats under replay and closed-loop interaction, respectively (difference $0.124$ nats; task-cluster bootstrap CI $[-0.117,0.519]$). The highest-scoring configuration differed between conditions in four of the eight task--backbone combinations. These comparisons reveal how agents' own choices alter the states they encounter, alongside changes in memory configuration's rankings (see Appendix~\ref{app:replay}).

We also compared complete trajectories from LLM agents with those generated autonomously by the frozen human-policy models across nine branching tasks. For each configuration, we generated 10,000 reference experiments, each matched to its initial scenarios and episode count. In Castle two-step, GPT with Cumulative memory achieved completeness of $1.249$ but repeated its first-door choice on 95.8\% of training transitions, compared with a reference mean of 49.5\% (95\% predictive interval $[39.0\%,62.6\%]$). These checks reveal differences in sustained behavioral patterns even when the completeness measure is high.

\section{Discussion}

We introduced \texttt{MindGym}, a suite of psychology-inspired interactive environments for evaluating LLM agents. 
Compared with approaches that turn human experiences in psychological tasks into static prompts, \texttt{MindGym} supports closed-loop evaluation, where agents make their own decisions and influence subsequent task states. Agents can therefore be evaluated under their own interaction histories without repeatedly returning to a human participant's experience. This introduces an evaluation challenge: little or no human data may be available for states rarely or never visited by human participants. We address this by using task-specific RNNs trained on human trajectories as learned behavioral references for evaluating agent-generated trajectories.

Using \texttt{mini-psych-agent}, we systematically varied how LLM agents manage their external memory files and incorporate past interactions into the context at each decision step. 
Overall, retaining more past interactions in context generally improved both task performance and compatibility with the human policy, although no single memory strategy performed best across all tasks and backbones. For the two backbones whose action probabilities we could measure, most of the gain in $\bar S$ reflected more deterministic choice; KL divergence from the human policy fell modestly and was lowest under Cumulative (Table~\ref{tab:kl_overall}). At the level of complete trajectories, memory moved history-dependent behavior toward the human range but also made agents more likely to repeat previous decisions than humans (Appendix~\ref{app:trajectory}; Figures~\ref{fig:agent_behavior_gpt}--\ref{fig:agent_behavior_qwen}). We therefore interpret the memory gain as greater predictability and compatibility with the human policy rather than as greater human-likeness. Cumulative outperformed Window in feedback learning and planning tasks, while Window outperformed Retrieval in working memory tasks. These patterns suggest that memory benefits depend on both the amount of past experience retained and the preservation of recent temporal order. Window and Surprise nevertheless captured much of Cumulative's benefit for compatibility while using substantially fewer tokens in prompts for decision making.

Alignment with optimal behavior and alignment with human behavior were found to be positively related, but could also diverge in certain tasks. This suggests that greater optimality can coincide with higher compatibility with the human policy in some cognitive domains. Agents with high compatibility scores could nevertheless exhibit qualitatively distinct behavioral patterns over complete trajectories. Therefore, task performance, conditional human-policy fit, and behavior across complete trajectories provide complementary perspectives on closed-loop interaction, supporting a more holistic evaluation of autonomous agents.

\textbf{Limitations and future research.}
Our environments and agents represent early steps toward psychology-inspired evaluation of LLM agents. We also note several limitations. First, the human-policy RNNs were validated on human histories, whereas agents can reach states that people rarely visit. Therefore, likelihood scores from these RNNs are extrapolations, and a more direct test would require human responses collected from agent-generated histories (Appendix~\ref{app:reference_validation}). Second, the environments are still text-based: they preserve the choice structure of each task but not perceptual interference or response timing, as in Stroop, and no human data were collected on the text interface, so our claims concern choice behavior. Third, $D_{\mathrm{KL}}$ is potentially a better metric, although it is only available for open-weight models and has to be approximated using model logits. Additional psychological tasks could be implemented as interactive environments within \texttt{MindGym} to examine a broader range of cognitive processes. More sophisticated memory strategies, particularly those that modify stored representations of individual trials, could be incorporated into \texttt{mini-psych-agent}, alongside additional skills and tool-use capabilities. Another promising direction is to train LLM agents to behave more like humans within psychological environments, using learned human-policy models as training targets. We believe these efforts could help develop autonomous agents whose decisions are better aligned with human behavior.

\clearpage
\subsection*{AI use statement}

We used generative AI tools to assist with implementing
and debugging code. LLM-generated trajectories are experimental outputs of the
evaluated agents and are distinguished from the human
behavioral data used to train the RNNs. The authors take responsibility for the final methods,
analyses, claims, and artifacts presented in this work.

\subsection*{Ethics statement}

This work uses previously collected human behavioral
datasets and computational experiments with LLM agents.
The learned human-policy models reflect patterns in their
source populations and should not be interpreted as
universal models of human cognition. The reported alignment
scores are not intended for clinical assessment or
individual psychological profiling, and higher scores do
not necessarily imply human-like reasoning or optimal
behavior. Data access and any redistribution remain
subject to the original datasets' licensing, access,
and privacy requirements.

\subsection*{Reproducibility statement}

Task protocols and environment adaptations are documented
in Appendix~\ref{app:environments}. RNN input encodings,
data splits, training configurations, and predictive
validation are described in
Appendix~\ref{app:human_policy}. Agent configurations,
decoding settings, and episode accounting are reported
in Appendix~\ref{app:agents}. Score definitions and their
interpretation are provided in
Appendix~\ref{app:justify_rnn_scores}, while aggregation,
seed matching, and statistical procedures are detailed
in Appendix~\ref{app:inference}.

We provide the code and experimental data in an online repository.\footnote{\url{https://github.com/YifeiCAO/MindGym}}

\bibliography{references,dataset_sources}
\bibliographystyle{unsrtnat}

\clearpage
\appendix

\section{Illustrations for interactive psychological environments in \texttt{MindGym}}

\label{app:environments}

Each environment exposes task instructions, a textual observation, and the currently legal responses. Executing a response updates the task state and returns the next observation, any protocol-permitted feedback, and a termination signal. The observation contains participant-visible information: for example, previously revealed outcomes, current offers, or a revealed portion of a payoff matrix. Hidden reward parameters, unrevealed outcomes, and evaluator-only optimality calculations are kept separate. The number of environment steps need not equal the number of experimental trials: information search, navigation, and multistage decisions can require several actions within one trial or round.

Tables~\ref{tab:environment_inventory_a}--\ref{tab:environment_inventory_b} summarize the 25 implemented tasks. All are untimed text adaptations; visual displays are described symbolically and response times are not modeled. Source schedules and feedback conditions are retained where available, while interface instructions and cover stories are not uniformly reproduced. Iowa gambling, momentary mood, and sparse-event estimation include declared reconstructions where original implementation details are unavailable.

\begin{table}[htbp]
\centering
\small
\setlength{\tabcolsep}{4pt}
\renewcommand{\arraystretch}{1.16}
\caption{Environment inventory (1/2). Data citations identify the source human datasets. Counts distinguish choices, trials, and rounds; a round may contain multiple actions.}
\label{tab:environment_inventory_a}
\begin{tabular}{@{}p{0.22\textwidth}p{\dimexpr0.78\textwidth-2\tabcolsep\relax}@{}}
\toprule
Task & Interaction and task-specific adaptation \\
\midrule
Stroop & \textit{Data:} \citet{busch2024individual}. Report the annotated color of a color word from three alternatives over 84 trials. Ink color is described in text, so the interface does not reproduce visual interference or response-time effects. \\
\addlinespace[3pt]
N-back & \textit{Data:} \citet{enkavi2019large}. Judge letter matches over 20 adaptive blocks, each with 20 scored positions and $N$ initial context positions. Load changes between blocks; trial feedback and the source 0-back control are absent. \\
\addlinespace[3pt]
Experiential risk & \textit{Data:} \citet{radulescu2020convergent}. Choose between labeled stimuli in 110 free choices, interleaved with 66 forced-exposure events presented as feedback. Some outcome pools deplete with use; stimulus images and spatial layout are abstracted. \\
\addlinespace[3pt]
Iowa gambling & \textit{Data:} \citet{steingroever2015data}. Make 100 draws from four decks, observing gains, losses, and balance. Published payoff tables instantiate source-specific schemes; finite-deck depletion includes an explicitly reconstructed rule. \\
\addlinespace[3pt]
Range learning & \textit{Data:} \citet{bavard2021two}. Choose between labeled symbols over 120 learning and 120 transfer trials. Source conditions determine block structure and partial or complete outcome feedback; practice is omitted. \\
\addlinespace[3pt]
Reversal learning & \textit{Data:} \citet{suthaharan2021paranoia,reed2020paranoia}. Choose among three machines over four 40-choice blocks. Only the selected outcome is shown; recent choices can trigger reward-probability reversals. Machines replace the source card/deck display. \\
\addlinespace[3pt]
Reward bandit & \textit{Data:} \citet{dan2025behavior}. Make 100 binary choices under hidden fixed reward schedules. The agent sees only the chosen reward outcome, remaining choices, and accumulated rewards; graphical rewards are rendered as text. \\
\addlinespace[3pt]
Weather & \textit{Data:} \citet{mi2026human}. Predict between two alternatives from cue counts replacing icons. Source sessions determine the cue sequence and length; feedback stages reveal the rewarded side, while test stages withhold outcome feedback. \\
\addlinespace[3pt]
Information sampling & \textit{Data:} \citet{eisenberg2019uncovering}. Reveal boxes on a $5\times5$ text board or choose its majority color. There are 10 fixed-reward and 10 decreasing-reward rounds; opening boxes reduces correct-answer rewards only in the latter condition. \\
\addlinespace[3pt]
Information demand & \textit{Data:} \citet{rischall2023inefficient}. Four 126-trial blocks combine lottery queries with Observe, Estimate, or Intervene responses. Bonuses are settled at block end; the middle-block order is seeded, and source practice/quiz stages are omitted. \\
\addlinespace[3pt]
Observe-or-bet & \textit{Data:} \citet{anvari2024testing}. Choose to observe a light or bet on red/blue: 25 practice choices followed by two 50-choice paying blocks. Bets hide immediate outcomes; observations reveal color and block ends reveal total score. \\
\addlinespace[3pt]
Costly search & \textit{Data:} \citet{krueger2024identifying}. Across 20 rounds, pay to reveal cells of a four-color, six-gamble payoff matrix, then choose a gamble. Text replaces the query interface; color probabilities and query charges are public. \\
\addlinespace[3pt]
Castle two-step & \textit{Data:} \citet{tomov2021multi}. Learn stable door mappings across 100 training rounds and one transfer round. Each round has two three-door choices; the terminal resource bundle and price-weighted reward appear after the second choice. \\
\bottomrule
\end{tabular}
\end{table}

\begin{table}[htbp]
\centering
\small
\setlength{\tabcolsep}{4pt}
\renewcommand{\arraystretch}{1.16}
\caption{Environment inventory (2/2). Data citations identify the source human datasets. Subjective reports and preference choices remain distinct from objective task-performance measures.}
\label{tab:environment_inventory_b}
\begin{tabular}{@{}p{0.22\textwidth}p{\dimexpr0.78\textwidth-2\tabcolsep\relax}@{}}
\toprule
Task & Interaction and task-specific adaptation \\
\midrule
Ring navigation & \textit{Data:} \citet{tomov2020discovery}. Move clockwise/counterclockwise on a text-described 9- or 10-node ring. Evaluated sessions contain 80 goal-reaching training rounds and one test probe, which terminates after its first valid move. \\
\addlinespace[3pt]
Risky lottery & \textit{Data:} \citet{peterson2021using}. Choose between two lotteries for 20 problems repeated five times. Payoffs, disclosed probabilities, and source feedback conditions are rendered as text; some problems contain undisclosed probabilities. \\
\addlinespace[3pt]
BART & \textit{Data:} \citet{frey2017risk}. Pump or cash out across 30 balloons. Each pump adds a pending point, while explosion removes that balloon's pending points; hidden source explosion thresholds are randomly ordered. \\
\addlinespace[3pt]
Columbia card & \textit{Data:} \citet{frey2017risk}. Turn another card or stop over 84 rounds, initially containing 32 cards each. Gain, loss, and loss-card count are public; sequential card outcomes and terminal scores are presented as text. \\
\addlinespace[3pt]
Patent race & \textit{Data:} \citet{hunter2022increased}. Invest 0--4 from a four-unit endowment over 80 rounds. A strictly larger investment wins a ten-unit prize; opponent investments are sampled from round-specific recorded-choice pools and revealed afterward. \\
\addlinespace[3pt]
Self-other & \textit{Data:} \citet{barnby2022knowing}. Make 18 own-allocation choices, 36 partner-choice predictions, two intention ratings, and one attribution. Allocations are text; recorded partner choices and prediction feedback provide the social-learning sequence. \\
\addlinespace[3pt]
Social bandit & \textit{Data:} \citet{hertz2021trusting}. Choose a lake over three 15-choice blocks with private outcomes and condition-specific observation/advice. Social information appears at its scheduled event; the source's additional final stage is omitted. \\
\addlinespace[3pt]
Ultimatum & \textit{Data:} \citet{heffner2022probabilistic}. Accept or reject 20 shuffled offers dividing 100 points. Acceptance implements the offer; rejection yields zero to both players. Only responder choices are retained, excluding source emotion ratings and third-party roles. \\
\addlinespace[3pt]
Retrieve-compare & \textit{Data:} \citet{garcia2023experiential}. Learn labeled-symbol values, then make transfer choices and probability reports. Source-scheduled practice and feedback conditions are retained; pie charts become stated probabilities and reports use 21 levels from 0\% to 100\%. \\
\addlinespace[3pt]
Risk/happiness & \textit{Data:} \citet{rutledge2023risky,rutledge2014computational}. Make 30 safe-versus-risky choices interleaved with 12 happiness reports, beginning before the first choice. Offers and outcomes are text; happiness uses a 0--100 response scale. \\
\addlinespace[3pt]
Momentary mood & \textit{Data:} \citet{mantas2023experimental}. Report baseline and post-reward mood on a 101-level scale from $-1$ to $1$. Mood-dependent rewards and progression use a declared reconstruction; screening and stopping can make session length variable. \\
\addlinespace[3pt]
Sparse estimation & \textit{Data:} \citet{shin2021biased}. Record numerical events in two 40-event contexts, then estimate both means. The untimed reconstruction includes source-dependent fillers/probes; integer-entry scales are public, and the human reference scores the final estimates. \\
\bottomrule
\end{tabular}
\end{table}

\section{Human-policy models and predictive validation}
\label{app:human_policy}

Each task has a separately fitted RNN model on training human trajectories. Encodings represent visible stimuli, protocol stage, and causally available response/outcome history (Table~\ref{tab:human_policy_config}). Discrete identities use indicators; numerical features use task-specific scaling. Candidate-based choices pool masked candidate encodings before recurrent updating, with shared utility heads scoring legal options. Previous responses are shifted: the probability of response $t$ is computed before that response enters history. Observation-only events can update recurrent state without contributing a response loss. State resets at source sequence boundaries.

Categorical heads minimize binary or multiclass cross-entropy, masking illegal actions and unscored events. Retrieve-compare uses separate two-choice and 21-category report heads; costly search uses conditional typed heads. Mood uses mean squared error; sparse-event estimation uses squared error normalized by protocol slider range. Happiness and self--other models combine categorical losses with a weight-10 smooth-$L_1$ rating loss on the $[0,1]$ scale. Continuous outputs are evaluated in native units. Adam-family optimization and validation-based checkpoint selection are used; the table records selected widths, learning rates, batch sizes, maximum epoch budgets, and stopping patience. Maximum epochs are budgets, not fitted epoch counts. Held-out groups are source-defined participant/session identities in the saved splits.

Across 23 categorical tasks, mean saved test NLL is $0.491\pm0.297$ nats (task mean $\pm$ task standard deviation). The participant-sequence-balanced human anchor, averaged equally across tasks, is $\bar S=-0.541$: it uses sequence means and agent-scored endpoints, including joint rather than head-averaged likelihood for costly search. Thus the summaries have different aggregation and endpoint conventions. Table~\ref{tab:human_policy_complete} reports categorical and continuous test metrics. Categorical trajectory scores use a probability floor of $10^{-12}$.

A sampled-reference check covers 16 tasks: the mean absolute task-level difference from the held-out human participants $\bar S$ is $0.031$ nats, with 12 tasks below $0.05$. Fourteen tasks average five environment seeds; N-back and Stroop average 64 and four unique yoked stimulus schedules, respectively. Sampled responses condition on their own preceding responses; yoked schedules are retained rather than regenerated adaptively.

N-back illustrates why predictive likelihood complements performance. Always choosing nonmatch yields $\bar S=-0.249$ but balanced accuracy $0.500$, compared with the human anchor $\bar S=-0.458$. GPT Stateless chooses nonmatch on $96.3\%$ of responses; cumulative memory raises balanced accuracy from $49.7\%$ to $77.5\%$ while lowering $\bar S$ from $-0.365$ to $-0.472$. Predictable repetition can receive high conditional likelihood without reproducing the human behavioral distribution.

\begin{table}[htbp]
\centering
\caption{Task encodings and frozen training configurations. Configuration entries give encoder/GRU width, followed by learning rate/batch size/maximum epochs/stopping patience. R denotes a continuous head; typed heads depend on response stage.}
\label{tab:human_policy_config}
\scriptsize
\setlength{\tabcolsep}{2.5pt}
\renewcommand{\arraystretch}{1.18}
\begin{tabular}{@{}p{0.22\textwidth}p{0.43\textwidth}p{0.30\textwidth}@{}}
\toprule
Task & Encoding; head & Configuration \\
\midrule
BART & Pumps, outcomes; 2 & 16/24;.003/24/20/6 \\
Columbia card & Risk, gain/loss, cards; 2 & 16/24;.003/24/160/10 \\
Experiential risk & Stimuli, exposures, outcomes; 2 & 24/32;.003/64/100/10 \\
Ring navigation & Station, goal, visible graph; 2 & 32/48;.002/64/120/8 \\
Information sampling & Revealed boxes, payoff; 27 & 20/32;.003/20/80/8 \\
Information demand & Offers, stage, public reveal; typed 2 & 24/40;.003/64/160/10 \\
Iowa gambling & Deck-specific choice/outcome history; 4 & 16/24;.003/24/40/6 \\
Weather & Six cues, sun side, phase; 2 & 48/96;.002/16/300/60 \\
Castle two-step & Stage, prices, past resources; 3 & 24/32;.003/64/220/12 \\
N-back & Stimulus identity, $n$, block; 2 & 16/24;.003/24/40/6 \\
Observe-or-bet & Revealed evidence, block; 3 & 16/24;.003/24/80/6 \\
Patent race & Endowments, investment, history; 5 & 20/32;.003/128/80/8 \\
Range learning & Cue identities, phase, feedback; 2 & 20/28;.003/16/300/60 \\
Reward bandit & Options, prior outcomes; 2 & 16/24;.003/256/80/10 \\
Risky lottery & Payoffs, probabilities, feedback; 2 & 24/32;.002/128/80/4 \\
Ultimatum & Offer, trial, offer history; 2 & 16/24;.003/24/40/6 \\
Retrieve-compare & Cues, phase, revealed outcomes; 2/21 & 48/48;.002/32/160/15 \\
Reversal learning & Option-specific outcome history; choice & 16/24;.003/24/40/6 \\
Momentary mood & Rewards, score, previous mood; R & 24/32;.003/128/200/12 \\
Risk/happiness & Offers, event type, past responses; 2+R & 32/48;.002/512/12/3 \\
Costly search & Probabilities, cost, revealed payoffs; typed & 32/48;.002/64/20/5 \\
Social bandit & Options, visible social cues; 2 & 24/32;.003/512/60/8 \\
Self--other & Offers, event type, prior feedback; 2/3+R & 32/48;.002/64/160/10 \\
Sparse estimation & Event type, revealed amounts; R & 32/48;.002/64/100/12 \\
Stroop & Ink/word/candidate relations; choice & 16/24;.003/24/40/6 \\
\bottomrule
\end{tabular}
\end{table}

\begin{table}[htbp]
\centering
\caption{Held-out human-response prediction. $N$: train/validation/test identity groups; $T$: held-out targets. Accuracy/NLL pairs are unitless/nats. Baselines are the source-designated logistic or grouped linear-choice model (L), or histogram gradient boosting (H). Dashes indicate unavailable comparisons. Continuous errors retain native units.}
\label{tab:human_policy_complete}
\begingroup\scriptsize
\setlength{\tabcolsep}{2.5pt}
\resizebox{\textwidth}{!}{%
\begin{tabular}{lrrrr}
\toprule
Task & $N$ & $T$ & GRU acc./NLL & Baseline acc./NLL \\
\midrule
BART & 1125/141/141 & 157,919 & 0.9822/0.0673 & L: 0.9819/0.0832 \\
Columbia card & 1125/141/140 & 63,335 & 0.9321/0.1922 & -- \\
Experiential risk & 451/56/56 & 6,160 & 0.7310/0.5114 & H: 0.7093/0.5351 \\
Ring navigation & 507/64/63 & 12,666 & 0.9392/0.1581 & H: 0.9386/0.1692 \\
Information sampling & 418/52/52 & 13,924 & 0.5753/1.4460 & -- \\
Information demand & 476/60/59 & 44,604 & 0.8776/0.2752 & H: 0.8517/0.3262 \\
Iowa gambling & 416/52/52 & 5,685 & 0.6065/0.9541 & L: 0.5147/1.1189 \\
Weather & 459/58/57 & 37,032 & 0.7982/0.4278 & H: 0.5511/0.6870 \\
Castle two-step & 493/62/62 & 12,524 & 0.6886/0.6623 & H: 0.5088/0.9402 \\
N-back & 376/47/47 & 17,990 & 0.7843/0.4613 & L: 0.6504/0.6429 \\
Observe-or-bet & 667/84/83 & 10,375 & 0.7502/0.5932 & -- \\
Patent race & 595/74/74 & 5,920 & 0.7113/0.7536 & H: 0.6767/0.8199 \\
Range learning & 640/80/80 & 19,194 & 0.7203/0.5211 & H: 0.7071/0.5280 \\
Reward bandit & 2666/333/333 & 33,300 & 0.7499/0.4923 & L: 0.7058/0.5569 \\
Risky lottery & 4625/578/578 & 57,800 & 0.7658/0.4620 & -- \\
Ultimatum & 535/67/67 & 1,340 & 0.9075/0.2194 & L: 0.7746/0.4348 \\
Retrieve-compare & 539/67/67 & 28,015 & 0.8118/0.4349 & -- \\
Reversal learning & 730/91/91 & 14,560 & 0.7887/0.5030 & L: 0.7044/0.6527 \\
Momentary mood & 963/121/120 & 7,313 & --/-- & -- \\
Risk/happiness & 37578/4697/4697 & 259,830 & 0.7646/0.4876 & H: 0.7307/0.5348 \\
Costly search & 1486/186/186 & 52,430 & 0.7605/0.6679 & H: 0.7910/0.5893 \\
Social bandit & 1191/149/149 & 6,705 & 0.7363/0.5172 & H: 0.7289/0.5271 \\
Self--other & 502/63/62 & 3,410 & 0.8393/0.3736 & H: 0.6850/0.5779 \\
Sparse estimation & 725/91/90 & 180 & --/-- & -- \\
Stroop & 91/11/11 & 924 & 0.9805/0.1042 & L: 0.9805/0.2548 \\
\bottomrule
\end{tabular}}
\par\smallskip
\resizebox{\textwidth}{!}{%
\begin{tabular}{llrrrrrr}
\toprule
Continuous endpoint & Units & $N_{\rm test}$ & $T$ & GRU MAE & HGB MAE & GRU RMSE & HGB RMSE \\
\midrule
Momentary mood & $[-1,1]$ & 120 & 7,313 & .1340 & .1405 & .2211 & .2297 \\
Happiness & $[0,100]$ & 4,697 & 103,932 & 9.6777 & 11.8931 & 15.1726 & 17.1013 \\
Harmful intent & $[0,100]$ & 62 & 62 & 20.3984 & 20.8407 & 28.0039 & 26.4278 \\
Self-interest & $[0,100]$ & 62 & 62 & 25.6910 & 27.2246 & 33.8311 & 33.9800 \\
Sparse estimate: post-training & Original slider points & 90 & 90 & 14.7322 & 17.0084 & 23.4832 & 25.0049 \\
Sparse estimate: post-test & Original slider points & 90 & 90 & 20.5072 & 19.5686 & 25.2041 & 24.0355 \\
\bottomrule
\end{tabular}}
\par\smallskip
\begin{minipage}{\textwidth}
Costly search metrics average typed heads (52,430 targets over 26,215 action events); trajectory scoring uses joint likelihood. Sparse-event slider ranges are 300/500; normalized RMSE is .050912/.052674 for GRU and .054583/.049906 for HGB (post-training/post-test). Retrieve-compare's categorical probability reports also have MAE 14.8996 percentage points over 1,245 reports. Continuous ratings do not enter categorical NLL averages.
\end{minipage}
\endgroup
\end{table}

\section{Agent Configurations and Episode Accounting}
\label{app:agents}

\paragraph{Memory and decoding.}
Window and Retrieval provide at most six interaction records; Retrieval
limits each retrieved interaction to 2,000 characters. Surprise uses
the preceding ten valid forecast scores for the same target, with a
five-score warm-up, and closes an event when surprise exceeds their
mean plus one standard deviation. Events contain at least two and at
most 32 completed decisions; summaries contain at most 1,800 characters.
The current event remains uncompressed.
All three backbones run without extended reasoning. DeepSeek and Qwen
use temperature 1 and top-$p$ 1; GPT uses provider-default sampling.
Actions are returned as structured JSON and checked against the current
legal action set. Invalid or truncated actions terminate the episode
without executing a substitute action. Failed runs are recorded
separately; retained episodes contain no fallback actions.

\paragraph{Episode accounting.}
The design specifies 1,875 episodes: 25 tasks, three backbones, five
configurations, and five environment seeds. Of these, 1,812 completed,
50 terminated at the momentary-mood responsiveness screen, and 13 were
incomplete. The latter comprise six sparse-estimation episodes and
seven categorical-task episodes. Among the 1,718 completed categorical
episodes, 1,690 have a matching seed in every configuration and enter
the 23-task human-policy analysis. The full matched performance cohort
contains 1,735 episodes; its 19 tasks with verified random and ideal
references contribute 1,360 episodes to normalized comparisons.
Environment seeds match initial conditions across configurations;
model sampling and subsequent trajectories can differ.

\paragraph{Context usage.}
Table~\ref{tab:appendix_context_usage} reports prompt tokens per
environment decision in the matched categorical cohort. Each episode's
total prompt tokens, including forecast and summary calls, is divided
by its decision count; these ratios are averaged over seeds and then
equally over tasks. Pooled comparisons first average backbones within
each task. Token shares are ratios of these means, and retained
completeness gains are ratios of mean improvements over Stateless.
Window uses 8.72\% of Cumulative's prompt tokens and retains 78.66\%
of its completeness gain; Surprise uses 34.37\% and retains 89.69\%.
These are context-use measures, not monetary costs or cache-adjusted
billing estimates.

\begin{table}[htbp]
\centering
\small
\caption{Mean prompt tokens per decision across the 23 categorical tasks.}
\label{tab:appendix_context_usage}
\begin{tabular}{lrrrrr}
\toprule
Backbone & Stateless & Cumulative & Window & Retrieval & Surprise \\
\midrule
GPT-5.6 Luna        & 2,124 & 35,026 & 4,075 & 4,078 & 12,470 \\
DeepSeek V4.1 Flash &   446 & 34,223 & 2,551 & 2,552 & 11,146 \\
Qwen3.8 Flash      &   405 & 36,269 & 2,577 & 2,565 & 12,653 \\
\bottomrule
\end{tabular}
\end{table}

\section{Aggregation and Exploratory Inference}
\label{app:inference}

Analyses retain complete, valid episodes from seeds available under all
five configurations within each task and backbone. This yields 1,690
episodes for categorical human-policy scoring and 1,360 episodes for the
19-task normalized-performance analysis. Matching seeds controls initial
scenarios, not the subsequent action-dependent trajectories.

Performance is normalized within episode as
$(v-v_{\mathrm{random}})/(v_{\mathrm{ideal}}-v_{\mathrm{random}})$,
using the corresponding task references, and then averaged over matched
seeds. For completeness, we first average episode-level agent and
conditional-uniform log scores, then compute
completeness as $(\overline S_{\mathrm{agent}}-\overline S_{\mathrm{random}})/
(\bar S_{\mathrm{human}}-\overline S_{\mathrm{random}})$.
Thus, the reported cell score is a ratio of means, not a mean of episode
ratios. The human anchor is the equal-sequence mean of held-out human
scores. Neither score is clipped.

``Any memory'' averages the four memory configurations before subtracting
Stateless. Pooled effects average available backbones within each task,
then weight tasks equally. Confidence intervals are the 2.5th and 97.5th
percentiles of 10,000 bootstrap means obtained by resampling tasks
(random seed 20260925), retaining their configurations and backbones
together. These intervals hold the fitted scorers and human anchors
fixed and are not multiplicity-adjusted. The resulting pooled effects
are $0.140$ $[0.056,0.234]$ for performance, $0.222$
$[0.119,0.336]$ for completeness, and $0.168$ $[0.096,0.246]$
nats for $\bar S$.

Completeness effects are tested using two-sided task-level Wilcoxon
signed-rank tests after removing exact zero differences. The Holm family
contains eight comparisons: any-memory effects for three backbones,
four pooled memory-versus-Stateless effects, and the pooled any-memory
effect. Backbone results are shown below. For the separate
Cumulative--Window comparison, the main text reports the unadjusted
$p=0.273$; Holm adjustment over all six memory-pair contrasts for each
of three backbones and the pooled analysis gives $p=1.000$.

\begin{table}[htbp]
\centering
\small
\caption{Task-level any-memory completeness contrasts across 23 tasks.}
\begin{tabular}{lrrr}
\toprule
Backbone & Any-memory $\Delta c$ & Raw $p$ & Holm $p$ \\
\midrule
GPT & 0.125 & 0.030118 & 0.030118 \\
DeepSeek & 0.284 & 0.007452 & 0.014904 \\
Qwen & 0.256 & 0.000073 & 0.000512 \\
\bottomrule
\end{tabular}
\end{table}

For the 18 categorical tasks with normalized performance, Spearman
correlations are computed across configurations within each
task--backbone pair, then averaged across available backbones and tasks.
Undefined correlations from constant profiles are omitted. Task
bootstrap gives $\bar\rho=0.406$ $[0.233,0.569]$; omitting Stateless
gives $0.311$ $[0.140,0.484]$. All analyses are exploratory; individual
decisions are not treated as independent replicates.

\section{Using learned human-policy model to evaluate agent-generated trajectories}
\label{app:justify_rnn_scores}

We evaluate an agent's decisions using the probabilities assigned by a learned human-policy model along the agent's own interaction history. Below, we derive this score from the trajectory likelihood and clarify its interpretation.

For a fixed task $\xi$, consider an agent-generated trajectory $\tau^A=(o_1^A,a_1^A,\ldots,o_T^A,a_T^A)$. Let $\eta_t^A=(o_1^A,a_1^A,\ldots,o_t^A)$ denote the public interaction history available before action $a_t^A$, and let $x_{1:t}^A$ denote its structured representation. The frozen RNN processes $x_{1:t}^A$ to obtain its hidden state $h_t^A$ and assigns the observed action probability
\begin{equation}
\pi_\text{RNN}(a_t^A\mid x_{1:t}^A)
=
\pi_\text{RNN}(a_t^A\mid h_t^A).
\end{equation}
The recurrent state is updated using the agent's preceding actions and feedback. The current action $a_t^A$ enters the history only after its probability has been evaluated. Thus, the RNN scores decisions under histories generated by the agent, without requiring a matching human trajectory.

For categorical decisions and a fixed horizon $T$, the probability of the same trajectory under the RNN policy interacting with the environment factorizes as
\begin{align}
P_\text{RNN}^{\xi}(\tau^A)
&=
\underbrace{
\mu_\xi(o_1^A) \cdot
\prod_{t=1}^{T-1}
P_\xi(o_{t+1}^A\mid\eta_t^A,a_t^A)
}_{\text{environment contribution}} \cdot
\underbrace{
\prod_{t=1}^{T}
\pi_\text{RNN}(a_t^A\mid h_t^A)
}_{\text{action contribution}},
\label{eq: RNN_factorization}
\end{align}
where $\mu_\xi$ is the initial-observation distribution and $P_\xi$ is the distribution of the next observation induced by the environment dynamics and rendering function. Conditioning $P_\xi$ on the participant-visible history accommodates environments whose internal states are only partially observed.

Our evaluation uses the action contribution,
\begin{equation}
S_\text{RNN}(\tau^A)
=
\prod_{t=1}^{T}
\pi_\text{RNN}(a_t^A\mid h_t^A).
\label{eq: raw_score}
\end{equation}
This is a causal action likelihood: each action is evaluated using only information available before that decision \citep{ziebart2010modeling}. It is generally different from conditioning the entire action sequence on the entire observation sequence, which would also condition earlier decisions on future observations. The factorization does not assume independent actions, because each probability depends on the preceding interaction history.

Scoring the action contribution evaluates the agent's choices while excluding the direct likelihood of the environment's stochastic outcomes. Those outcomes still affect subsequent scores through the interaction history. The environment contribution can differ across trajectories, so $S_\text{RNN}$ should not be interpreted as the full joint trajectory probability.

To express the score per decision, we take the logarithm and divide by the number of decisions:
\begin{align}
\bar{S}_\text{RNN}(\tau^A)
&=
\frac{1}{T}\log S_\text{RNN}(\tau^A) \\
&=
\frac{1}{T}\sum_{t=1}^{T}
\log \pi_\text{RNN}(a_t^A\mid h_t^A).
\label{eq: get_per_decision_score}
\end{align}
Exponentiating gives the geometric mean of the per-decision action probabilities:
\begin{equation}
\exp \bigl(\bar{S}_\text{RNN}(\tau^A)\bigr)
=
\left[
\prod_{t=1}^{T}
\pi_\text{RNN}(a_t^A\mid h_t^A)
\right]^{1/T}.
\label{eq: interpret_geometric_mean}
\end{equation}
Thus, $\exp \big( \bar{S}_\text{RNN} \big)$ is the geometric mean of likelihoods. They produce identical rankings because exponentiation is strictly increasing. This construction preserves the multiplicative structure of sequential action likelihoods while expressing the result on a per-decision scale. For variable-length episodes, we use each episode's number of decisions in the denominator. This normalization removes the direct dependence of the product on sequence length, although differences in task difficulty and visited histories remain.

\textbf{Interpreting the score.}
The logarithmic score is strictly proper when used to evaluate a predictive distribution against responses drawn from a fixed target distribution \citep{gneiting2007strictly}. This supports its use for training and validating the RNN on human trajectories. Evaluating agent-generated trajectories under a fixed RNN, however, measures their compatibility with the learned human policy rather than guaranteeing distributional matching.

To make this distinction explicit, let $\pi_\text{Agent}(a_t\mid\eta_t)$ denote the agent's induced policy over executed actions, including the effects of its LLM and harness. For a fixed horizon $T$, define its average conditional action entropy as
\begin{equation}
\mathcal{H}_\text{Agent}
=
-\frac{1}{T}
\mathbb{E}_{\tau\sim P_\text{Agent}^\xi}
\left[
\sum_{t=1}^{T}
\log\pi_\text{Agent}(a_t\mid\eta_t)
\right].
\end{equation}
When the agent and RNN policies interact with the same environment and initial-observation distribution,
\begin{equation}
\mathbb{E}_{\tau\sim P_\text{Agent}^\xi}
\left[\bar{S}_\text{RNN}(\tau)\right]
=
-\mathcal{H}_\text{Agent}
-
\frac{1}{T}
D_\mathrm{KL}
\left(
P_\text{Agent}^\xi \Vert P_\text{RNN}^\xi
\right).
\label{eq: score_action entropy_KL}
\end{equation}
This identity follows by expanding the trajectory KL divergence: the shared environment factors in Equation~\ref{eq: RNN_factorization} cancel, leaving the sum of action log-probability ratios.

A higher expected score can therefore reflect both a smaller divergence from the reference policy and lower action entropy. In particular, an agent that consistently chooses modal human responses can score higher than one that reproduces human variability. We consequently interpret the score as behavioral compatibility with a learned human policy. Computing it requires only the agent's sampled actions and the RNN's probabilities, without access to the agent's own action probabilities. Obtaining the latter can be difficult even when token-level probabilities are available, because multiple response strings and reasoning traces can lead to the same parsed action. Computing an action's probability generally requires marginalizing over these alternatives; the probability of an answer token conditioned on a single reasoning trace need not equal the agent's marginal action probability. Moreover, interpreting the mean log likelihoods as human-likeness also depends on the RNN's predictive validity under agent-generated histories. Performance on held-out human trajectories supports the use of the RNN as a reference, but does not by itself establish reliable extrapolation to rarely or never observed histories.

\section{Exploratory statistical analysis of task performance}
\label{app:performance_statistics}
We analyzed the existing scalar performance endpoints of all 19 tasks with a
defined improvement direction. The comparison family comprised 228 contrasts:
four memory configurations versus Stateless for each of three backbones and
19 tasks. Within each task and backbone, we retained seeds with complete,
finite outcomes under all five configurations and required at least three
matched seeds, yielding 224 estimable contrasts. Each episode contributed
one summary outcome; sequential decisions were not treated as independent
replicates. Positive differences indicate improved performance, with signs
reversed for lower-is-better endpoints.

We report mean paired differences, marginal unadjusted 95\% paired-$t$
confidence intervals, and both two-sided paired-$t$ and exact sign-flip
tests of the mean difference. Holm adjustment was applied separately to
each test procedure across the same 228 planned contrasts. Undefined tests
were flagged and assigned $p=1$ solely for adjustment bookkeeping.
Paired-$t$ inference assumes independent, approximately normal seed
differences; sign-flip inference requires the corresponding null symmetry
or within-pair exchangeability. With five pairs, the usual two-sided
exact test cannot yield $p<0.0625$. Small sample sizes limit assessment of
these assumptions. The confidence intervals are not multiplicity-adjusted.

These analyses were specified after inspection of descriptive outcomes and
are exploratory. Adjustment across the full comparison family does not
remove prior task or reporting selection. Matching protocol seeds does not
imply identical realized trajectories in adaptive or action-dependent tasks.
Uncertainty is conditional on the evaluated tasks, backbones, protocol,
frozen endpoints, and retained complete runs; neither test is treated as
independent confirmation or selected according to its significance.

\section{The best memory configuration depends on the task}

\begin{figure}[h!]
    \centering
    \includegraphics[width=\linewidth]{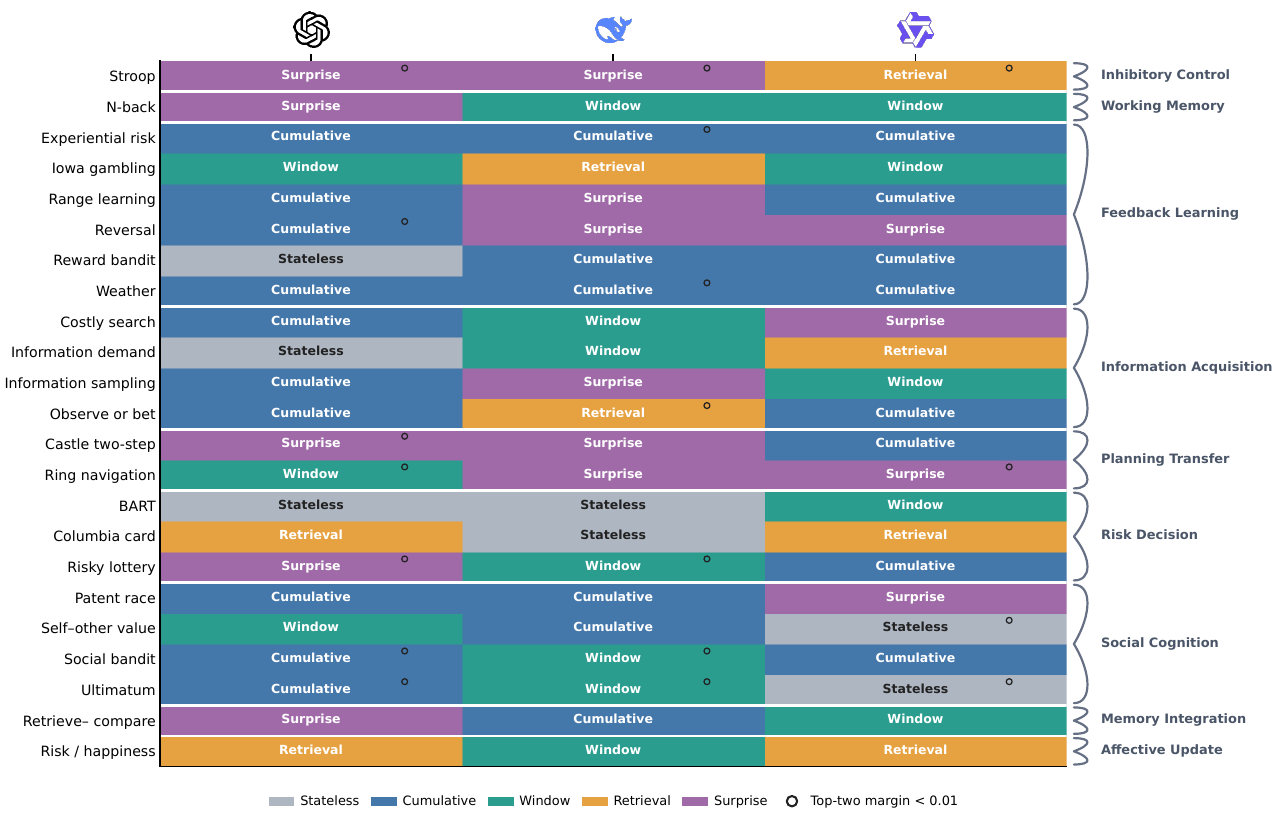}
    \caption{
    Task-level variation in human-policy likelihood using the latest audited
    result table.
    Each cell identifies the memory configuration with the highest $\bar{S}$ for a task and backbone. Open circles mark cells for
    which the difference between the two highest means is below
    $0.01$ nats per scored action. Colors encode descriptive mean winners and do not imply
    statistically significant pairwise superiority.
    }
    \label{fig:behavioral_variation}
\end{figure}

No single memory mechanism is the winner for every task and
backbone (Figure~\ref{fig:behavioral_variation}). Although Cumulative
memory performs best on average, the configuration that most closely
matches the learned human policy in raw likelihood is task- and
backbone-dependent; many
near-ties further caution against interpreting small mean differences as
reliable superiority.

\begin{figure*}[t]
    \centering
    \includegraphics[width=\textwidth]{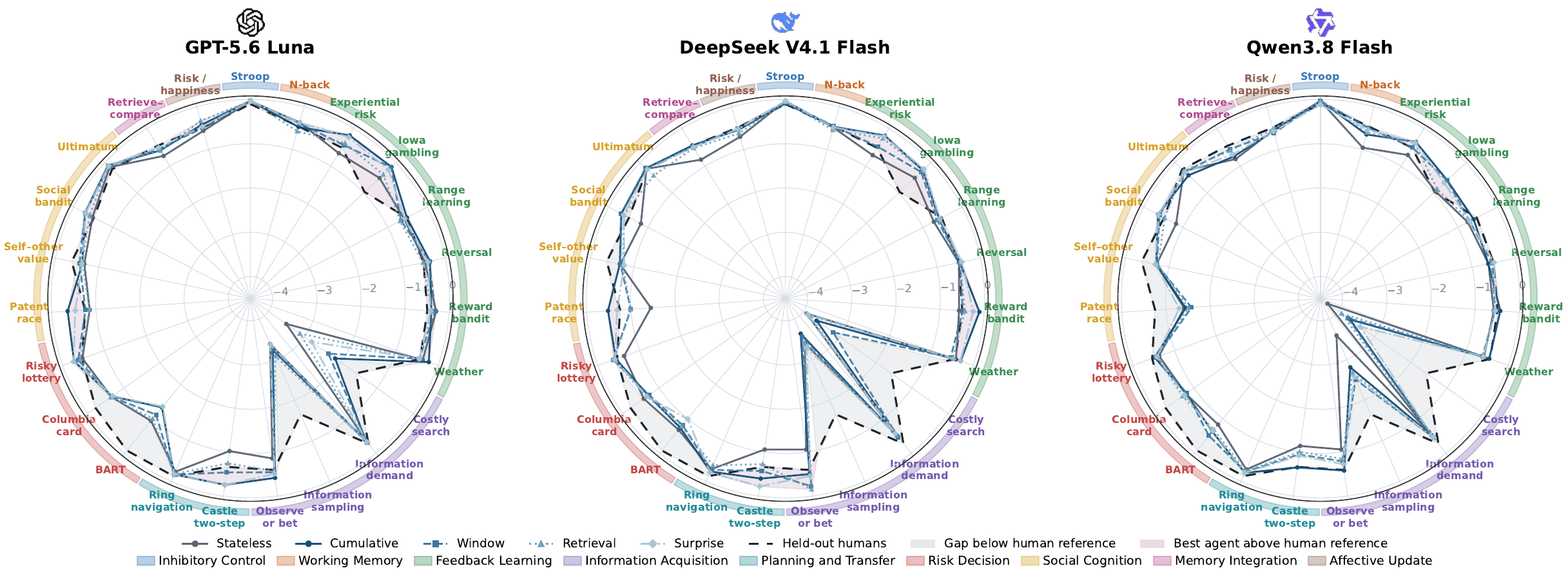}
    \caption{
    Human-policy fit $\bar S$ across the 23 categorical tasks. Each axis
    reports $\bar S$ in nats per scored action, so values closer to zero
    indicate that the learned human policy assigns greater likelihood to the
    agent's choices. The dashed line is the score of held-out human sequences
    and is an empirical reference rather than an upper bound. Outer colors
    denote psychological categories; lines denote memory configurations.
    Gray shading marks the gap between the best displayed agent configuration
    and the human reference; purple shading marks tasks on which the best
    agent score exceeds that reference.
    Because raw $\bar S$ depends on task-specific action spaces and predictability,
    radial distances and polygon areas should not be compared as aggregate
    scores across tasks.
    }
    \label{fig:human_policy_likelihood_radar}
\end{figure*}

\providecommand{\InteractionAppendixPath}{appendix_short}
\section{Replay and Autonomous Behavior}
\paragraph{Teacher-forced replay}
\label{app:replay}
We paired 160 episodes across four tasks, two backbones, four memory
configurations, and five seeds. For each task and seed, all conditions
shared one trajectory sampled from the frozen human-policy RNN.
Replay executed its actions and recorded them in memory; agent proposals
were scored without changing the path. Closed-loop interaction executed
the agent's own actions. Table~\ref{tab:replay} reports $\bar S$ under each
mode's realized history. Memory gain averaged Cumulative, Window, and
Retrieval relative to Stateless. Resampling tasks and then paired seeds,
keeping both backbones together, gave a replay-minus-closed-loop gain
of $0.124$ nats (10,000 bootstrap draws; 95\% interval $[-0.117,0.519]$). For weather classification, the shared trajectories were sampled from an earlier version of the human-policy RNN; all scores use the current RNN.

\begin{table}[htbp]
\centering
\scriptsize
\setlength{\tabcolsep}{3pt}
\caption{Mean log score $S$ (nats per decision), averaged over five matched seeds. Replay proposals are evaluated on forced histories; closed-loop actions on their own histories.}
\begin{tabular}{llrrrrrrrr}
\toprule
& & \multicolumn{4}{c}{Replay $S$} & \multicolumn{4}{c}{Closed-loop $S$} \\
\cmidrule(lr){3-6}\cmidrule(lr){7-10}
Task & Backbone & Stateless & Cumul. & Window & Retr. & Stateless & Cumul. & Window & Retr. \\
\midrule
Iowa & DeepSeek & $-0.738$ & $-0.741$ & $-0.736$ & $-0.756$ & $-0.501$ & $-0.252$ & $-0.305$ & $-0.247$ \\
Iowa & Qwen & $-1.209$ & $-1.081$ & $-0.809$ & $-0.975$ & $-0.945$ & $-0.581$ & $-0.586$ & $-0.886$ \\
Weather & DeepSeek & $-0.753$ & $-0.718$ & $-0.708$ & $-0.707$ & $-0.409$ & $-0.296$ & $-0.536$ & $-0.482$ \\
Weather & Qwen & $-0.712$ & $-0.729$ & $-0.704$ & $-0.722$ & $-0.575$ & $-0.451$ & $-0.613$ & $-0.589$ \\
Castle & DeepSeek & $-1.902$ & $-0.927$ & $-1.058$ & $-0.930$ & $-1.059$ & $-0.396$ & $-0.570$ & $-0.725$ \\
Castle & Qwen & $-1.900$ & $-0.977$ & $-1.191$ & $-1.021$ & $-1.140$ & $-0.651$ & $-0.947$ & $-1.004$ \\
Social bandit & DeepSeek & $-1.015$ & $-0.446$ & $-0.423$ & $-0.544$ & $-0.819$ & $-0.322$ & $-0.319$ & $-0.393$ \\
Social bandit & Qwen & $-0.974$ & $-0.425$ & $-0.434$ & $-0.507$ & $-0.824$ & $-0.364$ & $-0.403$ & $-0.535$ \\
\bottomrule
\end{tabular}
\label{tab:replay}
\end{table}

\begin{figure}[htbp]
\centering
\includegraphics[width=0.94\linewidth]{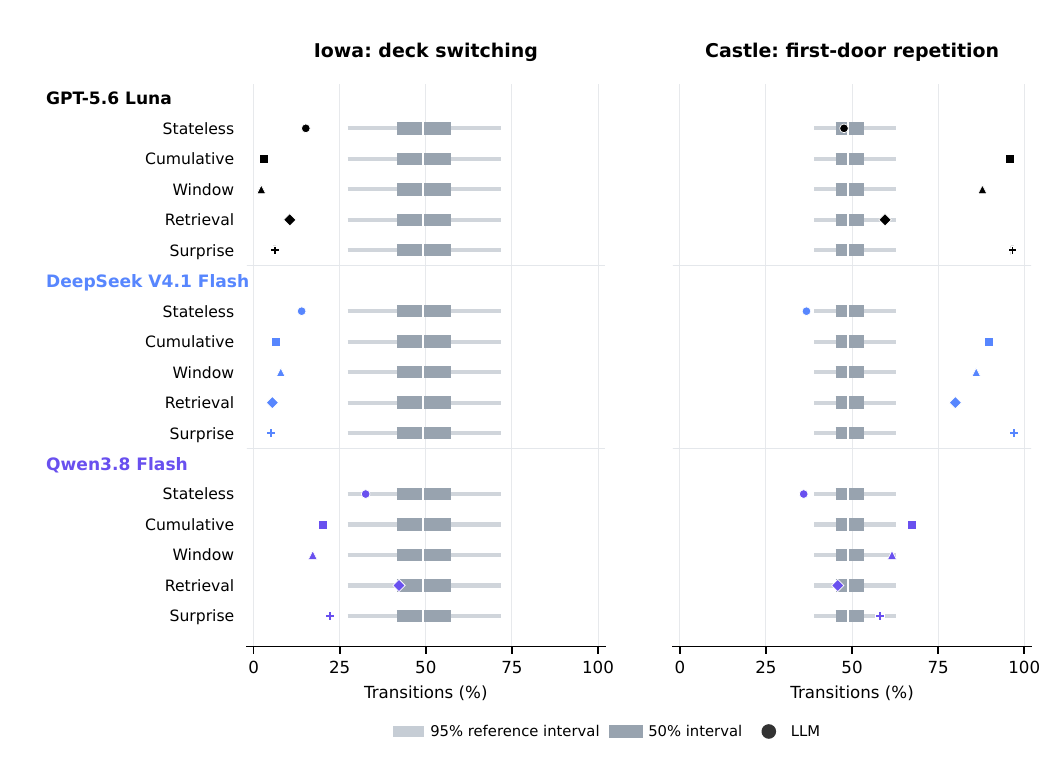}
\caption{Selected autonomous behavioral checks. Each marker summarizes
five LLM episodes; gray bands show matched RNN predictive intervals,
with a white tick at the median. Iowa measures switching between decks
on consecutive trials. Castle measures repetition of the first-door
choice across consecutive training rounds. Colors identify backbones;
rows identify memory configurations.}
\label{fig:autonomous_examples}
\end{figure}

\paragraph{Autonomous behavioral checks.}
\label{app:autonomous_behavior}
We compared 623 included episodes across nine branching tasks with
complete autonomous RNN trajectories. For each task--backbone--memory
group, 10,000 reference experiments matched its initial scenarios and
episode count, drawing one full trajectory per scenario. Ordinary
features used equal-episode means; conditional rates pooled event and
opportunity counts. Undefined statistics remained unavailable.
Figure~\ref{fig:autonomous_examples} shows the switching and repetition
features discussed in the main text, retaining all configurations.
Bands are predictive intervals under the frozen RNN, rather than
confidence intervals for LLM effects or human population means.
Complete feature results are provided in the supplementary data.

\section{Results on Continuous-Response Tasks}
\label{app:continuous_results}

For Momentary mood and Sparse estimation, we report mean absolute
deviation (MAD) between agent responses and the frozen human-policy
model's predictions, conditioned on the agent's own interaction history.
Lower MAD indicates closer agreement. Mood scores use the $[-1,1]$
scale; Sparse-estimation MAD averages deviations for the post-training
and post-test estimates in original slider points (ranges $0$--$300$
or $0$--$500$).
Table~\ref{tab:continuous_agent_results} averages episode-level MAD
over completed episodes. Momentary mood has 25 completed episodes and
50 protocol-screened terminations; Sparse estimation has 69 completed
episodes and six incomplete episodes. Cell counts are reported because
the available seeds differ across configurations.

\begin{table}[htbp]
\centering
\small
\setlength{\tabcolsep}{3pt}
\caption{Agent--human-policy MAD on the two continuous-only tasks
(lower is better). Values are means across completed episodes;
parentheses give episode counts. Dashes indicate no completed
episodes. Mood and estimation retain their respective native units.}
\label{tab:continuous_agent_results}
\begin{tabular}{llccccc}
\toprule
Task & Backbone & Stateless & Cumulative & Window & Retrieval & Surprise \\
\midrule
Momentary mood & GPT & -- & -- & -- & -- & -- \\
 & DeepSeek & 0.190 (1) & 0.073 (3) & 0.070 (3) & -- & 0.093 (4) \\
 & Qwen & 0.251 (1) & 0.184 (4) & 0.185 (3) & 0.248 (2) & 0.196 (4) \\
\addlinespace
Sparse estimation & GPT & -- & 4.3 (5) & 11.5 (5) & 11.4 (5) & 7.5 (5) \\
 & DeepSeek & 50.0 (5) & 6.3 (5) & 67.4 (5) & 21.9 (5) & 5.8 (5) \\
 & Qwen & 69.2 (4) & 7.5 (5) & 11.5 (5) & 19.1 (5) & 6.5 (5) \\
\bottomrule
\end{tabular}
\end{table}

\section{Prompt examples}

The following abridged example shows the participant-facing text for
the first Stroop decision in a saved Stateless episode (seed 101).
Line breaks are adjusted for readability. The task instructions and
current observation are supplied separately from any memory selected
by the configuration; no previous interaction exists at this first step.

\begin{samepage}
\textbf{Task instructions (excerpt).}
\begin{quote}
This is a three-colour Stroop response task. On every trial, report
the PRINTED COLOUR annotation, not the word's meaning. No response-time
measure is collected.

Actions: 0 = red, 1 = green, 2 = blue. There are 84 trials.
\end{quote}
\end{samepage}

\textbf{Current observation.}
\begin{verbatim}
[Trial 1 of 84]

Word: RED  (printed in: green)

What is the printed colour? (0=red, 1=green, 2=blue)
Reply with exactly one final line: `Action: N`.
\end{verbatim}

\textbf{Response-format contract (excerpt).}
The runtime appends an overriding instruction: ``Return only one JSON
object with exactly one key, action, whose value is an available integer
action. Do not explain the choice or repeat any observation or past
record.'' This JSON contract supersedes the legacy response-style
sentence visible in the observation above. Here the permitted values
are $\{0,1,2\}$; the response contains no additional fields or explanation.

\textbf{Valid response format (illustrative, not a quoted LLM output).}
\begin{verbatim}
{"action": 1}
\end{verbatim}
The harness parses and validates the numeric index before executing it.

\section{Action entropy and the KL divergence from the human policy}
\label{app:kl}

Equation~\ref{eq: score_action entropy_KL} splits the expected likelihood score into the agent's action entropy and its KL divergence from the human-policy RNN. This section describes how we estimated the action entropy based on the same agent trajectories.

\textbf{Definitions.}
Consider decision $t$ of an agent episode with public history $\eta_t$ and valid actions $\mathcal{A}_t$. Let $\pi_t(a)=\pi_\text{Agent}(a\mid\eta_t)$ denote the agent's distribution over executed actions and $p_t(a)=\pi_\text{RNN}(a\mid h_t)$ the RNN's distribution for the same history. The entropy of the agent's choice and its KL divergence from the RNN at this decision are
\begin{equation}
H_t=-\sum_{a\in\mathcal{A}_t}\pi_t(a)\log\pi_t(a),
\qquad
D_t=\sum_{a\in\mathcal{A}_t}\pi_t(a)\log\frac{\pi_t(a)}{p_t(a)}.
\label{eq:kl_per_decision}
\end{equation}
The expected score of the action sampled at this decision is the negative cross-entropy between the two distributions,
\begin{equation}
\mathbb{E}_{a_t\sim\pi_t}\bigl[\log p_t(a_t)\bigr]
=\sum_{a\in\mathcal{A}_t}\pi_t(a)\log p_t(a)
=-H_t-D_t .
\label{eq:kl_cross_entropy}
\end{equation}
Averaging over the $T$ decisions of a trajectory and taking the expectation over trajectories gives Equation~\ref{eq: score_action entropy_KL}: $\mathcal{H}_\text{Agent}=\mathbb{E}\bigl[\frac{1}{T}\sum_t H_t\bigr]$ and, by the chain rule for the KL divergence between trajectory distributions, $\frac{1}{T}D_\mathrm{KL}\bigl(P_\text{Agent}^\xi\Vert P_\text{RNN}^\xi\bigr)=\mathbb{E}\bigl[\frac{1}{T}\sum_t D_t\bigr]$. We report the per-decision KL divergence along the agent's own trajectory,
\begin{equation}
D_{\mathrm{KL}}=\frac{1}{T}\sum_{t=1}^{T}D_t ,
\label{eq:kl_definition}
\end{equation}
in nats per decision. It is zero if and only if the agent chooses with the RNN's probabilities at every decision it reaches. A deterministic policy has $H_t=0$, so that $D_{\mathrm{KL}}=-\bar S$, and a policy that samples from the RNN has $D_{\mathrm{KL}}=0$.

\textbf{Action probabilities.}
The original runs stored each sampled action but not the distribution it was drawn from. For every scored decision we re-submitted the saved request, with identical context and decoding settings, and requested token log probabilities. All runs used non-thinking mode with a constrained output in which the action is a single integer token, so no marginalization over response strings or reasoning traces is required. Both distributions are compared over the valid actions $\mathcal{A}_t$ only: the agent's probability of action $a$ is the probability $q_t(a)$ of the tokens that denote $a$ at the action position (for example, the digit with or without a leading space), renormalized over the valid actions:
\begin{equation}
\hat\pi_t(a)=\frac{q_t(a)}{\sum_{a'\in\mathcal{A}_t}q_t(a')},\qquad a\in\mathcal{A}_t .
\label{eq:kl_recovered}
\end{equation}
The APIs return probabilities only for the most likely tokens at each position (at most five for Qwen), so we requested the five most likely tokens for both backbones. Every analyzed decision has at most five valid actions, a valid action that is not returned has a lower probability than the fifth returned token (median $2\times10^{-7}$ for Qwen and $2\times10^{-6}$ for DeepSeek), and the saved action was among the returned tokens in all but 0.01\% of decisions. This was possible for DeepSeek V4.1 Flash and Qwen3.8 Flash on 20 of the 23 categorical tasks (980 matched episodes, 180,244 decisions); information sampling, risky information acquisition, and retrieve--compare use multi-token action labels and are excluded. GPT-5.6 Luna is a closed-source model that we accessed through an interface returning no log probabilities. The DeepSeek probabilities come from a dated snapshot of the same model version served by another provider.

\textbf{Estimators.}
Let $\hat H_t$ be the entropy of $\hat\pi_t$. For an episode with scored decisions $t=1,\dots,T$ and saved actions $a_t$, we compute
\begin{align}
\hat{\mathcal{H}}_\text{Agent}&=\frac{1}{T}\sum_{t=1}^{T}\hat H_t, \label{eq:kl_entropy_estimate}\\
\hat D_{\mathrm{KL}}&=-\bar S-\hat{\mathcal{H}}_\text{Agent}
=\frac{1}{T}\sum_{t=1}^{T}\bigl[-\log p_t(a_t)-\hat H_t\bigr], \label{eq:kl_plugin}\\
\tilde D_{\mathrm{KL}}&=\frac{1}{|\mathcal{T}'|}\sum_{t\in\mathcal{T}'}\log\hat\pi_t(a_t)-\bar S, \label{eq:kl_sample}
\end{align}
where $\mathcal{T}'$ contains the decisions whose saved action is among the returned tokens. The plug-in estimate $\hat D_{\mathrm{KL}}$ takes the entropy from the full recovered distribution and uses the sampled action only through $\bar S$; the sample-based estimate $\tilde D_{\mathrm{KL}}$ uses the sampled action in both terms. By Equation~\ref{eq:kl_cross_entropy}, both are unbiased for $D_{\mathrm{KL}}$ when $\hat\pi_t$ is the distribution that generated $a_t$. Otherwise $\tilde D_{\mathrm{KL}}$ is biased downward by the divergence of the generating distribution from $\hat\pi_t$, and $\hat D_{\mathrm{KL}}$ is biased by the difference between the two entropies. Their difference, $\hat D_{\mathrm{KL}}-\tilde D_{\mathrm{KL}}=-\hat{\mathcal{H}}_\text{Agent}-\frac{1}{|\mathcal{T}'|}\sum_{t\in\mathcal{T}'}\log\hat\pi_t(a_t)$, has expectation zero when the saved actions are draws from $\hat\pi_t$ and serves as a consistency check: pooled over configurations, it was below 0.05 nats in 18 of 20 tasks for Qwen and in 16 of 20 for DeepSeek. Benchmark-level changes differ by up to 0.03 nats between the two estimates, which is why we treat $D_{\mathrm{KL}}$ as approximate and report both. Unless noted otherwise, $D_{\mathrm{KL}}$ in tables denotes the plug-in estimate.

\textbf{Reference points and aggregation.}
For held-out participants, whose own action probabilities are unobservable, the analogous quantity is the RNN's cross-entropy on human histories minus its own entropy there, $-\bar S_\text{human}-\frac{1}{T}\sum_t H(p_t)$, which is zero for a calibrated model (the calibration gap in Table~\ref{tab:kl_reference_policies}). Estimates are computed per episode, averaged over matched seeds within task and configuration, and then averaged with equal weight over the 20 tasks. Changes relative to Stateless and contrasts between configurations are paired by task; intervals are 95\% task-bootstrap intervals (10,000 draws), and tests are two-sided Wilcoxon signed-rank tests on the 20 task means with Holm correction over the eight backbone-by-configuration contrasts.

\begin{figure}[htbp]
\centering
\IfFileExists{figures/fig_kl_decomposition.pdf}{\includegraphics[width=\linewidth]{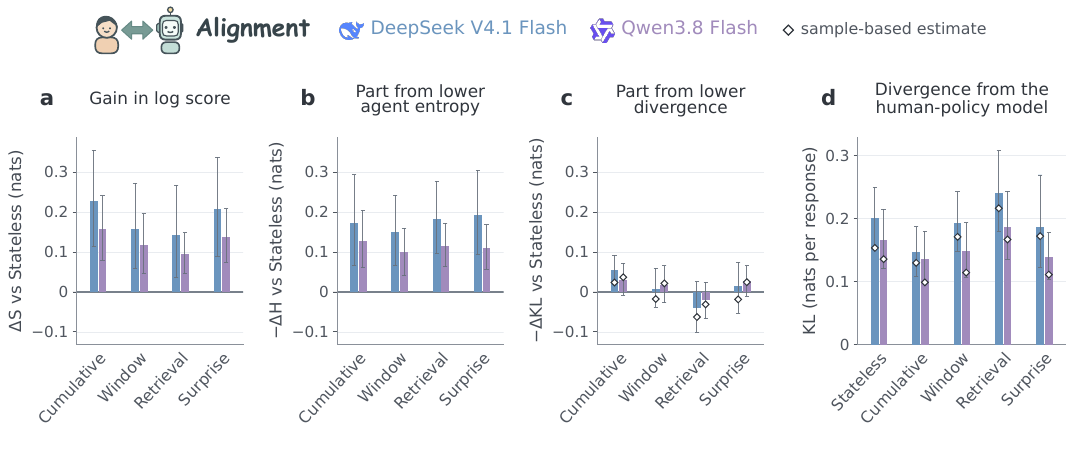}}{\fbox{\parbox{0.9\linewidth}{\centering figures/fig\_kl\_decomposition.pdf (to be uploaded)}}}
\caption{Decomposition of the change in $\bar S$ relative to Stateless into lower action entropy and lower divergence from the human-policy RNN, and $D_{\mathrm{KL}}$ by configuration, for DeepSeek and Qwen (20 tasks). Error bars denote 95\% task-bootstrap intervals; diamonds mark the sample-based estimate.}
\label{fig:kl_decomposition}
\end{figure}

\textbf{Results.}
Table~\ref{tab:kl_decomposition} and Figure~\ref{fig:kl_decomposition} give the decomposition. Memory lowered action entropy in every configuration, and this accounts for 76--95\% of the gain in $\bar S$ under Cumulative, Window, and Surprise; under Retrieval, divergence increased. No change in $D_{\mathrm{KL}}$ relative to Stateless is significant after Holm correction (smallest adjusted $p=0.08$, Cumulative for DeepSeek). Differences between memory designs are more reliable: $D_{\mathrm{KL}}$ under Retrieval exceeds that under Cumulative by 0.094 (CI $[0.051,0.136]$) for DeepSeek and 0.051 (CI $[0.013,0.088]$) for Qwen, and exceeds that under Window, which uses the same prompt budget, by 0.048 (CI $[0.014,0.082]$) and 0.038 (CI $[0.006,0.074]$). $D_{\mathrm{KL}}$ decreased relative to Stateless in 91 of 160 task-level comparisons, whereas $\bar S$ increased in 117.

\textbf{Simple reference policies.}
Table~\ref{tab:kl_reference_policies} scores simple policies with the same RNNs and seeds. Under $\bar S$, greedily choosing the RNN's most probable action and repeating the best constant action both score above held-out humans. Under $D_{\mathrm{KL}}$, held-out humans are at approximately zero, the greedy policy at 0.13, and constant and uniformly random policies at about 0.30; the LLM agents lie between 0.14 and 0.24. Table~\ref{tab:kl_nback} shows the N-back example of Section~\ref{sec:performance_alignment}: the always ``non-match'' policy exceeds held-out humans in $\bar S$ but has $D_{\mathrm{KL}}=0.249$, above every memory configuration of both backbones.

\begin{table}[htbp]
\centering
\scriptsize
\setlength{\tabcolsep}{3pt}
\caption{Decomposition of $\bar S$ into action entropy and divergence (Equation~\ref{eq: score_action entropy_KL}): equal-task means over 20 tasks, in nats per decision. $\Delta$ denotes the paired change from Stateless with its 95\% task-bootstrap interval; ``sample'' is the sample-based estimate $\tilde D_{\mathrm{KL}}$.}
\label{tab:kl_decomposition}
\resizebox{\linewidth}{!}{%
\begin{tabular}{llrrrrrrrr}
\toprule
Backbone & Configuration & $\bar S$ & $\mathcal{H}_\text{Agent}$ & $D_{\mathrm{KL}}$ & $D_{\mathrm{KL}}$ (sample) & $\Delta\bar S$ & $\Delta\mathcal{H}_\text{Agent}$ & $\Delta D_{\mathrm{KL}}$ & $\Delta D_{\mathrm{KL}}$ (sample) \\
\midrule
DeepSeek & Stateless & $-$0.637 & 0.436 & 0.200 & 0.153 & -- & -- & -- & -- \\
& Cumulative & $-$0.409 & 0.263 & 0.146 & 0.130 & +0.228 [+0.113, +0.355] & $-$0.174 & $-$0.054 [$-$0.092, $-$0.019] & $-$0.024 [$-$0.070, +0.021] \\
& Window & $-$0.478 & 0.286 & 0.192 & 0.171 & +0.159 [+0.058, +0.272] & $-$0.150 & $-$0.008 [$-$0.060, +0.039] & +0.018 [$-$0.037, +0.071] \\
& Retrieval & $-$0.493 & 0.253 & 0.241 & 0.216 & +0.143 [+0.036, +0.267] & $-$0.183 & +0.040 [$-$0.025, +0.100] & +0.063 [+0.001, +0.126] \\
& Surprise & $-$0.430 & 0.244 & 0.186 & 0.172 & +0.207 [+0.088, +0.337] & $-$0.192 & $-$0.014 [$-$0.074, +0.053] & +0.019 [$-$0.048, +0.098] \\
Qwen & Stateless & $-$0.713 & 0.547 & 0.166 & 0.136 & -- & -- & -- & -- \\
& Cumulative & $-$0.555 & 0.420 & 0.136 & 0.099 & +0.158 [+0.080, +0.241] & $-$0.127 & $-$0.031 [$-$0.072, +0.009] & $-$0.037 [$-$0.070, $-$0.005] \\
& Window & $-$0.596 & 0.448 & 0.148 & 0.114 & +0.117 [+0.046, +0.195] & $-$0.099 & $-$0.018 [$-$0.066, +0.027] & $-$0.021 [$-$0.055, +0.010] \\
& Retrieval & $-$0.619 & 0.432 & 0.186 & 0.167 & +0.095 [+0.045, +0.148] & $-$0.115 & +0.020 [$-$0.024, +0.067] & +0.031 [$-$0.008, +0.074] \\
& Surprise & $-$0.576 & 0.437 & 0.139 & 0.111 & +0.137 [+0.074, +0.208] & $-$0.110 & $-$0.028 [$-$0.066, +0.011] & $-$0.024 [$-$0.055, +0.007] \\
\bottomrule
\end{tabular}%
}
\end{table}

\begin{table}[htbp]
\centering
\small
\setlength{\tabcolsep}{3pt}
\caption{Humans, simple policies, and LLM agents scored by the same human-policy RNNs (closed loop, same seeds; equal-task means over 20 tasks). $D_{\mathrm{KL}}$ is given as plug-in / sample-based estimate. The last column counts tasks in which the policy's $\bar S$ is at least that of held-out humans.}
\label{tab:kl_reference_policies}
\begin{tabular}{lrrr}
\toprule
Policy & $\bar S$ & $D_{\mathrm{KL}}$ & Tasks with $\bar S\geq$ humans \\
\midrule
Held-out humans & $-$0.438 & $-$0.002 (calibration gap) & -- \\
Human-policy RNN, sampled & $-$0.452 & 0 (by definition) & 7/20 \\
Human-policy RNN, greedy & $-$0.130 & 0.130 & 20/20 \\
Best constant action & $-$0.297 & 0.297 & 17/20 \\
Uniform random & $-$1.151 & 0.302 & 0/20 \\
DeepSeek V4.1 Flash, Stateless & $-$0.637 & 0.200 / 0.153 & 5/20 \\
DeepSeek V4.1 Flash, Cumulative & $-$0.409 & 0.146 / 0.130 & 12/20 \\
DeepSeek V4.1 Flash, Window & $-$0.478 & 0.192 / 0.171 & 11/20 \\
DeepSeek V4.1 Flash, Retrieval & $-$0.493 & 0.241 / 0.216 & 8/20 \\
DeepSeek V4.1 Flash, Surprise & $-$0.430 & 0.186 / 0.172 & 11/20 \\
Qwen3.8 Flash, Stateless & $-$0.713 & 0.166 / 0.136 & 2/20 \\
Qwen3.8 Flash, Cumulative & $-$0.555 & 0.136 / 0.099 & 7/20 \\
Qwen3.8 Flash, Window & $-$0.596 & 0.148 / 0.114 & 4/20 \\
Qwen3.8 Flash, Retrieval & $-$0.619 & 0.186 / 0.167 & 3/20 \\
Qwen3.8 Flash, Surprise & $-$0.576 & 0.139 / 0.111 & 4/20 \\
\bottomrule
\end{tabular}
\end{table}

\begin{table}[htbp]
\centering
\small
\setlength{\tabcolsep}{3pt}
\caption{N-back under both measures (five seeds). $D_{\mathrm{KL}}$ is given as plug-in / sample-based estimate.}
\label{tab:kl_nback}
\begin{tabular}{lrr}
\toprule
Policy & $\bar S$ & $D_{\mathrm{KL}}$ \\
\midrule
Held-out humans & $-$0.458 & $-$0.008 (calibration gap) \\
Always non-match & $-$0.249 & 0.249 \\
Human-policy RNN, greedy & $-$0.248 & 0.248 \\
Uniform random & $-$0.950 & 0.259 \\
DeepSeek V4.1 Flash, Stateless & $-$0.486 & 0.265 / 0.230 \\
DeepSeek V4.1 Flash, Cumulative & $-$0.467 & 0.042 / 0.069 \\
DeepSeek V4.1 Flash, Window & $-$0.451 & 0.129 / 0.106 \\
DeepSeek V4.1 Flash, Retrieval & $-$0.539 & 0.197 / 0.149 \\
DeepSeek V4.1 Flash, Surprise & $-$0.481 & 0.083 / 0.085 \\
Qwen3.8 Flash, Stateless & $-$0.957 & 0.310 / 0.278 \\
Qwen3.8 Flash, Cumulative & $-$0.647 & 0.157 / 0.112 \\
Qwen3.8 Flash, Window & $-$0.493 & 0.096 / 0.079 \\
Qwen3.8 Flash, Retrieval & $-$0.599 & 0.087 / 0.107 \\
Qwen3.8 Flash, Surprise & $-$0.536 & 0.101 / 0.096 \\
GPT-5.6 Luna, Stateless & $-$0.365 & -- \\
GPT-5.6 Luna, Cumulative & $-$0.472 & -- \\
\bottomrule
\end{tabular}
\end{table}

\section{Additional validation of the human-policy RNNs}
\label{app:reference_validation}

\begin{figure}[htbp]
\centering
\IfFileExists{figures/fig_reference_behavior_16tasks.pdf}{\includegraphics[width=\linewidth]{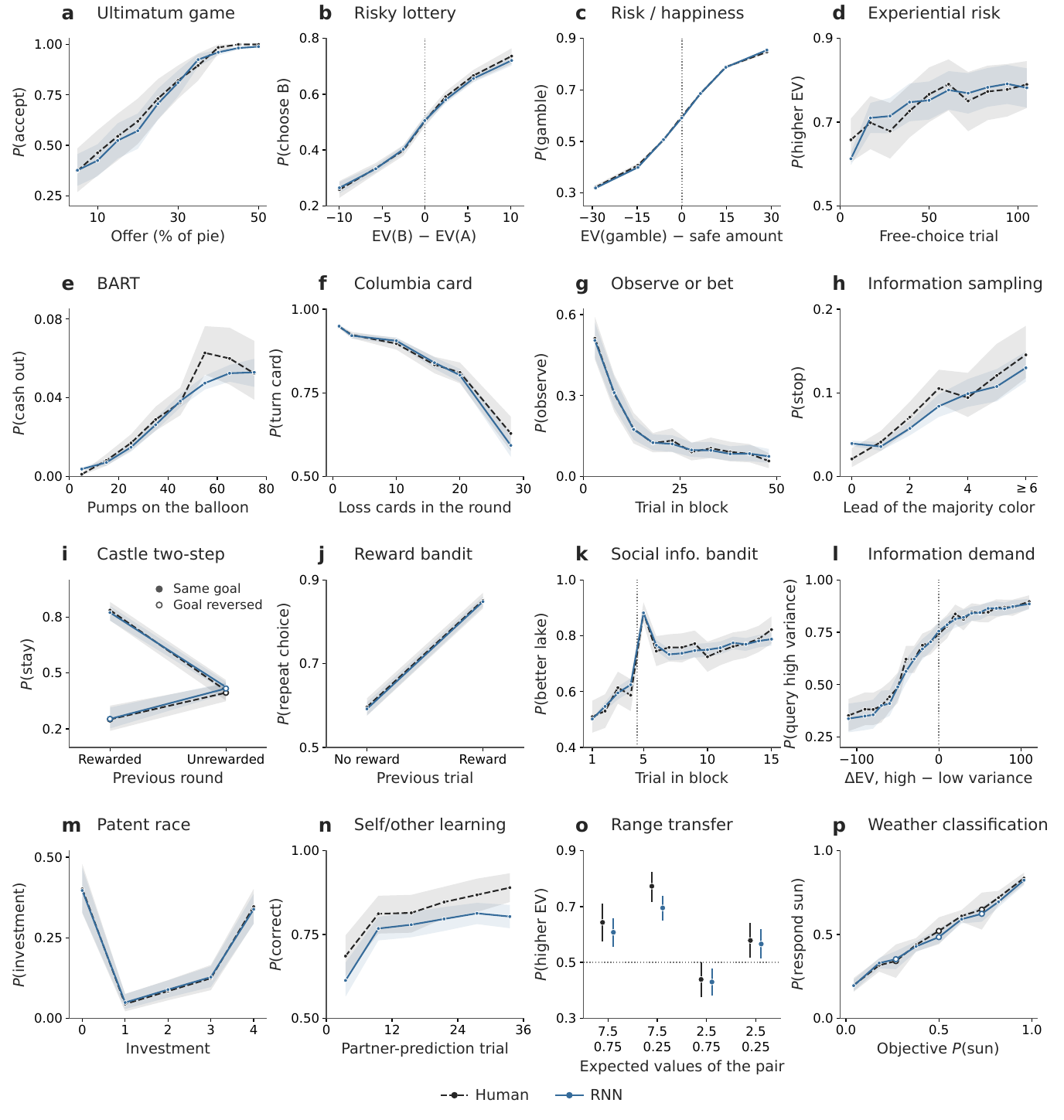}}{\fbox{\parbox{0.9\linewidth}{\centering figures/fig\_reference\_behavior\_16tasks.pdf (to be uploaded)}}}
\caption{Human responses and RNN probabilities on held-out participants for 16 tasks, in the format of Figure~\ref{fig:rnn_results}b--d. Shading and bars denote pointwise 95\% participant-cluster bootstrap intervals.}
\label{fig:reference_behavior}
\end{figure}

\textbf{Behavioral signatures on held-out participants.}
Figure~\ref{fig:reference_behavior} compares, for 16 tasks, the response proportions of held-out participants with the RNN probabilities for the same events, conditioned on each participant's own history as in Figure~\ref{fig:rnn_results}b--d. In 14 tasks the largest gap between the two curves is at most 0.06. The exceptions are one transfer pair in range learning (0.08) and self/other learning, where the RNN lies 0.04--0.09 below human accuracy. The patent-race panel shows a marginal distribution.

\textbf{Autonomous rollouts and shared histories.}
When the RNNs act on their own, their rollouts reproduce held-out human behavior on all 14 trajectory-level features that can be computed for both (Table~\ref{tab:rollout_features}). Table~\ref{tab:shared_histories} scores every policy on the same RNN-sampled histories of the replay experiment (Appendix~\ref{app:replay}), which removes any effect of the histories that an agent creates for itself: agents with memory ($-0.71$ to $-0.81$) are close to a sample from the RNN ($-0.69$) and well above constant ($-1.28$) and random ($-1.46$) policies.

\begin{table}[htbp]
\centering
\scriptsize
\setlength{\tabcolsep}{3pt}
\caption{Held-out humans and autonomous rollouts of the human-policy RNNs on trajectory-level features. The human mean lies inside the rollout interval for all 14 features. The last column gives the range over the 15 backbone-by-configuration cells of the LLM agents.}
\label{tab:rollout_features}
\resizebox{\linewidth}{!}{%
\begin{tabular}{llrrr}
\toprule
Task & Feature & Held-out humans & RNN rollouts, median [95\% interval] & LLM agents \\
\midrule
BART & Attempted pumps per balloon & 36.70 & 37.43 [30.67, 43.20] & 1.48 to 5.29 \\
BART & Adjusted pumps (cash only) & 40.24 & 39.71 [31.37, 47.78] & 1.50 to 5.44 \\
BART & Pop rate & 0.335 & 0.340 [0.253, 0.420] & 0.027 to 0.040 \\
BART & Late-minus-early pumping & 1.675 & 2.980 [$-$5.140, 11.540] & $-$1.860 to 1.260 \\
Columbia card & Cards attempted per round & 4.974 & 5.121 [3.738, 6.312] & 0.595 to 1.826 \\
Columbia card & Loss-card endpoint rate & 0.589 & 0.586 [0.443, 0.712] & 0.098 to 0.312 \\
Columbia card & Points per round & 32.28 & 35.08 [11.31, 59.89] & $-$3.31 to 37.57 \\
Columbia card & Higher-loss minus lower-loss cards & $-$1.357 & $-$1.145 [$-$2.682, 0.273] & $-$1.173 to $-$0.045 \\
Columbia card & Late minus early attempted cards & $-$0.716 & $-$0.721 [$-$1.779, 0.357] & $-$0.629 to 0.164 \\
Iowa gambling & Advantageous deck share & 0.564 & 0.514 [0.358, 0.672] & 0.370 to 0.910 \\
Iowa gambling & Change in advantageous deck share & 0.204 & 0.110 [$-$0.150, 0.380] & $-$0.060 to 0.550 \\
Iowa gambling & Switch rate & 0.517 & 0.491 [0.273, 0.719] & 0.022 to 0.422 \\
Iowa gambling & Shift after net loss & 0.719 & 0.711 [0.426, 0.893] & 0.291 to 0.983 \\
Iowa gambling & Shift after net win & 0.474 & 0.449 [0.231, 0.685] & 0.003 to 0.324 \\
\bottomrule
\end{tabular}%
}
\end{table}

\begin{table}[htbp]
\centering
\scriptsize
\setlength{\tabcolsep}{3pt}
\caption{Mean $\bar S$ on shared histories: a trajectory sampled from the human-policy RNN is imposed, and every policy's choice is scored on that same history (five seeds per task). For weather classification the imposed trajectories are those of Appendix~\ref{app:replay}, sampled from an earlier version of its RNN.}
\label{tab:shared_histories}
\resizebox{\linewidth}{!}{%
\begin{tabular}{lrrrrrrrrrrrr}
\toprule
 & \multicolumn{4}{c}{Reference policies} & \multicolumn{4}{c}{DeepSeek} & \multicolumn{4}{c}{Qwen} \\
\cmidrule(lr){2-5}\cmidrule(lr){6-9}\cmidrule(lr){10-13}
Task & RNN sample & RNN greedy & Best const. & Uniform & Stateless & Cumul. & Window & Retr. & Stateless & Cumul. & Window & Retr. \\
\midrule
Iowa gambling & $-$0.799 & $-$0.431 & $-$2.042 & $-$2.233 & $-$0.738 & $-$0.741 & $-$0.736 & $-$0.756 & $-$1.209 & $-$1.081 & $-$0.809 & $-$0.975 \\
Castle two-step & $-$0.763 & $-$0.536 & $-$1.710 & $-$1.864 & $-$1.902 & $-$0.927 & $-$1.058 & $-$0.930 & $-$1.900 & $-$0.977 & $-$1.191 & $-$1.021 \\
Social information bandit & $-$0.524 & $-$0.301 & $-$0.626 & $-$0.987 & $-$1.015 & $-$0.446 & $-$0.423 & $-$0.544 & $-$0.974 & $-$0.425 & $-$0.434 & $-$0.507 \\
Weather classification & $-$0.686 & $-$0.443 & $-$0.744 & $-$0.772 & $-$0.753 & $-$0.718 & $-$0.708 & $-$0.707 & $-$0.712 & $-$0.729 & $-$0.704 & $-$0.722 \\
Mean of four tasks & $-$0.693 & $-$0.428 & $-$1.281 & $-$1.464 & $-$1.102 & $-$0.708 & $-$0.731 & $-$0.734 & $-$1.199 & $-$0.803 & $-$0.785 & $-$0.806 \\
\bottomrule
\end{tabular}%
}
\end{table}

\section{Trajectory-level behavior and prompt budget}
\label{app:trajectory}

\begin{figure}[htbp]
\centering
\IfFileExists{figures/fig_agent_behavior_gpt.pdf}{\includegraphics[width=\linewidth]{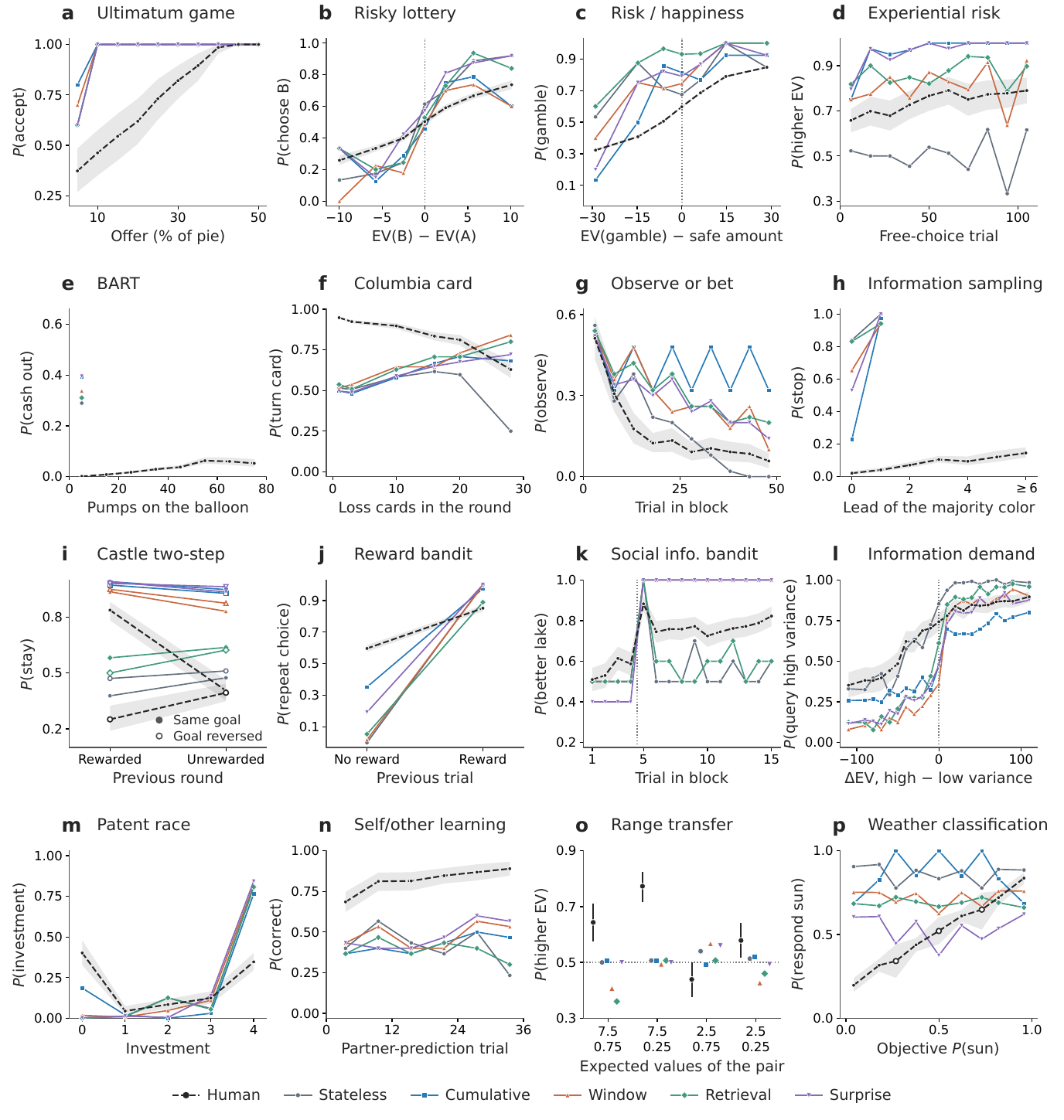}}{\fbox{\parbox{0.9\linewidth}{\centering figures/fig\_agent\_behavior\_gpt.pdf (to be uploaded)}}}
\caption{Behavioral effects of Figure~\ref{fig:reference_behavior} in the episodes of GPT-5.6 Luna: one line per memory configuration (five seeds pooled) against held-out participants (dashed line with its 95\% interval). Points with fewer than five events are omitted; several y-axes are wider than in Figure~\ref{fig:reference_behavior}.}
\label{fig:agent_behavior_gpt}
\end{figure}

\begin{figure}[htbp]
\centering
\IfFileExists{figures/fig_agent_behavior_deepseek.pdf}{\includegraphics[width=\linewidth]{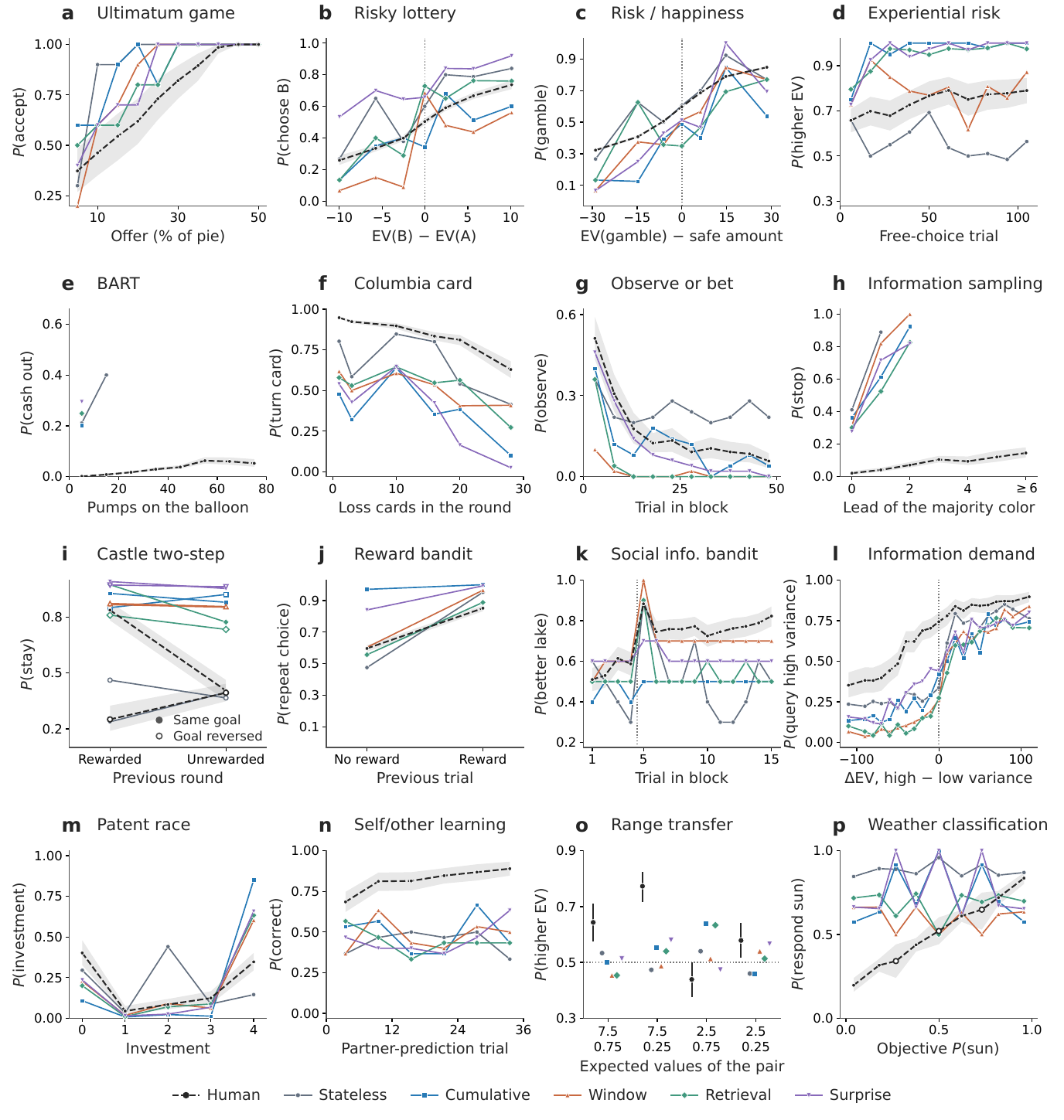}}{\fbox{\parbox{0.9\linewidth}{\centering figures/fig\_agent\_behavior\_deepseek.pdf (to be uploaded)}}}
\caption{As Figure~\ref{fig:agent_behavior_gpt}, for DeepSeek V4.1 Flash.}
\label{fig:agent_behavior_deepseek}
\end{figure}

\begin{figure}[htbp]
\centering
\IfFileExists{figures/fig_agent_behavior_qwen.pdf}{\includegraphics[width=\linewidth]{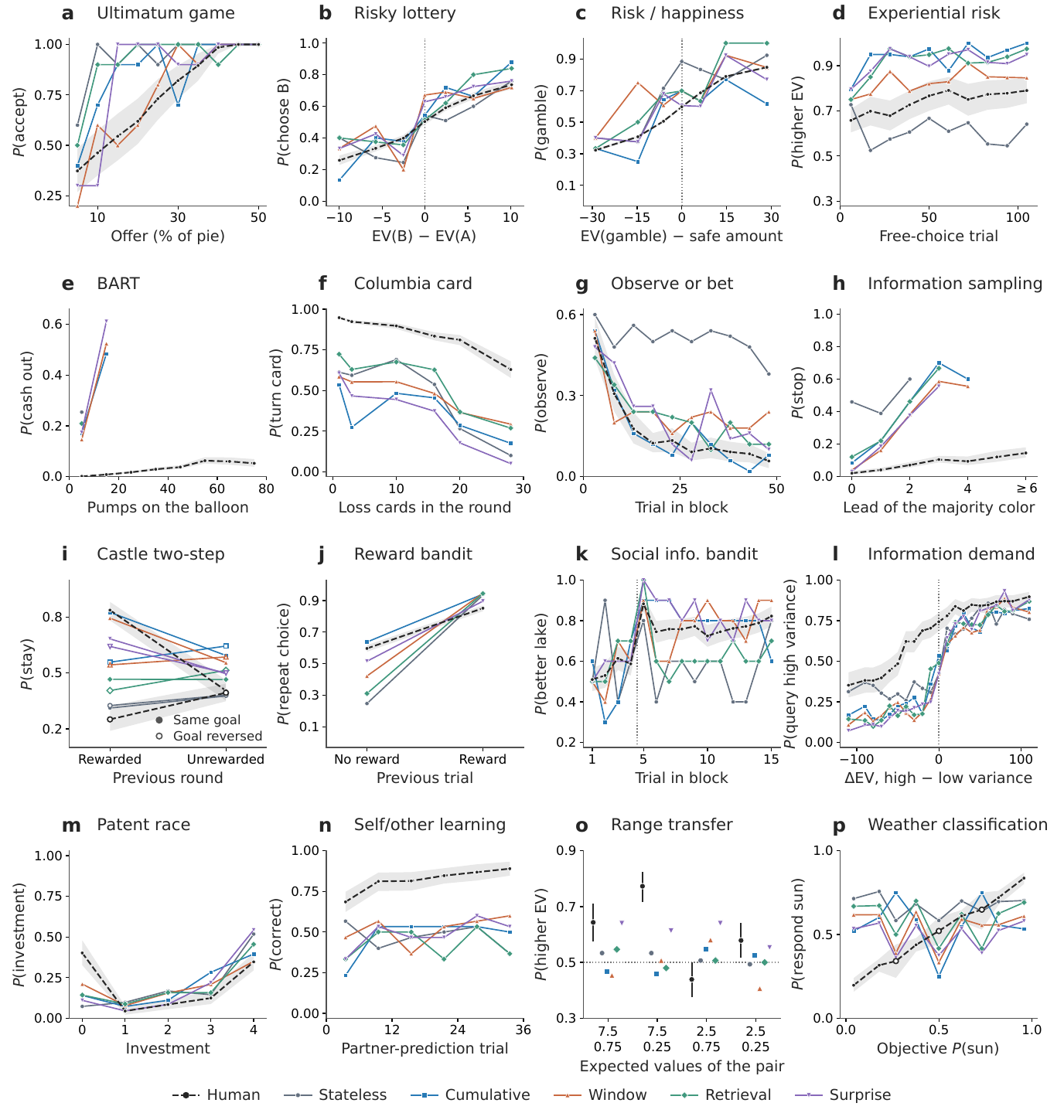}}{\fbox{\parbox{0.9\linewidth}{\centering figures/fig\_agent\_behavior\_qwen.pdf (to be uploaded)}}}
\caption{As Figure~\ref{fig:agent_behavior_gpt}, for Qwen3.8 Flash.}
\label{fig:agent_behavior_qwen}
\end{figure}

\textbf{Behavioral effects in the agents' own episodes.}
Figures~\ref{fig:agent_behavior_gpt}--\ref{fig:agent_behavior_qwen} repeat the analysis of Figure~\ref{fig:reference_behavior} for the LLM agents. For each backbone and memory configuration we computed the same behavioral effect on the agent's own episodes, pooling the five seeds, and plotted it against held-out participants. This comparison does not involve the human-policy RNNs. All 16 effects use the definitions applied to the participants. Memory moves several history-dependent effects toward the participants: the dependence of repeating on the previous reward in the reward bandit (DeepSeek with Window or Retrieval, Qwen with Cumulative), the decline of observing in observe-or-bet (Qwen with Cumulative), and acceptance in the ultimatum game (Qwen with Window). It also produces perseveration that participants do not show: after a rewarded round of Castle two-step, GPT (Cumulative, Window, Surprise) and DeepSeek (all four memory configurations) repeat their first-stage choice even when the goal has been reversed (0.81--0.99; participants 0.25). In five tasks every configuration of every backbone is far from the participants: responses do not depend on the objective probability in weather classification, there is no learning in self/other learning, transfer choices stay near 0.5 in range learning, and agents cash out after a few pumps in BART and turn far fewer cards in the Columbia card task. In BART and information sampling the agents reach only the first bins of the participants' range.

\textbf{Trajectory-level features.}
We computed 44 pre-specified behavioral features in the nine branching tasks of Appendix~\ref{app:autonomous_behavior} and asked whether each agent's value lies inside the central 95\% interval of the RNN's own rollouts (Table~\ref{tab:trajectory_inside}). Memory moves features that require using past outcomes into the human range, but it also makes agents more likely to repeat previous decisions than humans, most strongly under Cumulative: the share of features inside the interval falls from 53\% under Stateless to 34--47\% with memory (Table~\ref{tab:trajectory_examples}). In the risk tasks every configuration is far from human behavior.

\begin{figure}[htbp]
\centering
\IfFileExists{figures/fig_token_budget_regression.pdf}{\includegraphics[width=\linewidth]{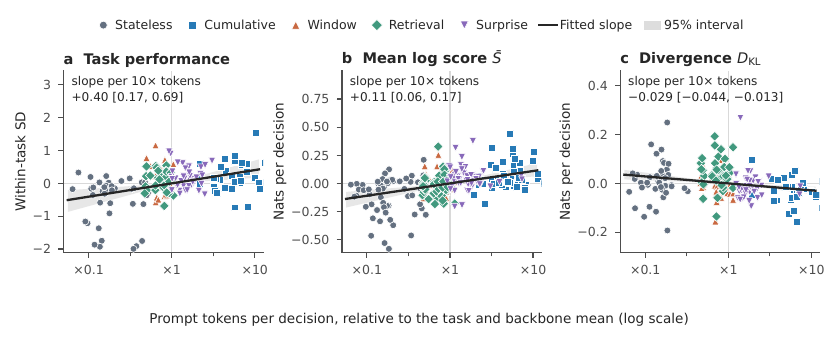}}{\fbox{\parbox{0.9\linewidth}{\centering figures/fig\_token\_budget\_regression.pdf (to be uploaded)}}}
\caption{Prompt length and outcomes within task and backbone. Each point is one task, backbone and memory configuration (mean over seeds); both axes are centered within task and backbone. Lines show the slope of a regression with task-by-backbone fixed effects fitted to the episodes, and the band its 95\% task-bootstrap interval. (a)~Task performance, standardized within task (13 tasks with a performance measure). (b)~$\bar S$ (20 tasks, three backbones). (c)~$D_{\mathrm{KL}}$ (20 tasks, DeepSeek and Qwen). With memory configuration in the regression, the slopes are $-0.07$ $[-0.47, 0.33]$, $0.04$ $[-0.04, 0.14]$ and $-0.03$ $[-0.08, 0.03]$.}
\label{fig:token_regression}
\end{figure}

\textbf{Prompt budget and seeds.}
Memory configurations differ in prompt tokens by design (Table~\ref{tab:appendix_context_usage}). Table~\ref{tab:budget_contrasts} separates mechanism from budget. Window and Retrieval use the same budget: their difference in $\bar S$ is small (0.015--0.044; the interval excludes zero only for GPT, narrowly), but Window has the lower divergence for both backbones with action probabilities. Cumulative uses five to nine times the tokens of Window for a gain in $\bar S$ whose interval includes zero for every backbone. Seed-to-seed variation accounts for about 2\% of the variance of the pooled memory effect, which has the same sign in every seed; five seeds do not suffice to rank individual configurations within a task. Figure~\ref{fig:token_regression} relates prompt length to the outcomes within task and backbone: a tenfold increase in prompt tokens per decision goes with higher task performance ($0.40$ within-task standard deviations, CI $[0.17, 0.69]$), higher $\bar S$ ($0.11$, CI $[0.06, 0.17]$) and lower $D_{\mathrm{KL}}$ ($-0.029$, CI $[-0.044, -0.013]$). Because each configuration has its own budget, prompt length and configuration are nearly collinear; with configuration in the regression, none of the three slopes differs from zero, so the equal-budget contrast in Table~\ref{tab:budget_contrasts} is the informative test.

\begin{table}[htbp]
\centering
\small
\setlength{\tabcolsep}{3pt}
\caption{Behavioral features inside the central 95\% interval of the human-policy RNN's rollouts (44 features, nine tasks), and the paired change in log distance to the RNN median relative to Stateless (95\% interval).}
\label{tab:trajectory_inside}
\begin{tabular}{lrrrrr}
\toprule
Configuration & All backbones & GPT & DeepSeek & Qwen & Distance vs.\ Stateless \\
\midrule
Stateless & 67/126 (53\%) & 21/38 (55\%) & 24/44 (55\%) & 22/44 (50\%) & -- \\
Cumulative & 43/127 (34\%) & 13/39 (33\%) & 11/44 (25\%) & 19/44 (43\%) & +0.44 [+0.23, +0.66] \\
Window & 55/127 (43\%) & 15/39 (38\%) & 18/44 (41\%) & 22/44 (50\%) & +0.22 [$-$0.01, +0.47] \\
Retrieval & 57/122 (47\%) & 17/39 (44\%) & 15/39 (38\%) & 25/44 (57\%) & +0.17 [$-$0.01, +0.34] \\
Surprise & 54/126 (43\%) & 15/38 (39\%) & 16/44 (36\%) & 23/44 (52\%) & +0.30 [+0.07, +0.55] \\
\bottomrule
\end{tabular}
\end{table}

\begin{table}[htbp]
\centering
\scriptsize
\setlength{\tabcolsep}{3pt}
\caption{Examples of behavioral features by memory configuration (range over the three backbones).}
\label{tab:trajectory_examples}
\resizebox{\linewidth}{!}{%
\begin{tabular}{lrrrrrr}
\toprule
Feature & RNN rollouts, median [95\%] & Stateless & Cumulative & Window & Retrieval & Surprise \\
\midrule
Two-step: effect of reward on repeating the first choice & 0.050 [$-$0.002, 0.105] & $-$0.013 to 0.007 & 0.038 to 0.071 & 0.029 to 0.058 & $-$0.084 to 0.029 & 0.047 to 0.103 \\
Two-step: repeats first choice & 0.49 [0.39, 0.63] & 0.36 to 0.48 & 0.67 to 0.96 & 0.62 to 0.88 & 0.46 to 0.80 & 0.58 to 0.97 \\
Two-step: entropy of first choices & 0.89 [0.72, 0.97] & 0.77 to 0.97 & 0.15 to 0.56 & 0.44 to 0.91 & 0.36 to 0.85 & 0.09 to 0.79 \\
Iowa: switch rate & 0.49 [0.27, 0.72] & 0.14 to 0.33 & 0.03 to 0.20 & 0.02 to 0.17 & 0.05 to 0.42 & 0.05 to 0.22 \\
Iowa: share of advantageous decks & 0.51 [0.36, 0.67] & 0.53 to 0.61 & 0.70 to 0.91 & 0.55 to 0.89 & 0.37 to 0.61 & 0.71 to 0.78 \\
BART: pumps per balloon & 37.4 [30.7, 43.2] & 2.4 to 3.6 & 1.5 to 4.3 & 1.9 to 5.3 & 2.1 to 3.7 & 1.5 to 4.3 \\
Columbia card: cards per round & 5.1 [3.7, 6.3] & 0.9 to 1.8 & 0.6 to 0.9 & 0.9 to 1.0 & 1.0 to 1.4 & 0.7 to 0.9 \\
\bottomrule
\end{tabular}%
}
\end{table}

\begin{table}[htbp]
\centering
\small
\setlength{\tabcolsep}{3pt}
\caption{Contrasts at equal prompt budget (Window minus Retrieval) and at a larger budget (Cumulative minus Window): paired differences of equal-task means with 95\% task-bootstrap intervals.}
\label{tab:budget_contrasts}
\begin{tabular}{llrrr}
\toprule
Contrast & Backbone & Token ratio & $\Delta\bar S$ & $\Delta D_{\mathrm{KL}}$ \\
\midrule
Window $-$ Retrieval & GPT & 1.0$\times$ & +0.044 [+0.000, +0.088] & -- \\
& DeepSeek & 1.0$\times$ & +0.015 [$-$0.028, +0.058] & $-$0.048 [$-$0.082, $-$0.014] \\
& Qwen & 1.0$\times$ & +0.023 [$-$0.017, +0.067] & $-$0.038 [$-$0.075, $-$0.006] \\
Cumulative $-$ Window & GPT & 5.5$\times$ & +0.057 [$-$0.004, +0.120] & -- \\
& DeepSeek & 9.0$\times$ & +0.069 [$-$0.001, +0.148] & $-$0.046 [$-$0.077, $-$0.014] \\
& Qwen & 9.2$\times$ & +0.041 [$-$0.008, +0.091] & $-$0.013 [$-$0.041, +0.019] \\
\bottomrule
\end{tabular}
\end{table}

\clearpage
\section{Task Workflow Illustrations}
\label{app:task_workflow_illustrations}

The following illustrations summarize the participant-facing interaction
flow of all 25 environments.  Displayed numerical stimuli show
representative rounds; task schedules, feedback conditions, and response
stages follow the protocols summarized in
Tables~\ref{tab:environment_inventory_a}--\ref{tab:environment_inventory_b}.
The Stroop panel depicts the source task's visual color conflict; our text
adaptation presents the ink color as an explicit annotation.

\begin{center}
\includegraphics[width=\textwidth,height=0.13\textheight,trim=0 60bp 0 60bp,clip,keepaspectratio]{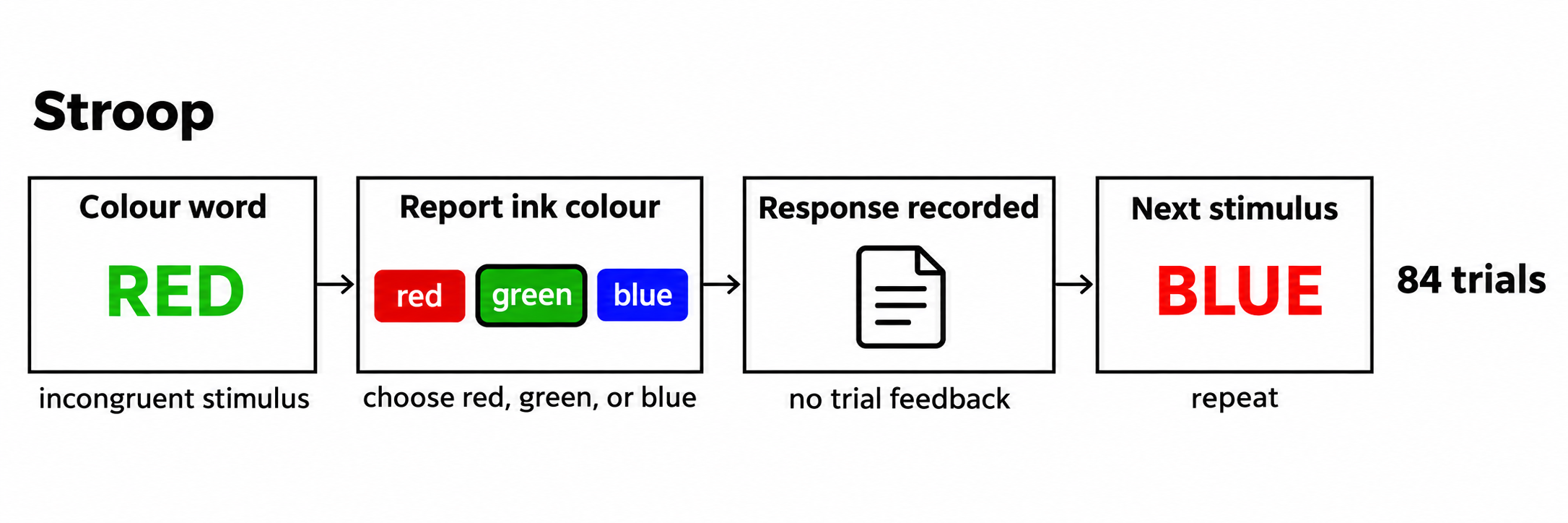}\par\vspace{1mm}
\includegraphics[width=\textwidth,height=0.13\textheight,trim=0 60bp 0 60bp,clip,keepaspectratio]{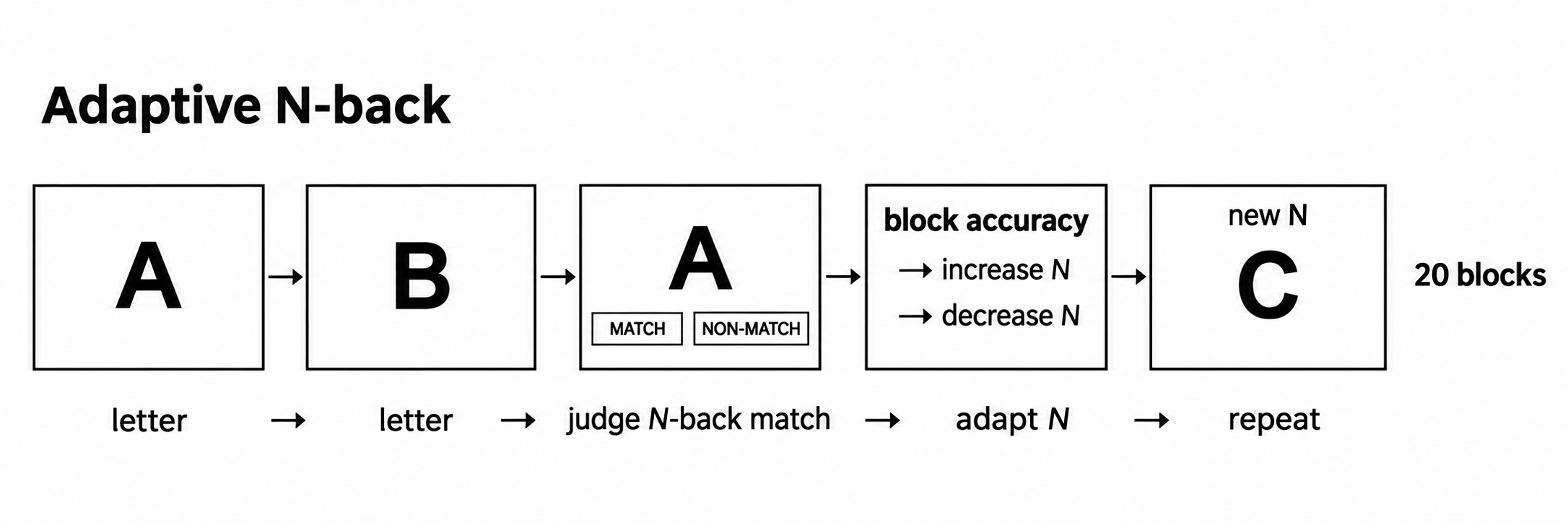}\par\vspace{1mm}
\includegraphics[width=\textwidth,height=0.13\textheight,trim=0 60bp 0 60bp,clip,keepaspectratio]{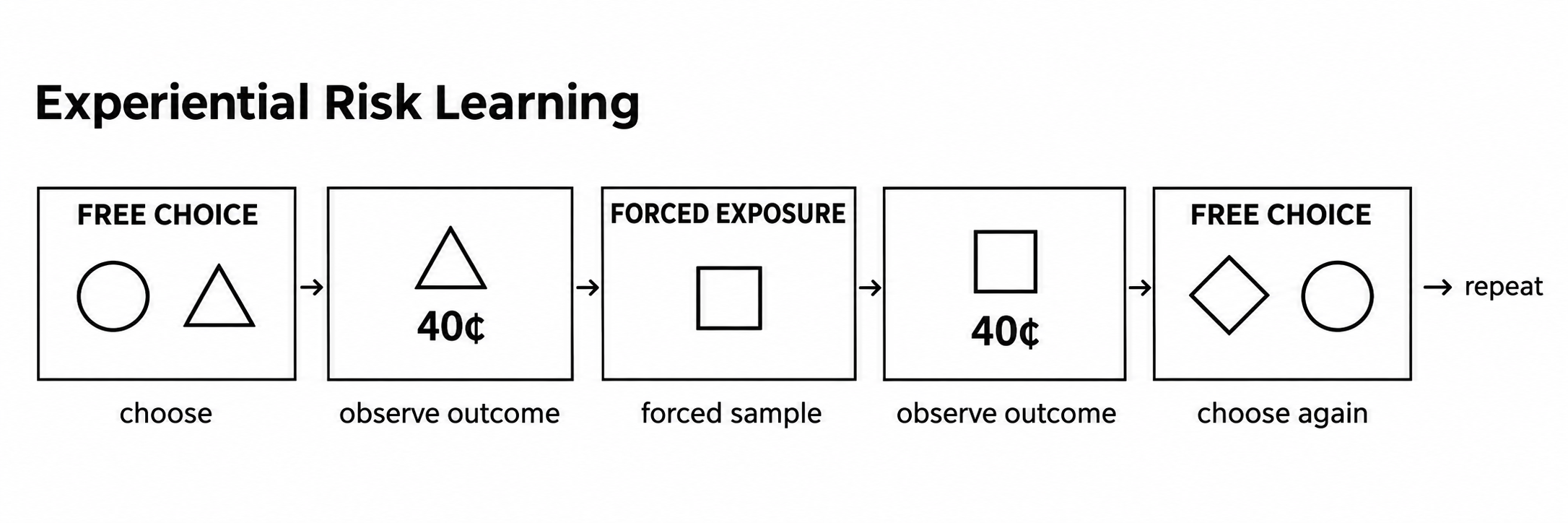}\par\vspace{1mm}
\includegraphics[width=\textwidth,height=0.19\textheight,trim=0 50bp 0 50bp,clip,keepaspectratio]{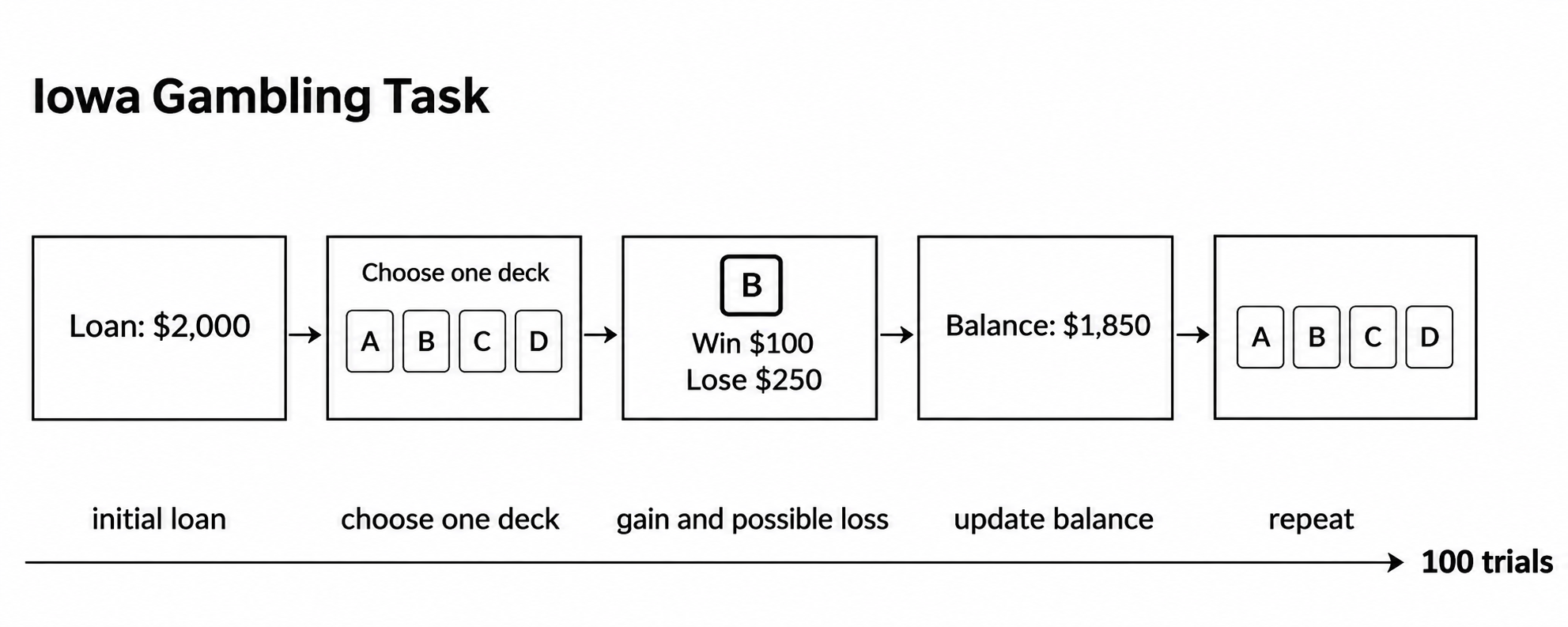}\par\vspace{1mm}
\includegraphics[width=\textwidth,height=0.18\textheight,trim=0 25bp 0 45bp,clip,keepaspectratio]{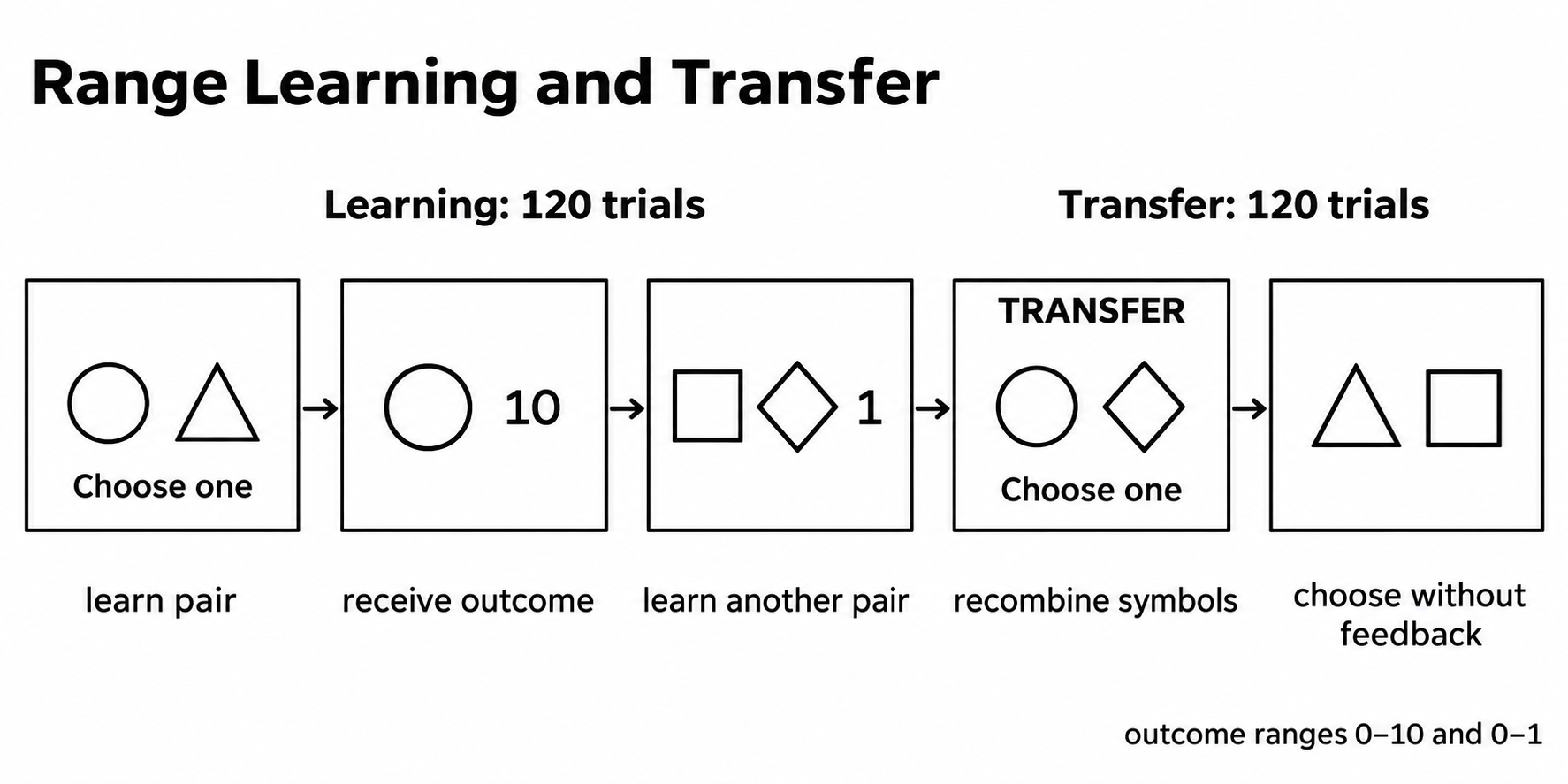}
\end{center}

\clearpage
\begin{center}
\includegraphics[width=\textwidth,height=0.15\textheight,trim=0 60bp 0 60bp,clip,keepaspectratio]{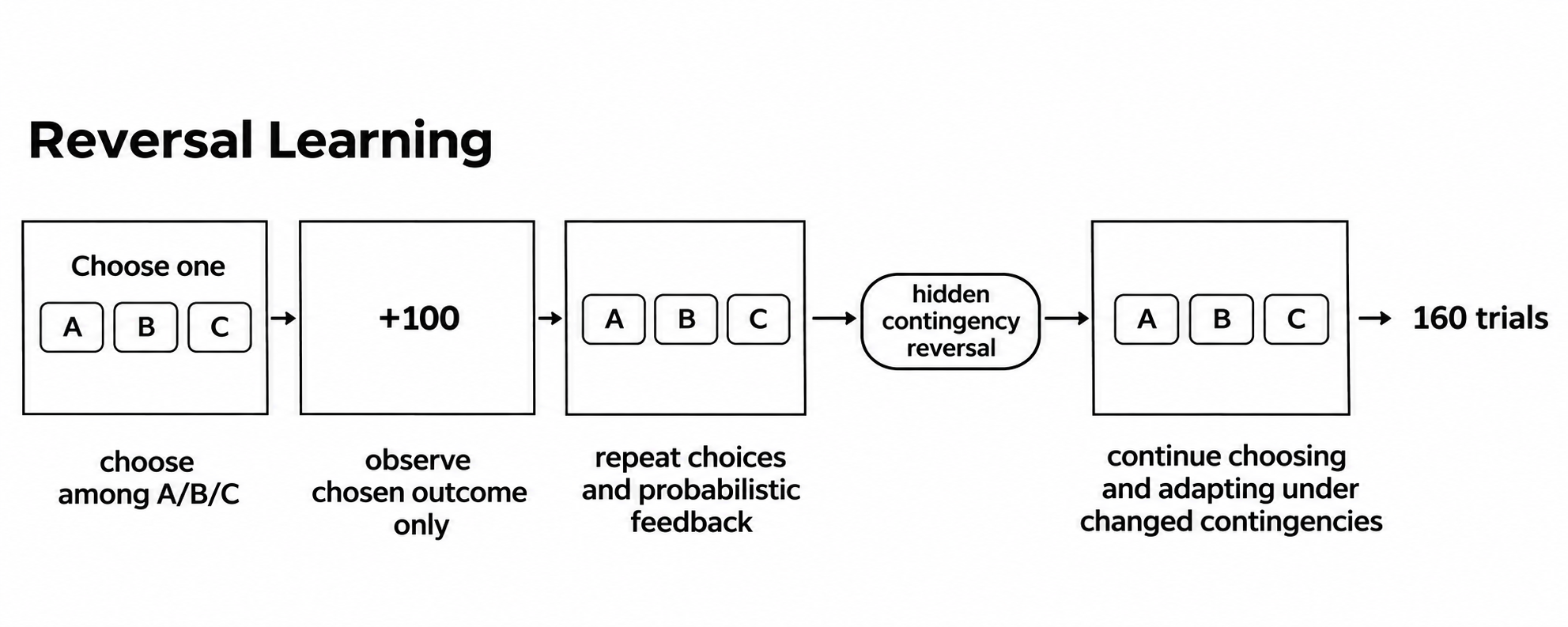}\par\vspace{1mm}
\includegraphics[width=\textwidth,height=0.14\textheight,trim=0 60bp 0 60bp,clip,keepaspectratio]{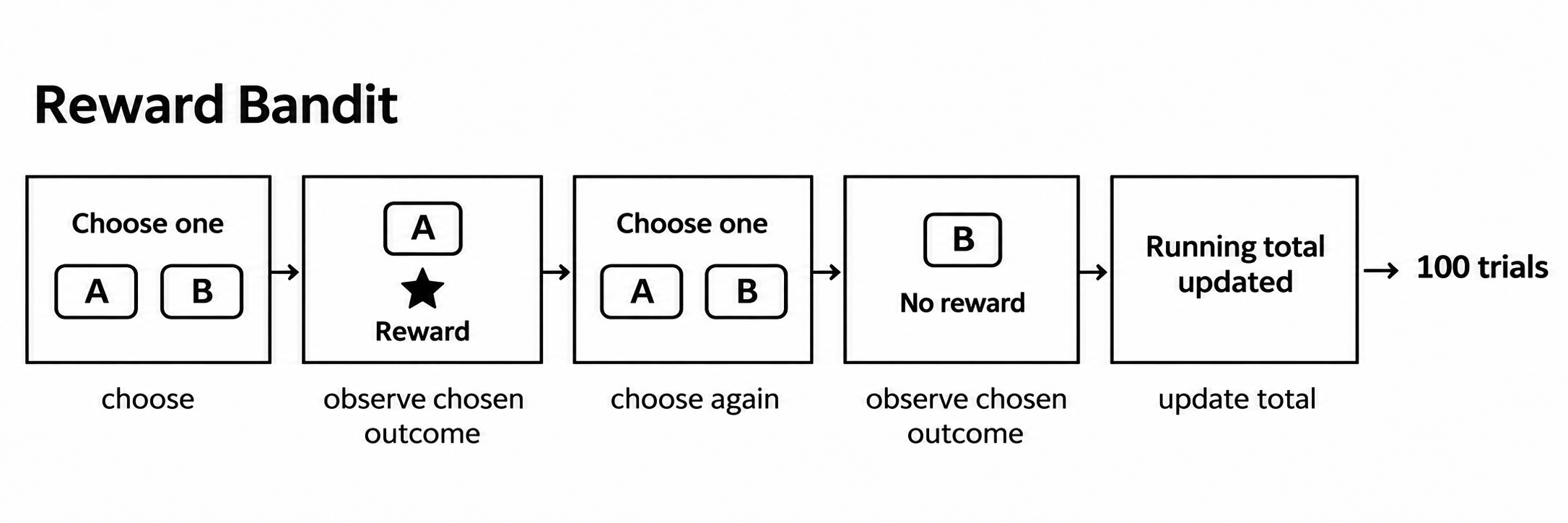}\par\vspace{1mm}
\includegraphics[width=\textwidth,height=0.13\textheight,trim=0 60bp 0 60bp,clip,keepaspectratio]{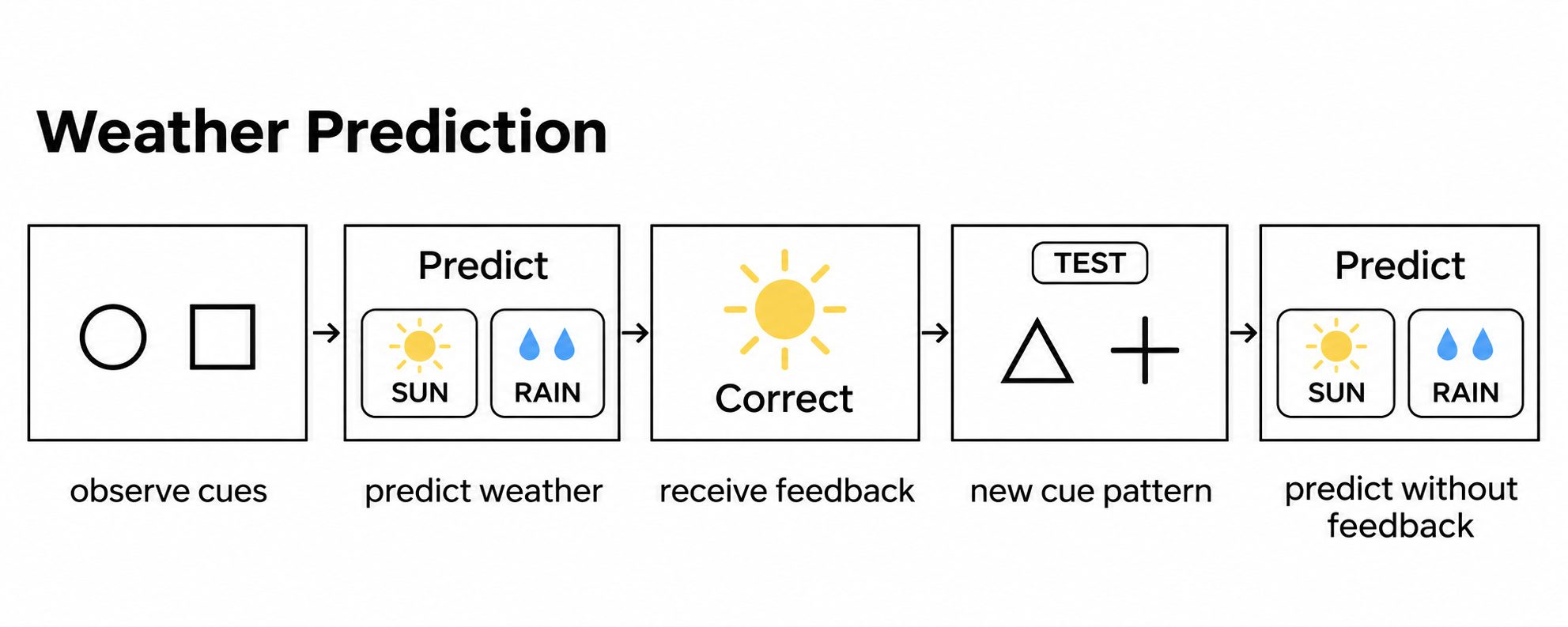}\par\vspace{1mm}
\includegraphics[width=\textwidth,height=0.17\textheight,keepaspectratio]{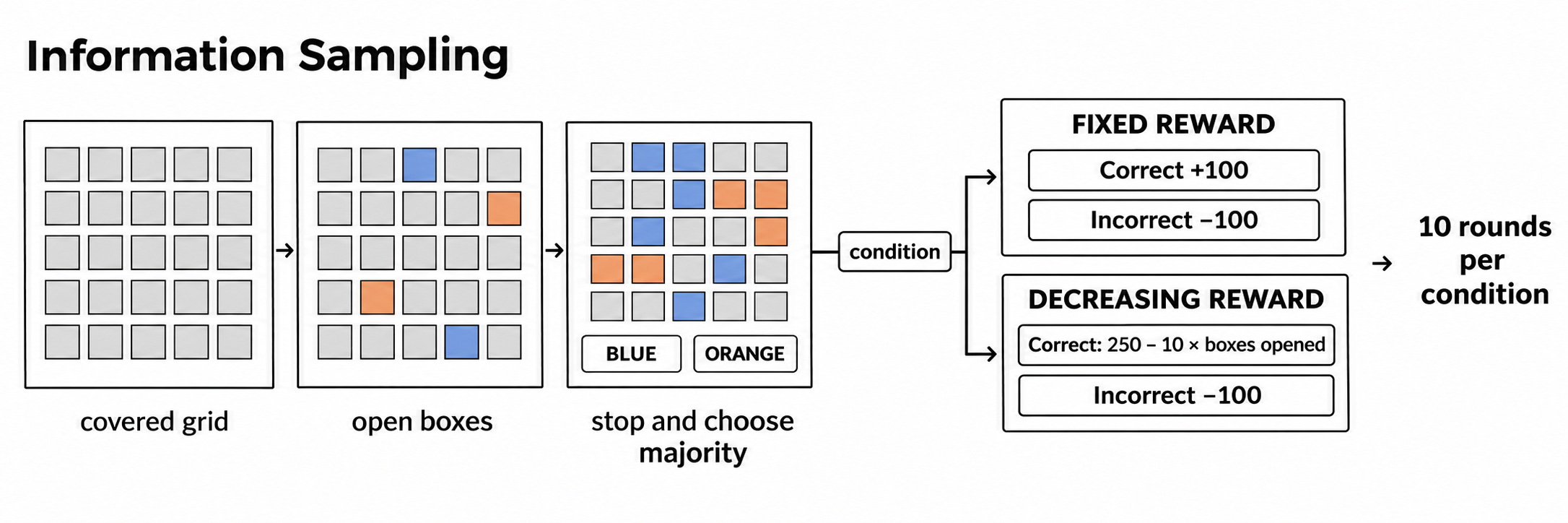}\par\vspace{1mm}
\includegraphics[width=\textwidth,height=0.20\textheight,keepaspectratio]{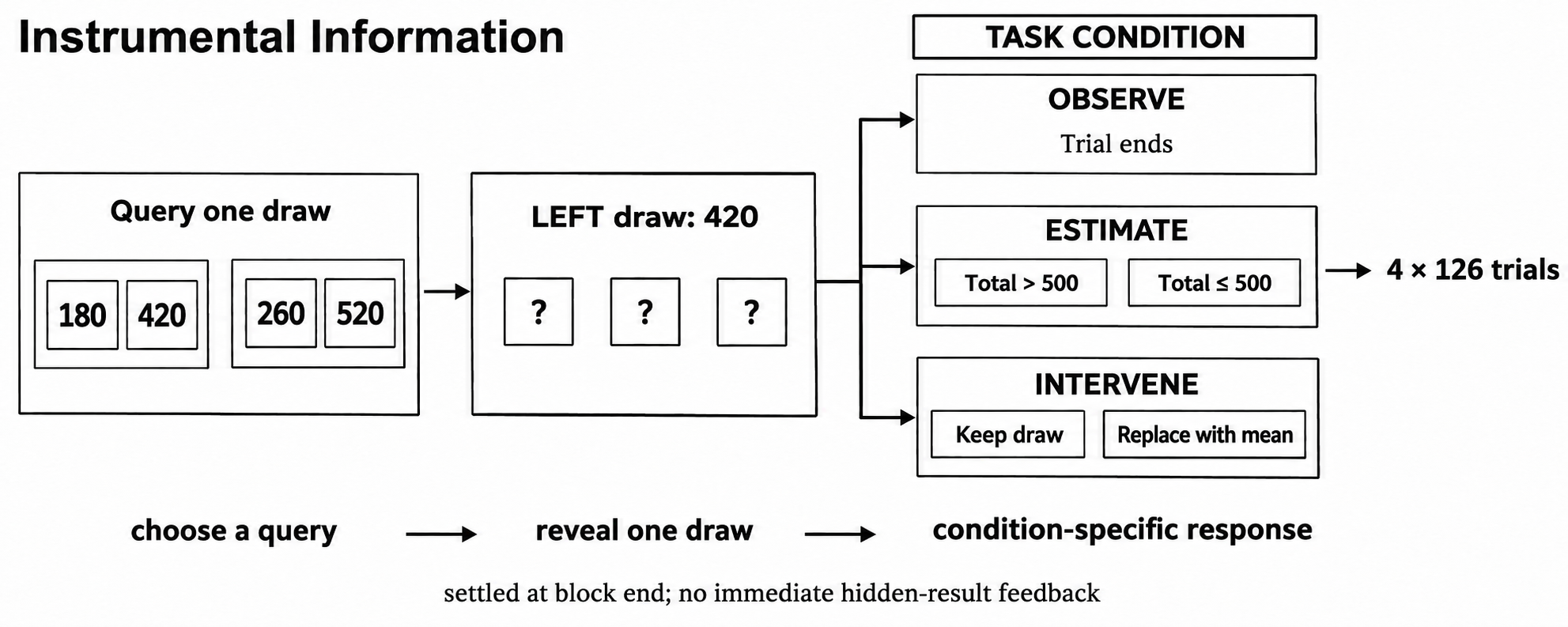}
\end{center}

\clearpage
\begin{center}
\includegraphics[width=\textwidth,height=0.17\textheight,keepaspectratio]{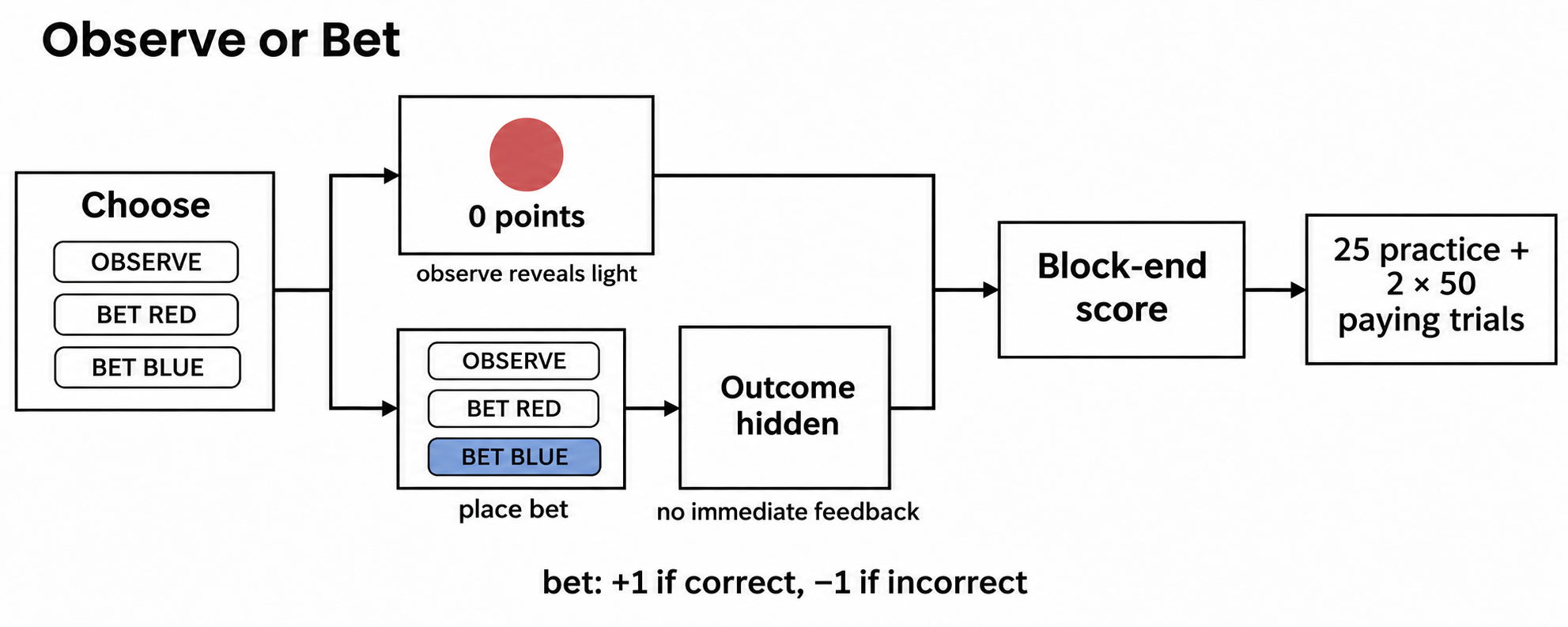}\par\vspace{1mm}
\includegraphics[width=\textwidth,height=0.16\textheight,trim=0 60bp 0 60bp,clip,keepaspectratio]{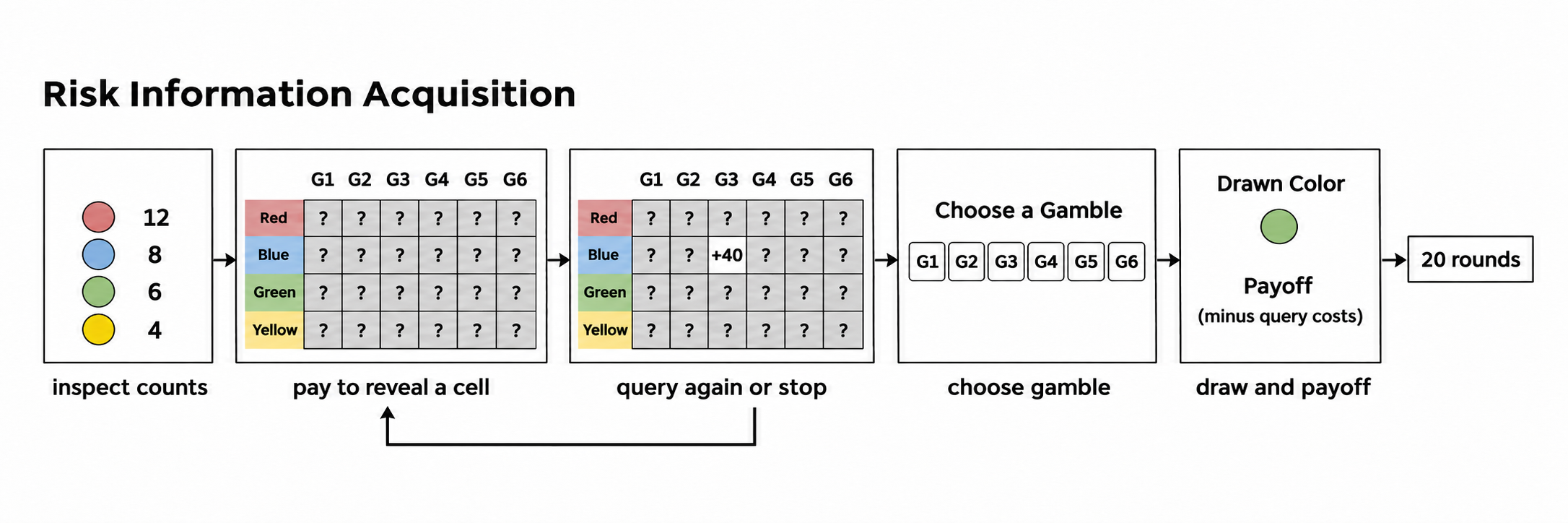}\par\vspace{1mm}
\includegraphics[width=\textwidth,height=0.14\textheight,trim=0 60bp 0 60bp,clip,keepaspectratio]{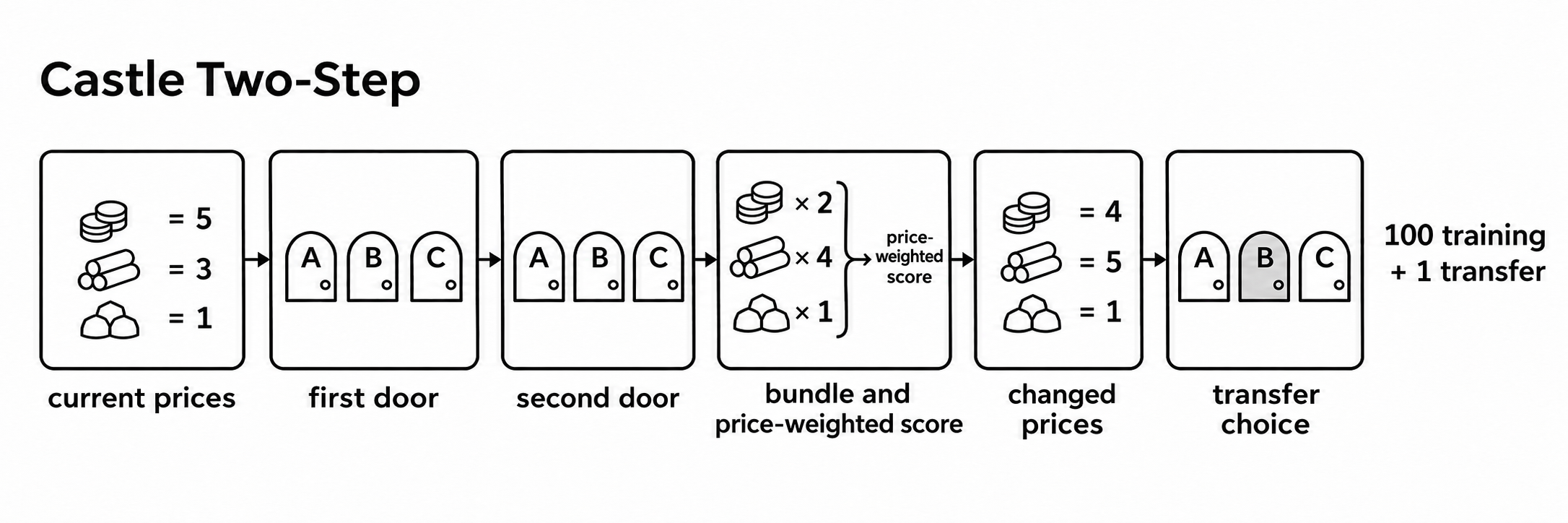}\par\vspace{1mm}
\includegraphics[width=\textwidth,height=0.13\textheight,trim=0 60bp 0 45bp,clip,keepaspectratio]{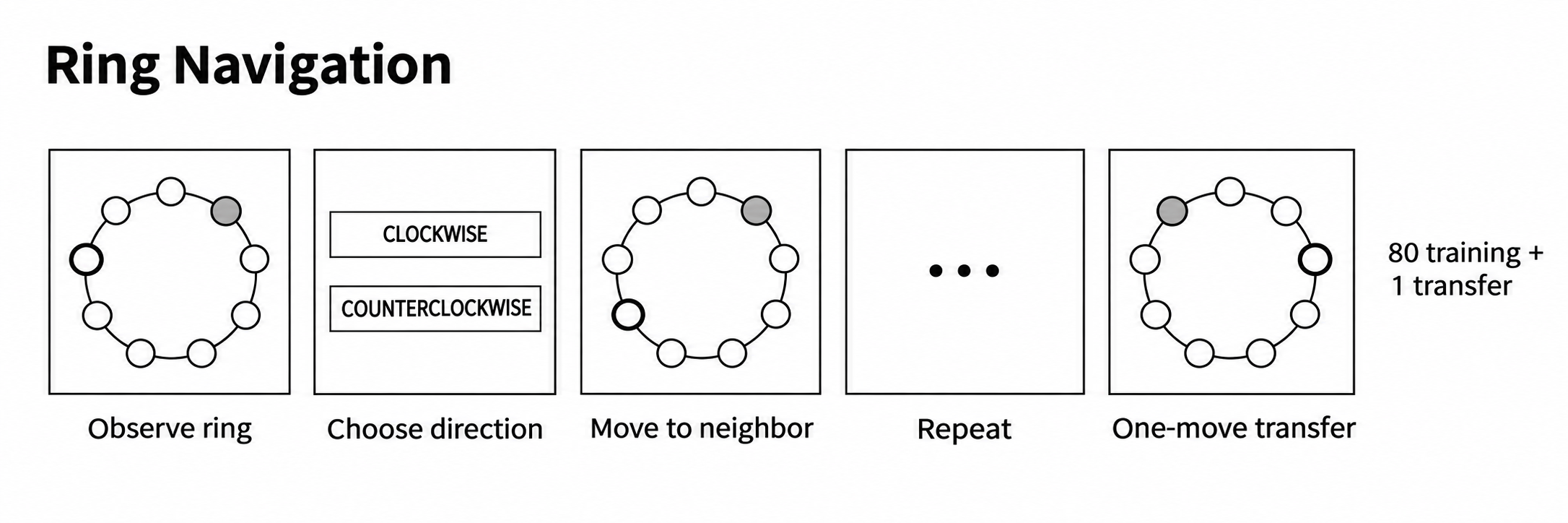}\par\vspace{1mm}
\includegraphics[width=\textwidth,height=0.17\textheight,trim=0 60bp 0 60bp,clip,keepaspectratio]{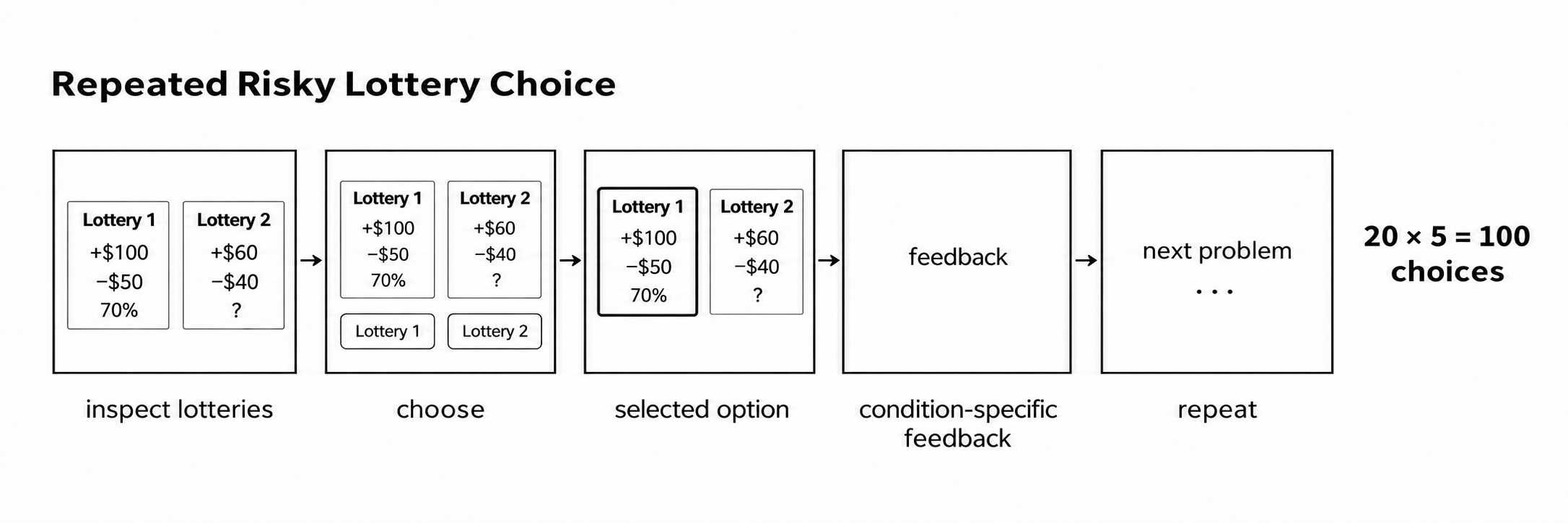}
\end{center}

\clearpage
\begin{center}
\includegraphics[width=\textwidth,height=0.20\textheight,keepaspectratio]{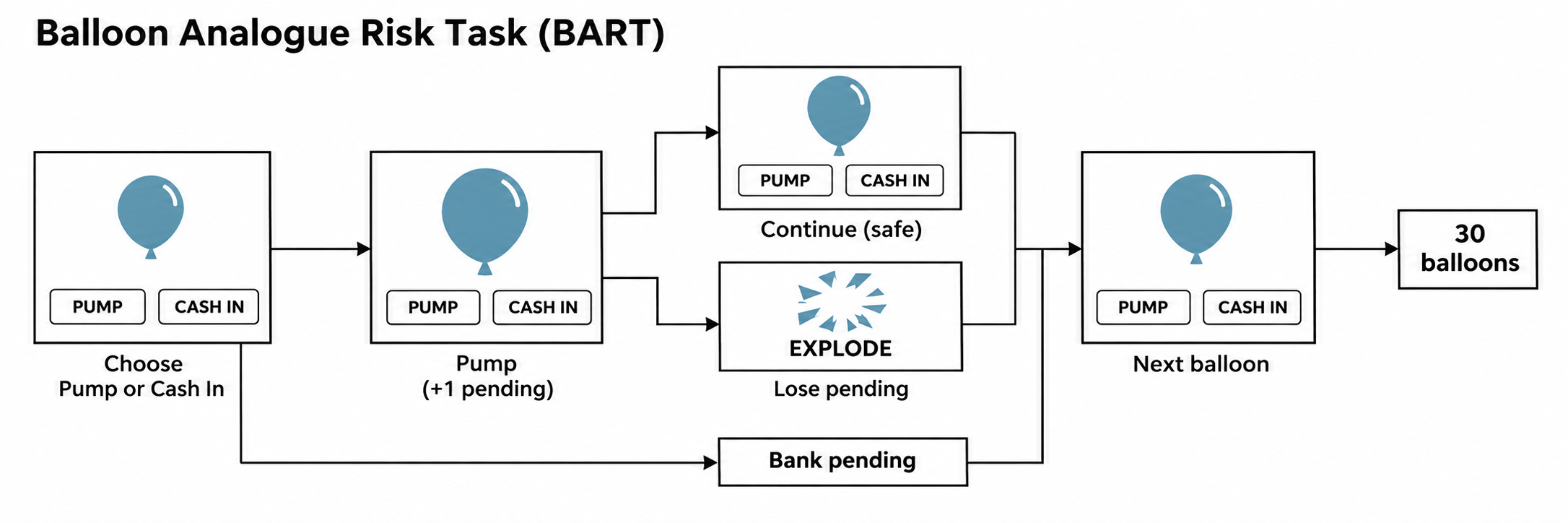}\par\vspace{1mm}
\includegraphics[width=\textwidth,height=0.24\textheight,keepaspectratio]{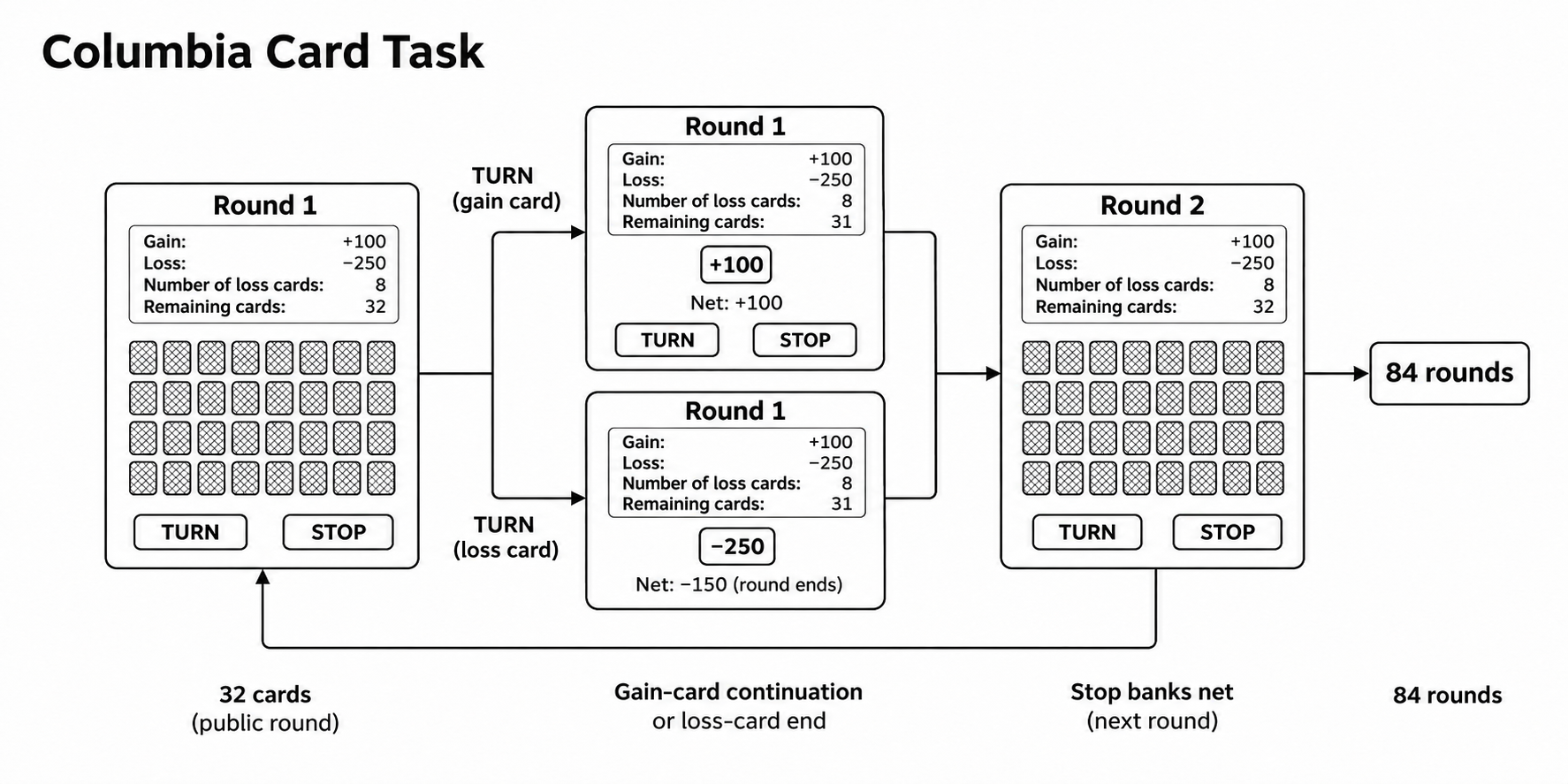}\par\vspace{1mm}
\includegraphics[width=\textwidth,height=0.12\textheight,trim=0 60bp 0 60bp,clip,keepaspectratio]{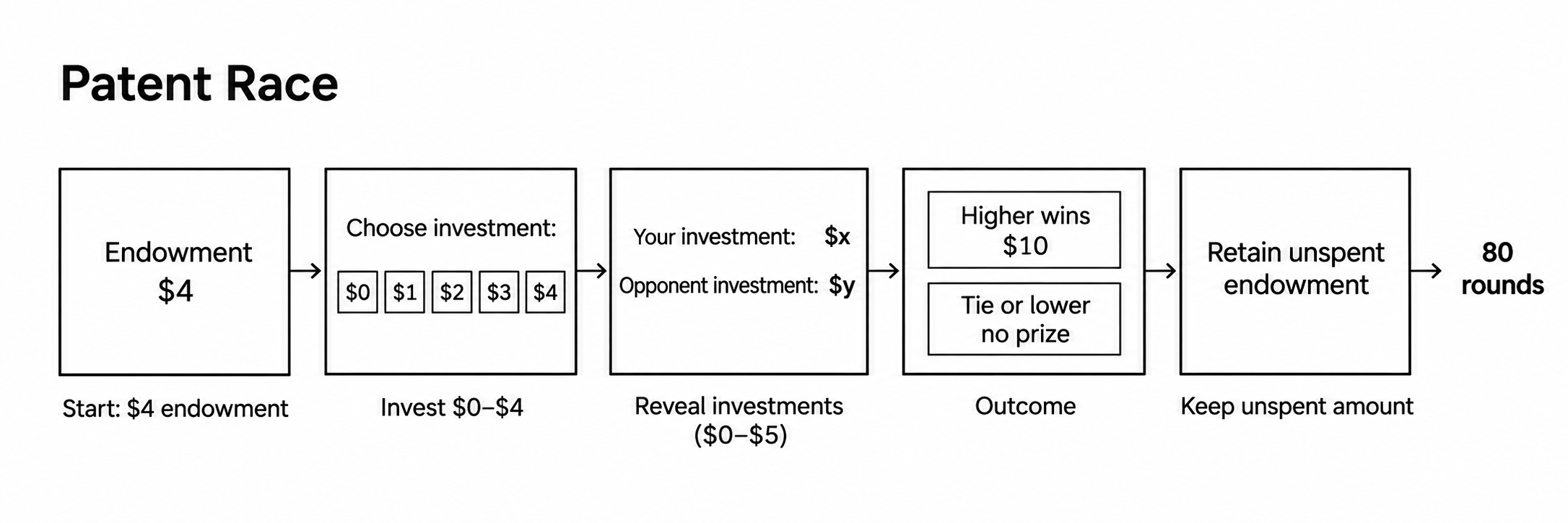}\par\vspace{1mm}
\includegraphics[width=\textwidth,height=0.14\textheight,trim=0 60bp 0 60bp,clip,keepaspectratio]{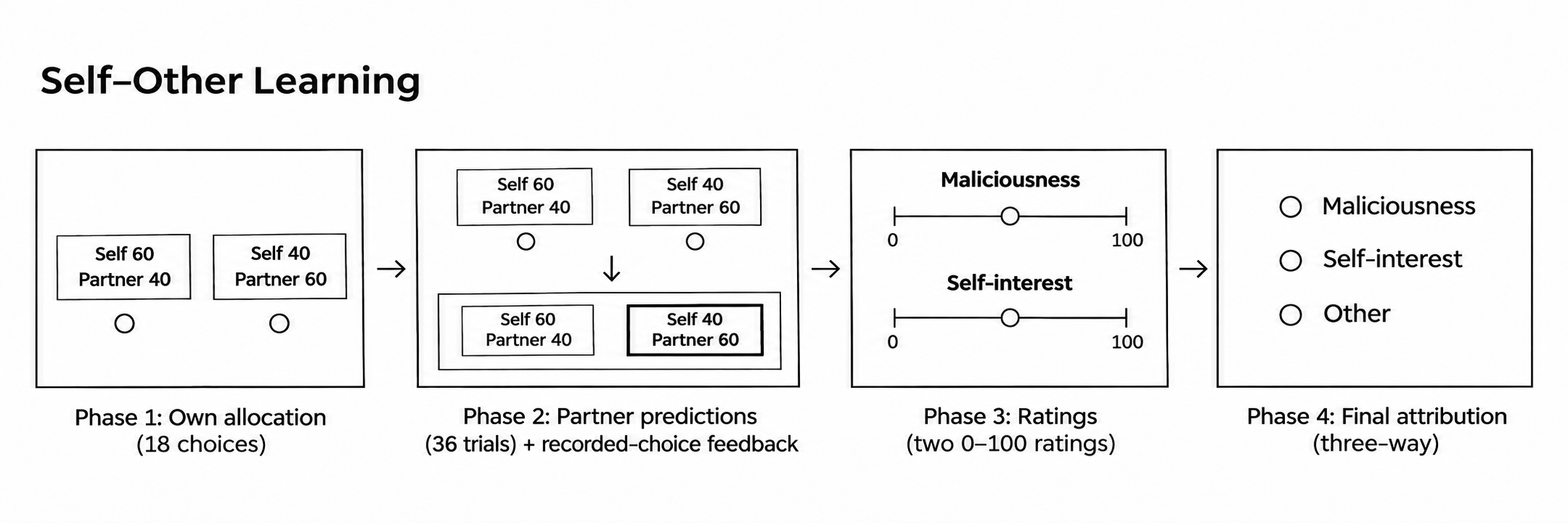}\par\vspace{1mm}
\includegraphics[width=\textwidth,height=0.16\textheight,keepaspectratio]{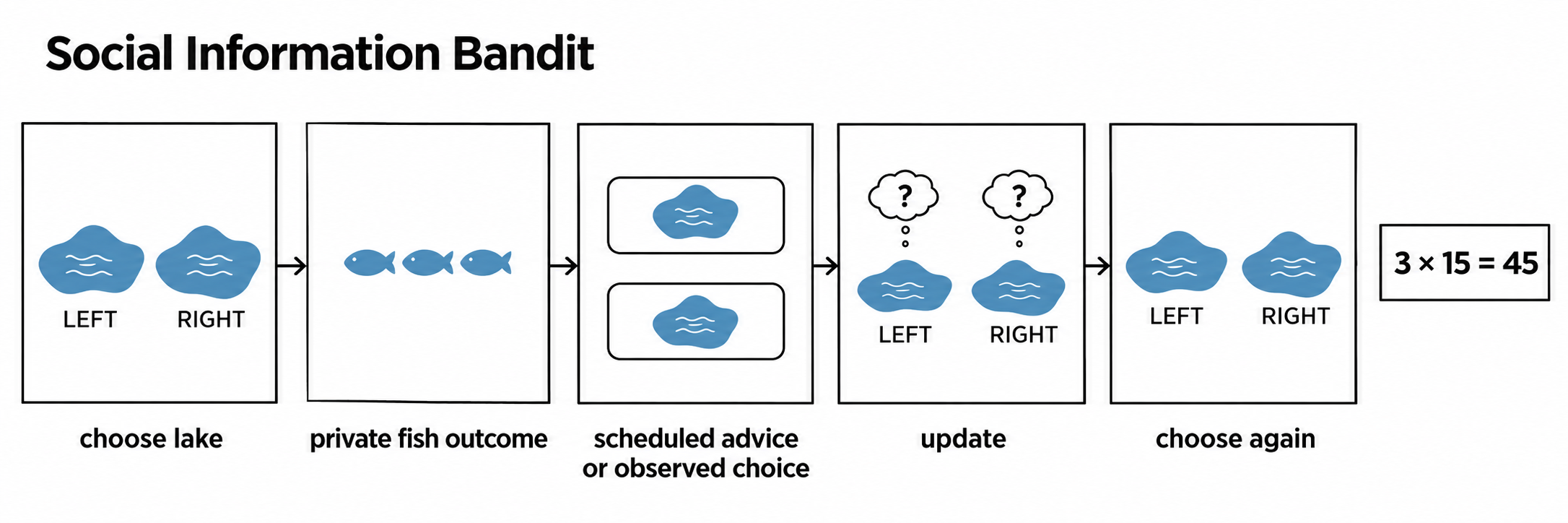}
\end{center}

\clearpage
\begin{center}
\includegraphics[width=\textwidth,height=0.15\textheight,trim=0 60bp 0 60bp,clip,keepaspectratio]{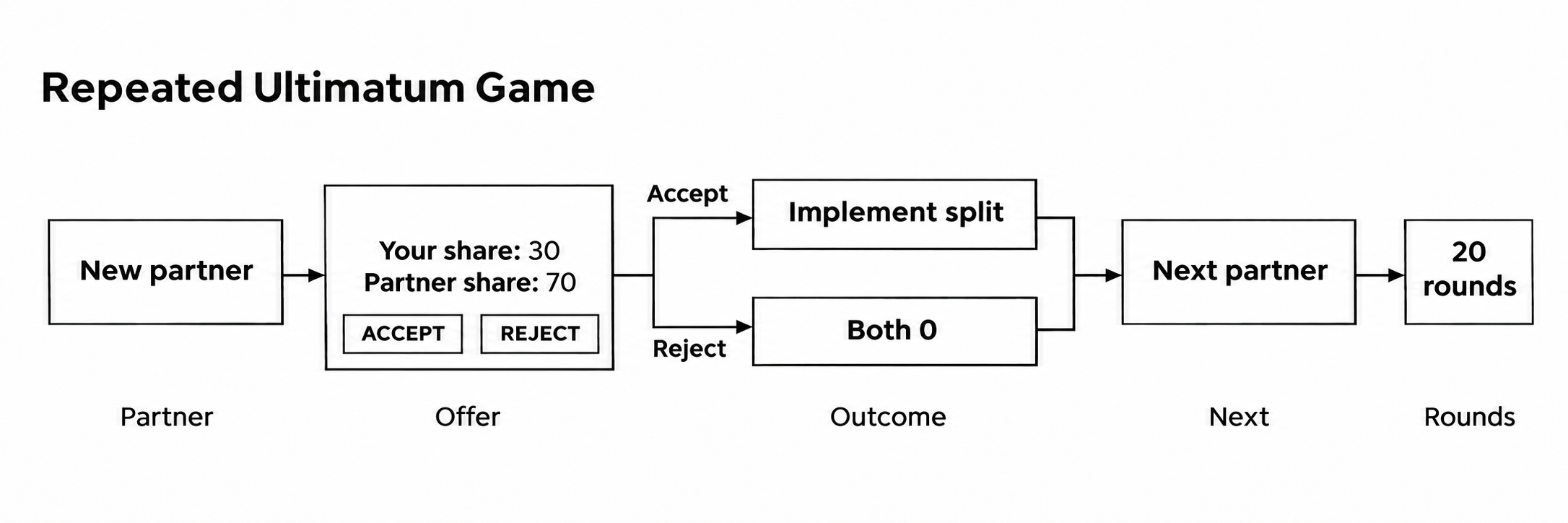}\par\vspace{1mm}
\includegraphics[width=\textwidth,height=0.16\textheight,keepaspectratio]{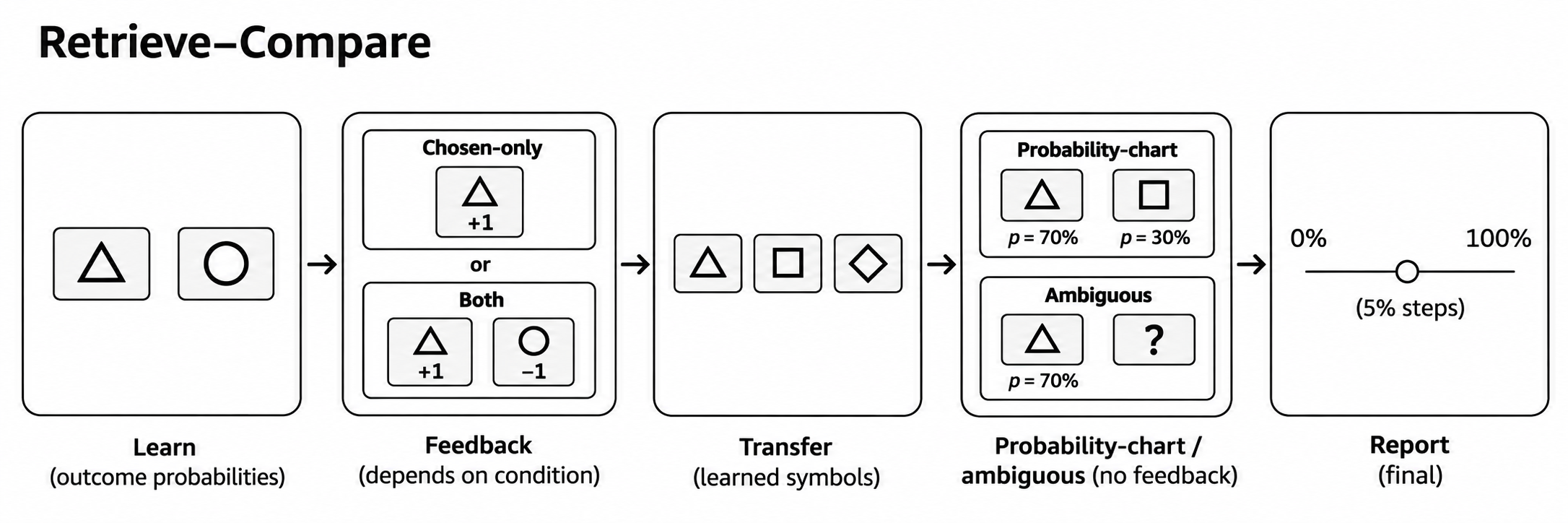}\par\vspace{1mm}
\includegraphics[width=\textwidth,height=0.15\textheight,trim=0 60bp 0 60bp,clip,keepaspectratio]{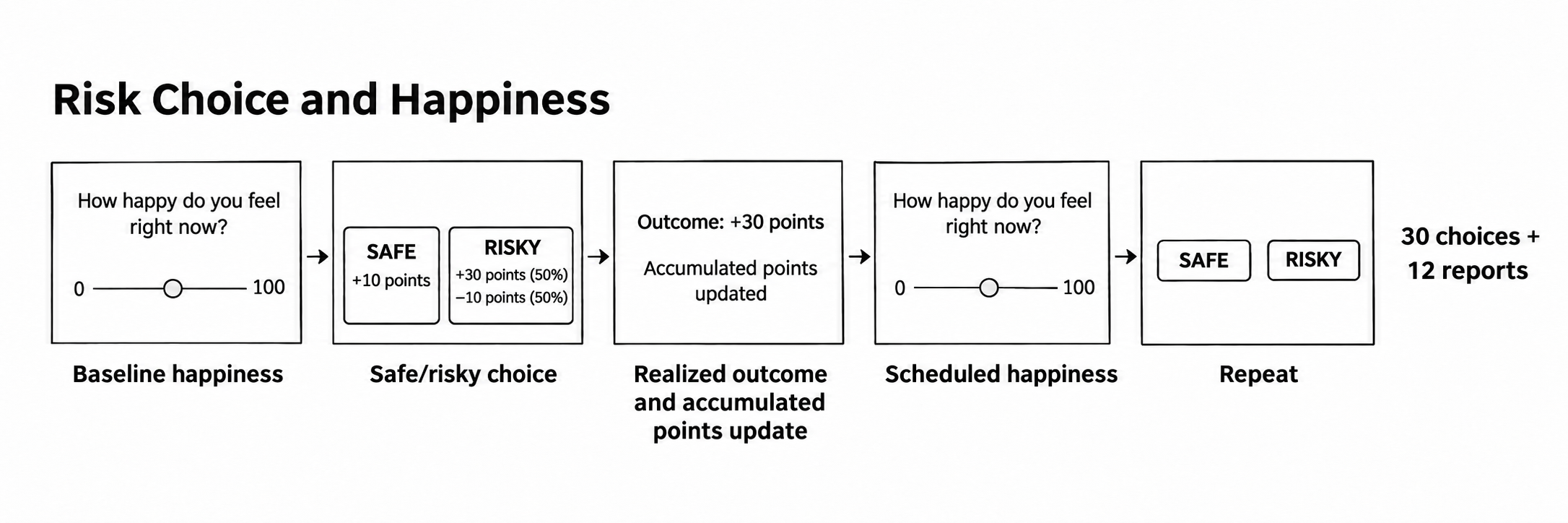}\par\vspace{1mm}
\includegraphics[width=\textwidth,height=0.16\textheight,trim=0 80bp 0 60bp,clip,keepaspectratio]{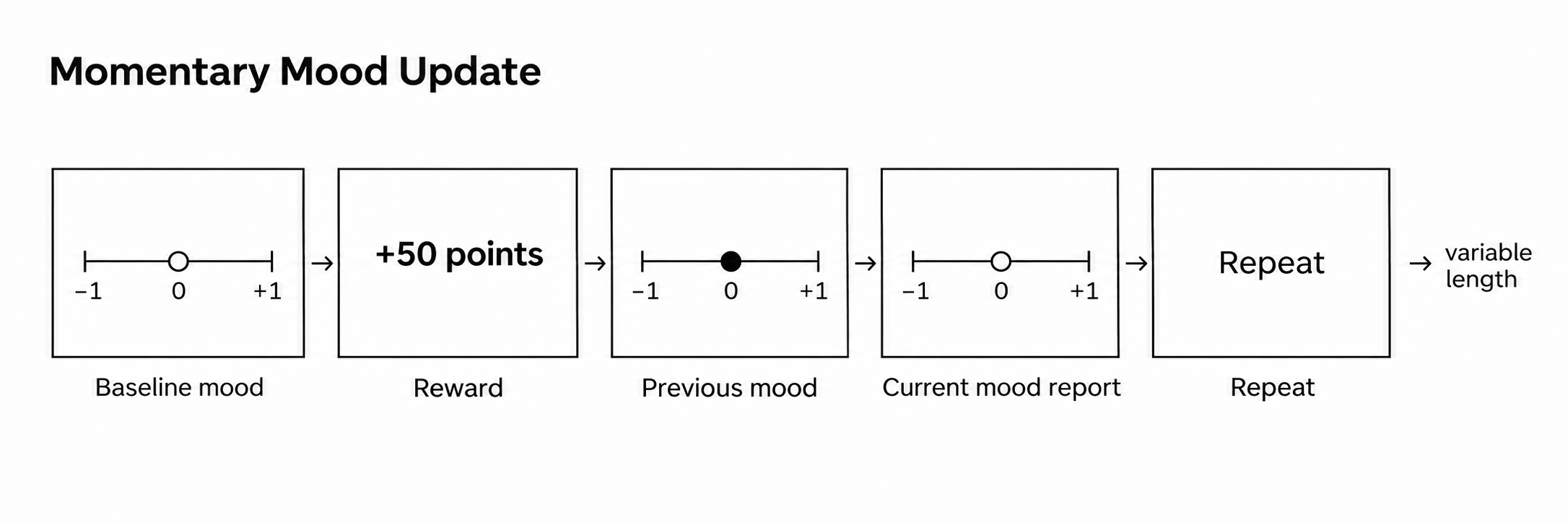}\par\vspace{1mm}
\includegraphics[width=\textwidth,height=0.16\textheight,trim=0 60bp 0 60bp,clip,keepaspectratio]{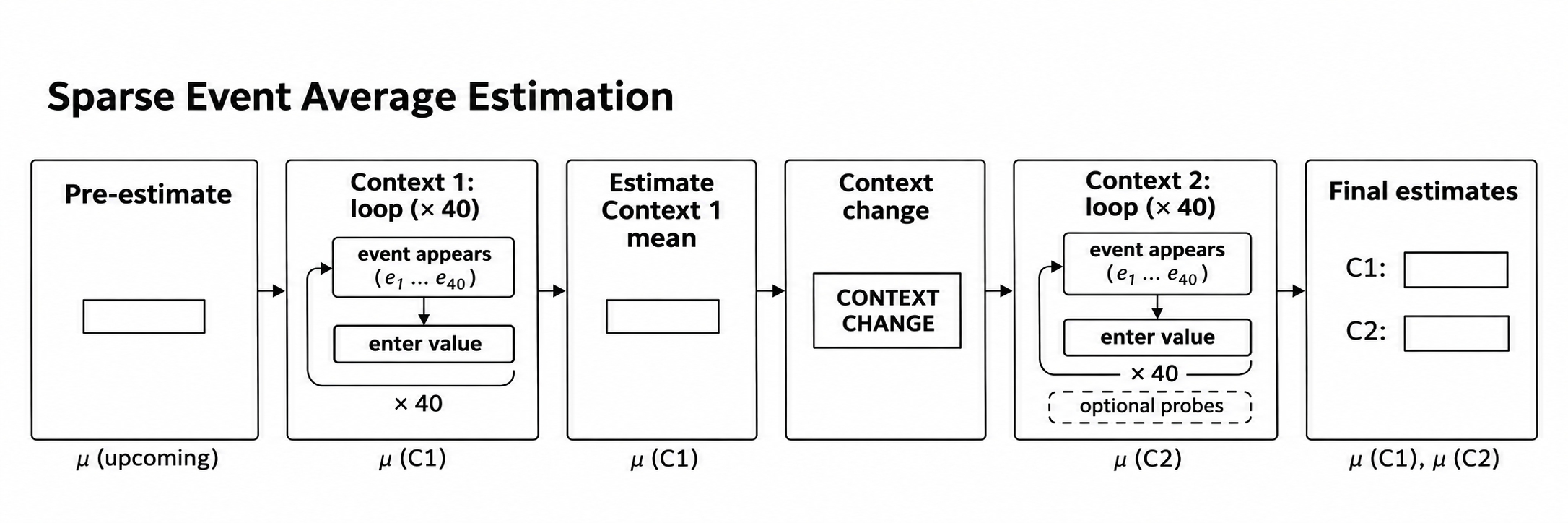}
\end{center}

\clearpage

\end{document}